%% file: main.tex
\documentclass{article}

\usepackage[autonum]{shortex}

\usepackage[accepted]{icml2023}

\theoremstyle{plain}

\theoremstyle{definition}

\theoremstyle{remark}

\usepackage[textsize=tiny]{todonotes}

\icmltitlerunning{MixFlows: principled variational inference via mixed flows}

\begin{document}

\twocolumn[
\icmltitle{MixFlows: principled variational inference via mixed flows}

\begin{icmlauthorlist}
\icmlauthor{Zuheng Xu}{ubc}
\icmlauthor{Naitong Chen}{ubc}
\icmlauthor{Trevor Campbell}{ubc}
\end{icmlauthorlist}

\icmlaffiliation{ubc}{University of British Columbia, Department of Statistics, Vancouver, Canada}

\icmlcorrespondingauthor{Trevor Campbell}{trevor@stat.ubc.ca}

\icmlkeywords{Machine Learning, ICML}

\vskip 0.3in
]

\printAffiliationsAndNotice{}  

\begin{abstract}
This work presents \emph{mixed variational flows (MixFlows)}, a new variational
family that 
consists of a mixture of repeated applications of a map to an initial
reference distribution. First, we provide efficient algorithms for 
\iid sampling, density evaluation, and unbiased ELBO estimation.
We then show that MixFlows have MCMC-like convergence
guarantees when the flow map is ergodic and measure-preserving,
and provide bounds on the accumulation of error for practical 
implementations where the flow map is approximated.
Finally, we develop an implementation of MixFlows 
based on uncorrected discretized Hamiltonian dynamics
combined with deterministic momentum refreshment.
Simulated and real data experiments show that MixFlows 
can provide more reliable posterior approximations than several black-box
normalizing flows, 
as well as samples of comparable quality to those obtained from state-of-the-art MCMC
methods.
\end{abstract}

\input{introduction}

\input{background}

\input{mixflows}

\input{guarantees}

\input{hamiltonian_mixflows}

\input{experiments}

\input{conclusion}

\section*{Acknowledgements}
The authors gratefully acknowledge the support
of a National Sciences and Engineering Research Council of Canada
(NSERC) Discovery Grant and a UBC four year doctoral fellowship.

\bibliography{sources}
\bibliographystyle{icml2023}

\newpage
\appendix
\onecolumn
\input{proofs}

\input{apdx_hamflow}
\input{apdx_extensions}

\input{apdx_expt}


\end{document}

%% file: introduction.tex
\section{Introduction}\label{sec:introduction}
Bayesian statistical modelling and inference provides a principled approach to
learning from data.  However, for all but the simplest models, exact
inference is not possible and computational approximations are
required. A standard methodology for Bayesian inference 
is Markov chain Monte Carlo (MCMC) [\citealp{Robert04};
\citealp{Robert11}; \citealp[Ch.~11,12]{Gelman13}], which involves 
simulating a Markov chain whose stationary distribution is the Bayesian posterior distribution, and 
then treating the sequence of states as draws from the posterior.
MCMC methods are supported by theory that guarantees
that if one simulates the chain for long
enough, Monte Carlo averages based on the sequence of states will
converge to the exact posterior expectation of interest (e.g., \cite{Roberts04}). 
This ``exactness'' property is quite compelling: regardless of how well one
is able to tune the Markov chain, the method is guaranteed 
to eventually produce an accurate result given enough computation time.
Nevertheless, it remains a challenge to assess and optimize the performance of MCMC in practice with a finite computational budget.
One option is to use a Stein discrepancy \citep{Gorham15,Liu16,Chwialkowski16,Gorham17,Anastasiou21}, which 
quantifies how well a set of MCMC samples approximates the posterior distribution.
But standard Stein discrepancies are not reliable in
the presence of multimodality \citep{Wenliang20}, are computationally expensive to
estimate, and suffer from the curse of dimensionality, although 
recent work addresses the latter two issues to an extent \citep{Huggins18,Gong21}. 
The other option is to use a traditional diagnostic, e.g., Gelman--Rubin \citep{Gelman92,Brooks98}, effective sample size \citep[p.~286]{Gelman13},
\citet{Geweke92}, or others \citep{Cowles96}. These diagnostics detect mixing issues,
but do not comprehensively quantify how well the MCMC samples approximate the posterior. 

Variational inference (VI) \citep{Jordan99,Wainwright08,Blei17} is an alternative 
to MCMC that does provide a straightforward quantification of posterior approximation error.
In particular, VI involves approximating the posterior with a probability distribution---typically selected
from a parametric family---that enables both \iid sampling and density
evaluation \citep{Wainwright08,Rezende15,Ranganath16,Papamakarios21}.
Because one can both obtain \iid draws and evaluate the density, one can estimate the 
ELBO \cite{Blei17}, i.e., the Kullback-Leibler (KL) divergence \citep{Kullback51} to the posterior up to a constant.
The ability to estimate this quantity, in turn, enables 
scalable tuning via straightforward stochastic gradient descent algorithms \citep{Hoffman13,Ranganath14},
optimally mixed approximations
\citep{Jaakkola98,Gershman12,Zobay14,Guo16,Wang16,Miller17,Locatello18,Locatello18b,Campbell19b},
model selection \citep{Corduneau01,Masaaki01,Constantinopoulos06,Ormerod17,CheriefAbdellatif18,Tao18}, and more.
However, VI typically does not possess the same ``exactness regardless of tuning'' that MCMC does.
The optimal variational distribution is not usually equal to the posterior due
to the use of a limited parametric variational family; and even if it were, one
could not reliably find it due to nonconvexity of the KL objective
\citep{Xu22}. Recent work addresses this problem by
constructing variational families from parametrized Markov chains targeting the posterior.
Many are related to annealed importance sampling \citep{Salimans15,Wolf16,Geffner21,Zhang21,Thin21,Jankowiak21};
these methods introduce numerous auxiliary variables and only have convergence guarantees 
in the limit of increasing dimension of the joint distribution.
Those based on flows \citep{Neal05,Caterini18,Chen22} avoid the
increase in dimension with flow length, 
but typically do not have guaranteed convergence to the target.
Methods involving the final-state marginal of finite simulations of standard MCMC methods (e.g., \citealp{Zhang20}) do 
not enable density evaluation.

\textbf{The first contribution} of this work is a new family of \emph{mixed variational flows (MixFlows)}, 
constructed via averaging over repeated applications of a pushforward map to an 
initial reference distribution. We develop efficient methods for \iid sampling,
density evaluation, and unbiased ELBO estimation.
\textbf{The second contribution} is a theoretical analysis of MixFlows.
We show (\cref{thm:weakconvergence,thm:tvconvergence}) that when the 
map family is ergodic and measure-preserving, MixFlows 
converge to the target distribution for any value of the variational parameter, and hence
have guarantees regardless of tuning as in MCMC. 
We then extend these results (\cref{thm:approxtvconvergence,cor:approxtvconvergence}) to MixFlows based
on approximated maps---which are typically necessary in practice---with bounds
on error with increasing flow length.
\textbf{The third contribution} of the work is an implementation of 
MixFlows via uncorrected discretized Hamiltonian dynamics. 
Simulated and real data experiments 
compare performance to the No-U-Turn sampler (NUTS) \citep{Hoffman14},
standard Hamiltonian Monte Carlo (HMC) \citep{Neal11}, 
and several black-box
normalizing flows \citep{rezende2015variational,Dinh17}. Our results
demonstrate a comparable sample quality to  
NUTS, similar computational efficiency to HMC, and more reliable posterior
approximations than standard normalizing flows.

\paragraph*{Related work.}
Averages of Markov chain state distributions in general were studied in early work
on shift-coupling \citep{Aldous93}, with convergence guarantees established by 
\citet{Roberts97}. However, these guarantees involve minorization and drift conditions
that are designed for stochastic Markov chains, and
are not applicable to MixFlows.
Averages of deterministic pushforwards specifically to enable density evaluation 
have also appeared
in more recent work. \citet{Rotskoff19,Thin21b} use an average of pushforwards generated by
simulating nonequilibrium dynamics as an importance sampling proposal. 
The proposal distribution itself does not come with any convergence
guarantees---due to the use of non-measure-preserving, 
non-ergodic damped Hamiltonian dynamics---or tuning guidance.
Our work provides comprehensive convergence theory and
establishes a convenient means of optimizing hyperparameters. 

MixFlows are also related to past work on deterministic 
MCMC \citep{Murray12,Neal12,ver2021hamiltonian,neklyudov2021deterministic}.
\citet{Murray12} developed a Markov chain Monte Carlo procedure based on an arbitrarily
dependent random stream via augmentation and measure-preserving bijections. 
\citet{ver2021hamiltonian} designed a specialized momentum
distribution that generates valid Monte Carlo samples solely through the
simulation of deterministic Hamiltonian dynamics. 
\citet{neklyudov2021deterministic} proposed a general
form of measure-preserving dynamics that can be utilized to construct
deterministic Gibbs samplers.
These works all involve only deterministic updates,
but do not construct variational approximations,  
provide total variation convergence guarantees,
or provide guidance on hyperparameter tuning. 
Finally, some of these works involve discretization
of dynamical systems, but do not characterize
the resultant error \cite{ver2021hamiltonian,neklyudov2021deterministic}.
In contrast, our work provides a comprehensive convergence theory,
with error bounds for when approximate maps are used.

%% file: background.tex
\section{Background}\label{sec:background}

\subsection{Variational inference with flows}\label{sec:viflows}
Consider a set $\mcX \subseteq \reals^d$ 
and a target probability distribution $\pi$ on $\mcX$
whose density with respect to the Lebesgue measure we denote $\pi(x)$ for $x\in\mcX$.
In the setting of Bayesian inference, $\pi$ is the posterior distribution 
that we aim to approximate, and we are only able to evaluate a function $p(x)$
such that $p(x) = Z\cdot\pi(x)$ for some unknown normalization constant $Z>0$.
Throughout,  we will assume all distributions have densities with respect to the Lebesgue measure on $\mcX$,
and will use the same symbol to denote a distribution and its density; it will be clear from context what is meant.

Variational inference involves approximating the target distribution $\pi$
by minimizing the Kullback-Leibler (KL) divergence
from $\pi$ to members of a parametric family $\{q_\lambda : \lambda \in \Lambda\}$,
$\Lambda \subseteq \reals^p$, i.e.,
\[ \label{eq:klmin}
\begin{aligned} 
\lambda^\star &= \argmin_{\lambda\in\Lambda}\kl{q_\lambda}{\pi} \\
& = \argmin_{\lambda\in\Lambda}\int q_\lambda(x)\log\frac{q_\lambda(x)}{p(x)}\dee x.
\end{aligned}
\]
The two objective functions in \cref{eq:klmin} differ only by the constant $\log Z$. 
In order to be able to optimize $\lambda$ using standard techniques,
the variational family $q_\lambda$, $\lambda\in\Lambda$ must 
enable both \iid sampling and density evaluation.
A common approach to constructing such a family
is to pass draws from a simple reference distribution $q_0$
through a measurable function 
$T_\lambda : \mcX \to \mcX$; $T_\lambda$ is often
referred to as a \emph{flow} when comprised of repeated composed functions \citep{Tabak13,Rezende15,Kobyzev21}. 
If $T_\lambda$ is a \emph{diffeomorphism}, i.e., continuously differentiable and has a continuously differentiable inverse, then
we can express the density of $X = T_\lambda(Y)$, $Y \distas q_0$ as
\[
	\forall x\!\in\!\mcX,\, q_\lambda(x) \!\! = \!\!\frac{q_0(T_\lambda^{-1}(x))}{J_\lambda(T_\lambda^{-1}(x))}, \,
	J_\lambda(x) \!\!= \!\!\left| \det \!\nabla_x T_\lambda(x)\right|. \!\! \label{eq:transformationvars}
\]
In this case the optimization in \cref{eq:klmin} can be rewritten using a transformation
of variables as 
\[
	\lambda^\star &= \argmin_{\lambda\in\Lambda}\int q_0(x) \log \frac{q_0(x)}{J_\lambda(x)p(T_\lambda(x))}\dee x.\label{eq:reparammin}
\]
One can solve the optimization problem \cref{eq:reparammin} using unbiased stochastic estimates of 
the gradient\footnote{We assume throughout that differentiation
and integration can be swapped wherever necessary.} with respect to 
$\lambda$ based on draws from $q_0$ \citep{Salimans13,Kingma14b},
\[
	\nabla_\lambda \kl{q_\lambda}{\pi}\! \approx \!\nabla_\lambda \!\log \!\frac{q_0(X)}{J_\lambda(X)p(T_\lambda(X))}, \, X \distas q_0. \!\!\label{eq:gradest}
\]

\subsection{Measure-preserving and ergodic maps}
Variational flows are often constructed from a flexible, general-purpose parametrized
family $\{T_\lambda : \lambda \in \Lambda\}$ that is not specialized for any
particular target distribution \citep{Papamakarios21}; it is the job of the KL divergence minimization
\cref{eq:reparammin} to adapt the parameter $\lambda$ such that 
$q_\lambda$ becomes a good approximation of the target $\pi$. However, 
there are certain
functions---in particular, those that are both \emph{measure-preserving} and
\emph{ergodic} for $\pi$---that naturally provide a means to approximate
expectations of interest under $\pi$ without the need for tuning.
Intuitively, a measure-preserving map $T$ will not change the
distribution of draws from $\pi$: if $X \distas \pi$, then $T(X) \distas \pi$.
And an ergodic map $T$, when applied repeatedly, will not get ``stuck'' in 
a subset of $\mcX$ unless it has probability either 0 or 1 under $\pi$.
The precise definitions are given in \cref{def:measurepreserving,def:ergodic}.

\bdef[Measure-preserving map {\citep[pp.~73, 105]{Eisner15}}]\label{def:measurepreserving}
$T : \mcX\to\mcX$ is \emph{measure-preserving} for $\pi$ if $T\pi = \pi$, where
$T\pi$ is the pushforward measure given by $\pi(T^{-1}(A))$ for each measurable
set $A\subseteq \mcX$. 
\edef
\bdef[Ergodic map {\citep[pp.~73, 105]{Eisner15}}]\label{def:ergodic}
 $T :\mcX\to\mcX$ is \emph{ergodic} for $\pi$ if for all measurable sets $A\subseteq
\mcX$, $T(A) = A$ implies that $\pi(A) \in \{0,1\}$.  
\edef

If a map $T$ satisfies both \cref{def:measurepreserving,def:ergodic}, then long-run averages
resulting from repeated applications of $T$ will converge to expectations under $\pi$,
as shown by \cref{thm:pointwiseergodic}.
When $\mcX$ is compact, this result shows that the discrete 
measure $\frac{1}{N}\sum_{n=0}^{N-1} \delta_{T^n x}$ converges weakly to $\pi$ \citep[Theorem 6.1.7]{Dajani08}.
\bthm[{Ergodic Theorem [\citealp{Birkhoff31}; \citealp[p.~212]{Eisner15}]}]\label{thm:pointwiseergodic}
Suppose $T : \mcX \to \mcX$ is measure-preserving and ergodic for $\pi$,
and $f \in L^1(\pi)$. Then
\[
	\lim_{N\to\infty} \frac{1}{N}\sum_{n=0}^{N-1} f(T^nx) = \int f\dee \pi
	\qquad \text{$\pi$-\aev}\, x\in\mcX.
\]
\ethm
Based on this result, one might reasonably consider building a measure-preserving variational 
flow, i.e., $X = T(X_0)$ where $X_0 \distas q_0$. However, 
it is straightforward to show that measure-preserving bijections $T$ do not 
decrease the KL divergence (or any other $f$-divergence, e.g., total variation and Hellinger \citep[Theorem 1]{qiao2010study}):
\[
\kl{Tq}{\pi} = \kl{T q}{T \pi} = \kl{q}{\pi}.
\]

%% file: mixflows.tex
\section{Mixed variational flows (MixFlows)}\label{sec:ergodicflows}
In this section, we develop a general family of \emph{mixed variational flows} (MixFlows)
as well as algorithms for tractable \iid sampling, density evaluation, and ELBO estimation.
MixFlows consist of a mixture of normalizing flows obtained via repeated application 
of a pushforward map. This section introduces general MixFlows; later in \cref{sec:guarantees,sec:hamiltonianmixflows} we will 
provide convergence guarantees and examples based on Hamiltonian dynamics.

\subsection{Variational family}
Define a reference distribution $q_0$ on $\mcX$ for which \iid sampling and density evaluation is tractable, 
and a collection of measurable functions $T_\lambda : \mcX\to\mcX$ parametrized by $\lambda \in \Lambda$.
Then the MixFlow family generated by $q_0$ and $T_\lambda$  is 
\[
q_{\lambda,N} = \frac{1}{N} \sum_{n=0}^{N-1} T^n_\lambda q_0 \qquad \text{for} \qquad \lambda \in \Lambda, \, N \in \nats,
\]
where $T^n_\lambda q_0$ denotes the pushforward of the distribution $q_0$ under
$n$ repeated applications of $T_\lambda$. 

\subsection{Density evaluation and sampling} \label{sec:densityandsample}

If $T_\lambda:\mcX\to\mcX$ is a diffeomorphism with Jacobian $J_\lambda : \mcX\to\reals$,
we can express the density of $q_{\lambda,N}$ by 
using a transformation of variables formula on each component in the mixture:
\[
q_{\lambda,N}(x) &= \frac{1}{N}\sum_{n=0}^{N-1} \frac{q_0(T_\lambda^{-n}x)}{\prod_{j=1}^n J_\lambda(T_\lambda^{-j}x)}.
\]
This density can be computed efficiently using $N-1$ evaluations of $T^{-1}_\lambda$ and $J_\lambda$ each (\cref{alg:density}).
For sampling, we can obtain an independent draw $X\distas q_{\lambda,N}$ by treating $q_{\lambda,N}$ 
as a mixture of $N$ distributions:
\[
K &\distas\distUnif\{0, 1, \dots, N-1\} \quad X_0 \distas q_0 \quad X = T^K_\lambda(X_0).
\]
The procedure is given in \cref{alg:draw}. On average, this computation 
requires $\EE[K] = \frac{N-1}{2}$ applications of $T_\lambda$, and at most it requires $N-1$ applications.
However, one often takes samples from $q_{\lambda, N}$ to estimate the
expectation of some test function $f:\mcX \to \reals$. In this case, one
can use all intermediate states over a single pass of a trajectory rather than individual \iid draws.
In particular, we obtain an unbiased estimate of $\int f(x)q_{\lambda,N}(\dee x)$ via
\[
  X_0 \sim q_0 \quad \frac{1}{N}\sum_{n = 0}^{N-1}f(T_\lambda^n X_0). \label{eq:traj_average} 
\]
We refer to this estimate as the \emph{trajectory-averaged estimate} of $f$.
The trajectory-averaged estimate of $f$ is preferred over the na\"ive estimate
based on a single draw $f(X), X\sim q_{\lambda, N}$,
as its cost is of the same order ($N-1$ applications of $T_\lambda$)
and its variance is bounded above by that of 
the na\"ive estimate $f(X)$, as shown in \cref{prop:smallervar}.  
See \cref{apdx:2d}, and \cref{fig:var_compare} in particular, for empirical verification of \cref{prop:smallervar}.
\bprop\label{prop:smallervar}
\[
 \var\left[\frac{1}{N} \sum_{n =0}^{N-1} f(T^n X_0)\right] \leq \var\left[f(X)\right].
\]
\eprop

\subsection{ELBO estimation}\label{sec:param_opt}

We can minimize the KL divergence from $q_{\lambda,N}$ to $\pi$
by maximizing the ELBO \citep{Blei17}, given by 
\[
\mathrm{ELBO}\left(\lambda,N\right) &= \int q_{\lambda,N}(x)\log \frac{p(x)}{q_{\lambda,N}(x)}\dee x\\
&= \int q_0(x)\left(\frac{1}{N}\sum_{n=0}^{N-1}\log \frac{p(T_\lambda^nx)}{q_{\lambda,N}(T^n_\lambda x)}\dee x\right).
\]
The trajectory-averaged ELBO estimate is thus
\[\label{eq:betterelbo}
\!\!X_0 \distas q_0, \,
\widehat{\mathrm{ELBO}}(\lambda,N) 
\!= \!\frac{1}{N}\sum_{n=0}^{N-1} \log  \frac{p(T_\lambda^n X_0)}{q_{\lambda, N}(T_\lambda^n X_0)}.\!\!
\]
The na\"ive method to compute this estimate---sampling $X_0$ and then computing the log density ratio for each term---requires $O(N^2)$
computation because each evaluation of $q_{\lambda,N}(x)$ is $O(N)$.
\cref{alg:betterelbo} provides an efficient way of computing $\widehat{\text{ELBO}}(\lambda, N)$ in $O(N)$
operations, which is on par with taking a single draw from $q_{\lambda,N}$ or evaluating $q_{\lambda,N}(x)$ once.
The key insight in \cref{alg:betterelbo} is that we can evaluate the collection of values 
$\{ q_{\lambda, N}(X_0), q_{\lambda, N}(T_\lambda X_0), \dots, q_{\lambda, N}(T_\lambda^{N-1}X_0) \}$
incrementally, starting from $q_{\lambda, N}(X_0)$ and iteratively computing each
$q_{\lambda, N}(T_\lambda^n X_0)$ for increasing $n$ in constant time. 
Specifically, in practice one computes $ q_{\lambda, N}(x)$ with a complexity of $O(N)$ and iteratively
updates $ q_{\lambda, N}(T_\lambda^{k}x)$ and the Jacobian for $k = 1, 2, \dots N-1$.
Each update requires only constant cost
if one precomputes and stores $T^{-N+1}_\lambda x, \dots, T^{N-1}_\lambda x$ and 
$J_\lambda(T_\lambda^{-N+1}x), \dots, J_\lambda(T_\lambda^{N-1}x)$.
This then implies that \cref{alg:betterelbo} requires $O(N)$ memory; 
\cref{fig:efficientelbo} in \cref{sec:efficient_elbo} provides a slightly more complicated $O(N)$ time, $O(1)$ 
memory implementation of \cref{eq:betterelbo}.

\begin{algorithm}[t!]
\caption{$\texttt{Sample}(q_{\lambda,N})$: Take a draw from $q_{\lambda,N}$}\label{alg:draw}
\begin{algorithmic}
\STATE{\bfseries Input:} reference distribution $q_0$, flow map $T_\lambda$, number of steps $N$
\STATE $K \gets \texttt{Sample}(\distUnif\{0, 1, \dots, N-1\})$
\STATE $x_0 \gets \texttt{Sample}(q_0)$
\STATE{\bfseries Return:} $T_\lambda^K(x_0)$
\end{algorithmic}
\end{algorithm}

\begin{algorithm}[t!]
\caption{$\log q_{\lambda,N}(x)$: Evaluate the log-density of $q_{\lambda,N}$}\label{alg:density}
\begin{algorithmic}
\STATE{\bfseries Input:} location $x$, reference distribution $q_0$, flow map $T_\lambda$, Jacobian $J_\lambda$, number of steps $N$
\STATE $L \gets 0$
\STATE $w_0 \gets \log q_0(x)$
\FOR{$n=1, \dots, N-1$}
\STATE $x \gets T^{-1}_\lambda(x)$
\STATE $L \gets L + \log J_\lambda(x)$
\STATE $w_n \gets \log q_0(x) - L$
\ENDFOR
\STATE{\bfseries Return:} $\texttt{LogSumExp}(w_0, \dots, w_{N-1}) - \log N$
\end{algorithmic}
\end{algorithm}

\begin{algorithm}[t!]
\caption{$\texttt{EstELBO}(\lambda,N)$: Obtain an unbiased estimate of the ELBO for $q_{\lambda,N}$. }\label{alg:betterelbo}
\begin{algorithmic}
\STATE{\bfseries Input:} reference $q_0$, unnormalized target $p$, flow map $T_\lambda$, Jacobian $J_\lambda$, number of flow steps $N$

\STATE $x_0 \gets \texttt{Sample}(q_0), \quad J_0 \gets J_\lambda(x_0)$
\FOR{$n=1,\dots,N-1$}
\STATE $x_n \gets T_\lambda(x_{n-1}), \quad J_n \gets J_\lambda(x_n)$ 
\STATE $x_{-n} \gets T_\lambda^{-1}(x_{-n+1}), \quad J_{-n} \gets J_\lambda(x_{-n})$ 
\ENDFOR
\STATE $z_0 \gets q_{\lambda, N}(x_0)$ (via \cref{alg:density})
\STATE $J \gets \prod_{j=1}^{N-1} J_{-j} $
\FOR{$n=1,\dots,N-1$}
\STATE $\bq_n \gets \frac{1}{N} q_0(x_{-N+n})/J $
\STATE $z_n \gets \left(z_{n-1} - \bq_n\right)/J_{n-1} + \frac{1}{N}q_0(x_n)$
\IF{$n < N-1$}
\STATE $J \gets J \cdot J_{n-1}/ J_{-N+n}$
\ENDIF
\ENDFOR
\STATE $\widehat{\mathrm{ELBO}}(\lambda,N) \gets \frac{1}{N}\sum_{n=0}^{N-1}
(\log p(x_n) - \log z_n)$ 
\STATE{\bfseries Return:} $\widehat{\mathrm{ELBO}}(\lambda,N)$
\end{algorithmic}
\end{algorithm}

%% file: guarantees.tex
\section{Guarantees for MixFlows}\label{sec:guarantees}
In this section, we show that when $T_\lambda$ is 
measure-preserving and ergodic for $\pi$---or approximately so---MixFlows come with 
convergence guarantees and bounds on their approximation error as a function of flow length.
Proofs for all results may be found in \cref{sec:proofs}.

\subsection{Ergodic MixFlows}\label{sec:exactergflow}
\emph{Ergodic MixFlow families} are those where 
$\forall \lambda\in\Lambda$, $T_\lambda$ is measure-preserving and ergodic for $\pi$.
In this setting, \cref{thm:weakconvergence,thm:tvconvergence}
show that MixFlows
converge to the target weakly and in total variation as $N\to\infty$ regardless of
the choice of $\lambda$ (i.e., regardless of parameter tuning effort). 
Thus, ergodic MixFlow families provide
the same compelling convergence result as MCMC, but with the added benefit of
unbiased ELBO estimates.
We first demonstrate setwise and weak convergence.
Recall that a sequence of distributions $q_n$ converges \emph{weakly} to $\pi$
if for all bounded, continuous $f:\mcX \to \reals$, 
 $\lim_{n\to\infty} \int f(x) q_n(\dee x) = \int f(x) \pi(\dee x)$, 
and converges \emph{setwise} to $\pi$ if for all measurable $A\subseteq \mcX$, 
$\lim_{n\to\infty} q_n(A) = \pi(A)$.
\bthm\label{thm:weakconvergence}
Suppose $q_0 \ll \pi$ and $T_\lambda$ is measure-preserving and ergodic for $\pi$. Then
$q_{\lambda,N}$ converges both setwise and weakly to $\pi$ as $N\to\infty$.
\ethm
Using \cref{thm:weakconvergence} as a stepping stone, we can obtain convergence in total variation.
Recall that a sequence of distributions $q_n$ converges in \emph{total variation} to $\pi$
if $\tvd{q_n}{\pi} = \sup_{A} \left|q_n(A) - \pi(A)\right| \to 0$ as $n\to\infty$.
Note that similar nonasymptotic results exist for the
ergodic average law of Markov chains \citep{Roberts97}, but as previously
mentioned, these results do not apply to deterministic Markov kernels.

\bthm\label{thm:tvconvergence}
Suppose $q_0 \ll \pi$ and $T_\lambda$ is measure-preserving and ergodic for $\pi$. Then
$q_{\lambda,N}$ converges in total variation to $\pi$ as $N\to\infty$.
\ethm

\subsection{Approximated MixFlows}\label{sec:approxergflow}
In practice, it is rare to be able to evaluate a measure-preserving
map exactly; but \emph{approximations} are commonly available. For example,
in \cref{sec:hamiltonianmixflows} we will create a measure-preserving map
using Hamiltonian dynamics, and then approximate that map with a discretized
leapfrog integrator. We therefore need to extend the result in
\cref{sec:exactergflow}
to apply to approximated maps.

Suppose we have two maps, $T$ and $\hT$,
with corresponding MixFlow families $q_{n}$ and $\hq_{n}$
(suppressing $\lambda$ for brevity). 
\cref{thm:approxtvconvergence} shows that 
the error of the MixFlow family $\hq_N$ reflects an accumulation of the difference between the pushforward of
each $q_n$ under $\hT$ and $T$. 
\bthm\label{thm:approxtvconvergence}
Suppose $\hT$ is a bijection. Then
\[
 \tvd{\hq_{N}}{\pi}\leq   \tvd{q_{N}}{\pi}+ \sum_{n = 0}^{N-1} \frac{n}{N}\tvd{T q_n}{\hT q_n}.
\]
\ethm
Suppose $T$ is measure-preserving and ergodic for $\pi$ and $q_0 \ll N$.
Then \cref{thm:tvconvergence} implies that $\tvd{q_N}{\pi}\to 0$,
and for the second term, 
we (very loosely) expect that $\tvd{T q_n}{\hT q_n} \approx \tvd{\pi}{\hT \pi}$ for large $n$. 
Therefore, the second term should behave like  $O(N \cdot \tvd{\pi}{\hT \pi})$, i.e., increase
linearly in $N$ proportional to how ``non-measure-preserving'' $\hT$ is.
This presents a trade-off between flow length and approximation quality: better approximations
enable longer flows and lower minimal error bounds. 
Our empirical findings in \cref{sec:experiments} generally confirm this behavior.
\cref{cor:approxtvconvergence} further provides a more explicit characterization of 
this trade-off when $\hT$ and its log-Jacobian $\log \hJ$ are uniformly close to 
their exact counterparts $T$ and $\log J$, and 
$\log q_n$ and $\log J$ are uniformly (over $n \in \nats$) Lipschitz continuous.
The latter assumption is reasonable in practical settings where
we observe that the log-density of $q_n$ closely approximates $\pi$ (see the experimental results in \cref{sec:experiments}).
\bcor \label{cor:approxtvconvergence}
Suppose that for all $x\in\mcX$,
\[
\max\left\{ \|\hT^{-1}x - T^{-1}x\|, |\log \hJ(x) - \log J(x)|\right\} \leq \eps,
\]
for all  $n\in [N-1]$, $\log q_n$ and $\log J$ are $\ell$-Lipschitz continuous,
and that $T, \hT$ are diffeomorphisms. Then
\[
    \tvd{\hq_{N}}{\pi}\leq   \tvd{q_{N}}{\pi}+ N\epsilon (\ell+1) e^{(2\ell+1)\epsilon}.
\]
\ecor
This result states that the overall map-approximation error is $O(N\eps)$ when $\eps$ is small.
Theorem 1 of \citet{Butzer71} also suggests that $\tvd{q_N}{\pi} = O(1/N)$ in many 
cases, which hints that the bound should decrease roughly until
$N = O(1/\sqrt{\epsilon})$, with error bound $O(\sqrt{\epsilon})$.
We leave a more careful investigation of this trade-off for future work.

\subsection{Discussion}
Our main convergence results (\cref{thm:weakconvergence,thm:tvconvergence}) require
that $\pi$ dominates the reference $q_0$, and that
$T$ is both measure-preserving and ergodic.
It is often straightforward to design a dominated reference $q_0$.
Designing a measure-preserving
map $T$ with an implementable approximation $\hT\approx T$ is also often 
feasible. However, verifying the ergodicity of $T$ conclusively
is typically a very challenging task.
As a
consequence, most past work involving measure-preserving transformations just
asserts the \emph{ergodic hypothesis} without proof; see the discussions in 
\citet[p.~4]{ver2021hamiltonian} and \citet[p.~2-3]{Tupper05}.  

But because MixFlows provide the ability to estimate the Kullback-Leibler divergence
up to a constant, it is not critical
to prove convergence a priori (as it is in the case of MCMC, for example). Instead, we suggest 
using the results of
\cref{thm:weakconvergence,thm:tvconvergence,thm:approxtvconvergence,cor:approxtvconvergence} as 
a guiding recipe for constructing principled variational families.
First, start by designing a family of measure-preserving maps $T_\lambda$, $\lambda \in \Lambda$.
Next, approximate $T_\lambda$ with some tractable $T_{\lambda,\epsilon}$
including a tunable fidelity parameter $\epsilon \geq 0$, such as that $\epsilon \to 0$, $T_{\lambda,\epsilon}$ becomes 
closer to $T_\lambda$. Finally, build a MixFlow from $T_{\lambda,\epsilon}$, and tune both $\lambda$ and $\epsilon$ by maximizing the ELBO.
We follow this recipe in \cref{sec:hamiltonianmixflows} and verify it empirically in \cref{sec:experiments}.

%% file: hamiltonian_mixflows.tex
\section{Uncorrected Hamiltonian MixFlows}\label{sec:hamiltonianmixflows}
In this section, we provide an example of how to design a MixFlow by
starting from an exactly measure-preserving map and then creating a tunable approximation to it.
The construction is inspired by 
Hamiltonian Monte Carlo (HMC) \citep{Neal11,Neal96}, in which each iteration involves simulating 
Hamiltonian dynamics followed by a stochastic momentum refreshment; our method replaces the stochastic
refreshment with a deterministic transformation.
In particular, consider the \emph{augmented} target density
on $\mcX\times \reals^d \times [0,1]$,
\[
\bpi(x,\rho,u)\! =\! \pi(x)m(\rho) \ind[0\!\leq\! u\! \leq\! 1],  \;\; m(\rho)\! =\! \prod_{i=1}^d r(\rho_i),
\]
with auxiliary variables $\rho\in\reals^d$, $u\in[0,1]$, and some almost-everywhere 
differentiable univariate probability density $r : \reals\to\reals_+$.
The $x$-marginal distribution of $\bpi$ is the original target distribution
$\pi$.

\subsection{Measure-preserving map via Hamiltonian dynamics}
We construct $T_\lambda$ by composing the following three steps,
which are all measure-preserving for $\bpi$:

\paragraph{(1) Hamiltonian dynamics} We first apply 
\[
 (x', \rho') \gets H_L(x, \rho),
\] 
where $H_L : \mcX\times \reals^d \to \mcX \times \reals^d$ 
is the map of Hamiltonian dynamics with position $x$ and 
momentum $\rho$ simulated for a time interval of length $L\in\reals_+$,
\[
\der{\rho}{t} &= \nabla \log \pi(x) & \der{x}{t} &= -\nabla \log m(\rho). \label{eq:hamdynamics}
\]
One can show that this preserves density (and hence is measure preserving) and
is also unit Jacobian \citep{Neal11}. 
 
\paragraph{(2) Pseudotime shift} Second, we apply a constant shift to the pseudotime variable $u$,
\[
u' \gets u+ \xi \mod 1, 
\]
where $\xi\in\reals$  is a
fixed irrational number (say, $\xi=\pi/16$). As this is a constant shift, it is
unit Jacobian and density-preserving (and hence measure-preserving). The
$u$ component will act as a notion of ``time'' of the flow, and
ensures that the refreshment of $\rho$ in step (3) below will take 
a different form even if $x$ visits the same location again. 

\paragraph{(3) Momentum refreshment} Finally, we refresh each of the momentum variables via
\[
\forall i=1,\dots, d, \quad \rho''_i \gets R^{-1}(R(\rho_i') + z(x_i', u') \mod 1),
\]
where $R$ is the cumulative distribution function (CDF) of density $r$, and $z : \mcX \times [0,1] \to \reals$
is any differentiable function;
this generalizes the map from \citet{Neal12,Murray12} to enable the shift to depend on the state $x$ and pseudotime $u$.
This step (3) is an attempt to replicate the independent resampling of $\rho \distas m$ from HMC using only
deterministic maps.
This map is measure-preserving as it involves mapping the momentum to a $\distUnif[0,1]$ random variable via the CDF,
shifting by an amount that depends only on $x, u$, and then mapping back using the inverse CDF. The 
Jacobian is the momentum density ratio $m(\rho')/m(\rho'')$.

\subsection{Approximation via uncorrected leapfrog integration}
In practice, we cannot simulate the dynamics in step (1) perfectly. Instead, we approximate 
the dynamics in \cref{eq:hamdynamics} by running $L$ steps of the leapfrog
method, where each leapfrog map $\hH_{\epsilon} : \reals^{2d}\to\reals^{2d}$ 
involves interleaving three discrete transformations with step size $\epsilon >0$,
\[
\begin{aligned}
\hrho_{k+1} &= \rho_k + \frac{\epsilon}{2}\nabla \log\pi(x_k) \\
x_{k+1} &= x_{k} - \epsilon \nabla \log m(\hrho_{k+1}) \\
\rho_{k+1} &= \hrho_{k+1} + \frac{\epsilon}{2}\nabla \log\pi(x_{k+1}) 
\end{aligned}.
\label{eq:leapfrog}
\]
Denote the map $T_{\lambda,\epsilon}$ to be the composition of the three steps 
with Hamiltonian dynamics replaced by the leapfrog integrator;
\cref{alg:hamflow} in \cref{sec:hamflow_code} provides the pseudocode.
The final variational tuning parameters for the MixFlow are the step 
size $\epsilon$---which controls how close $T_{\lambda,\epsilon}$
is to being measure-preserving for $\bpi$---the Hamiltonian simulation length $L \in \nats$,
and the flow length $N$. In our experiments we tune the number of leapfrog steps $L$ and the
step size $\epsilon$ by maximizing the ELBO. We also tune the number
of refreshments $N$ to achieve a desirable computation-quality tradeoff by
visually inspecting the convergence of the ELBO.

\subsection{Numerical stability}\label{sec:stability}
Density evaluation (\cref{alg:density}) and ELBO estimation (\cref{alg:betterelbo})
both involve repeated applications of $T_\lambda$ and $T_\lambda^{-1}$.  
This poses no issue in theory, but in
a computer---with floating point computations---one should
code the map $T_\lambda$ and its inverse $T^{-1}_\lambda$ in a numerically
precise way such that $(T^K_\lambda) \circ (T^{-K}_\lambda)$
is the identity map for large $K\in\nats$. \cref{fig:norm_flow_err} in
\cref{apdx:2d} displays the
severity of the numerical error when $T_\lambda$ and $T^{-1}_\lambda$ are not
carefully implemented.
In practice, we check the stability limits of
our flow by taking draws from $q_0$ and evaluating
$T^K_\lambda$ followed by $T^{-K}_\lambda$ (and vice versa) for increasing $K$
until we cannot reliably invert the flow. See \cref{fig:stability} in the
appendix for an example usage of this diagnostic. Note that for sample generation
specifically (\cref{alg:draw}), numerical stability is not 
a concern as it only requires forward evaluation of the map
$T_\lambda$.

%% file: experiments.tex
\section{Experiments}\label{sec:experiments}

\captionsetup[subfigure]{labelformat=empty, width=\columnwidth}
\begin{figure}[t!]
\centering 
\begin{subfigure}[b]{.32\columnwidth}
    \centering 
    \scalebox{1}{\includegraphics[width=\columnwidth]{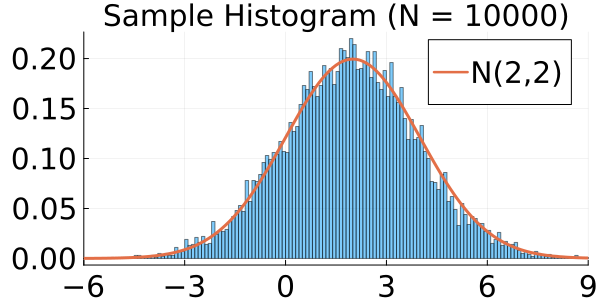}}
    \includegraphics[width=0.48\columnwidth]{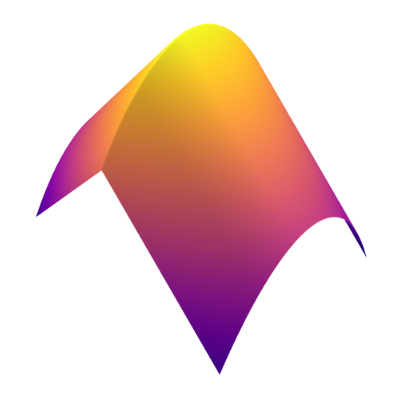}
    \includegraphics[width=0.48\columnwidth]{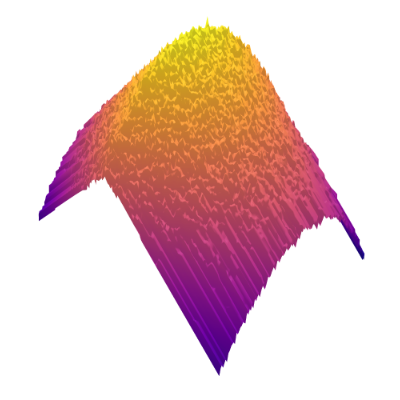}
    \caption{(a) Gaussian  \label{fig:gausshist}}
\end{subfigure}
\hfill
\centering 
\begin{subfigure}[b]{.32\columnwidth} 
    \centering 
    \scalebox{1}{\includegraphics[width=\columnwidth]{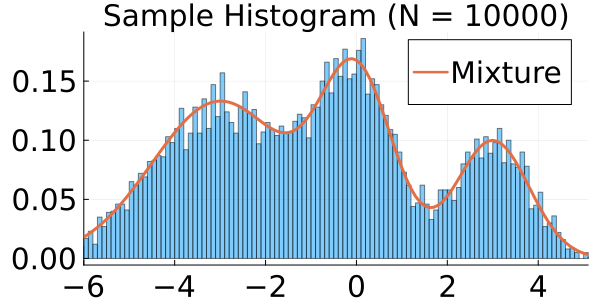}}
    \includegraphics[width=0.48\columnwidth]{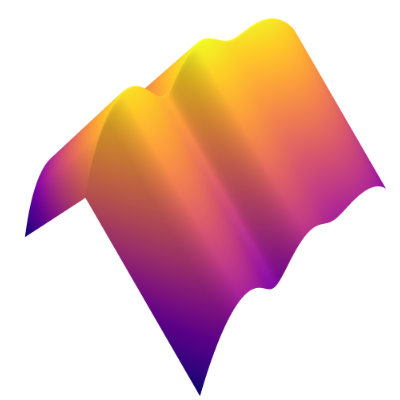}
    \includegraphics[width=0.48\columnwidth]{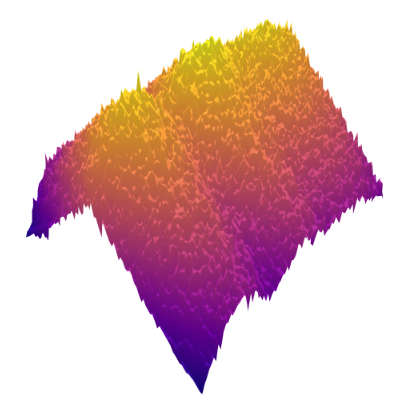}
    \caption{(b) Gaussian mixture\label{fig:mixhist}}
\end{subfigure}
\hfill
\centering
\begin{subfigure}[b]{.32\columnwidth} 
    \centering 
    \scalebox{1}{\includegraphics[width=\columnwidth]{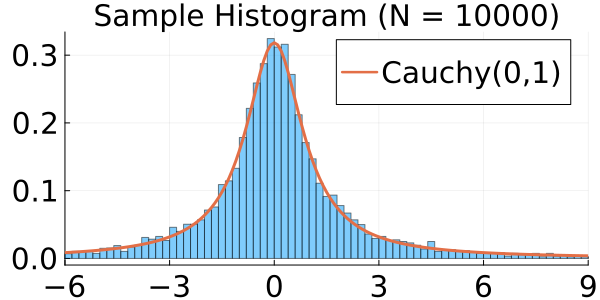}}
    \includegraphics[width=0.48\columnwidth]{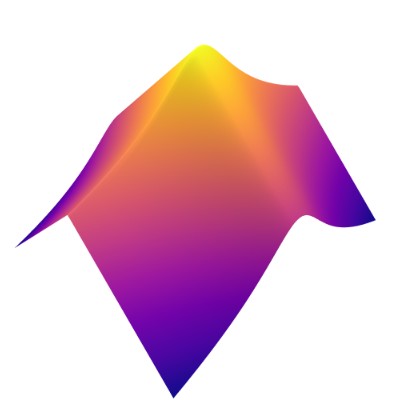}
    \includegraphics[width=0.48\columnwidth]{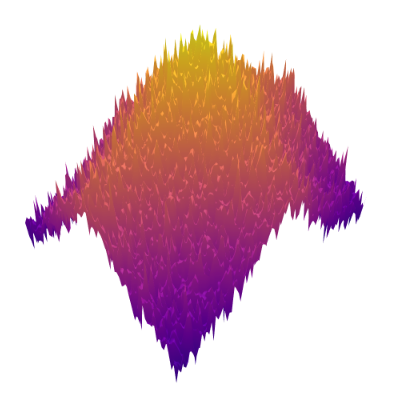}
    \caption{(c) Cauchy \label{fig:cauchyhist}}
\end{subfigure}
\hfill
    \centering 
\begin{subfigure}[b]{.24\columnwidth} 
    \scalebox{1}{\includegraphics[width=\columnwidth, clip]{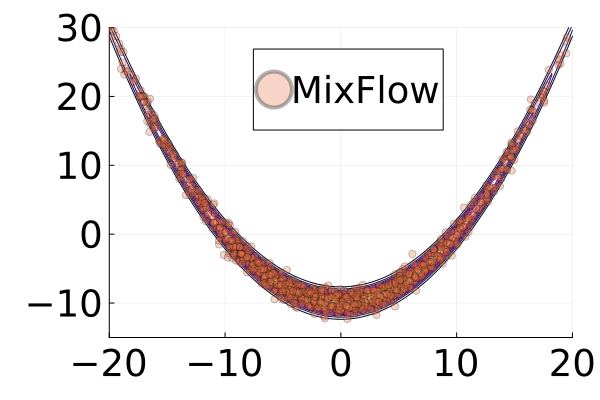}}
    \includegraphics[width=0.48\columnwidth]{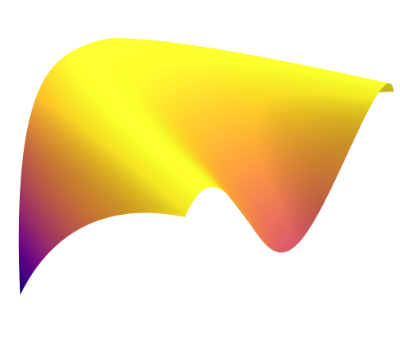}
	\includegraphics[width=0.48\columnwidth]{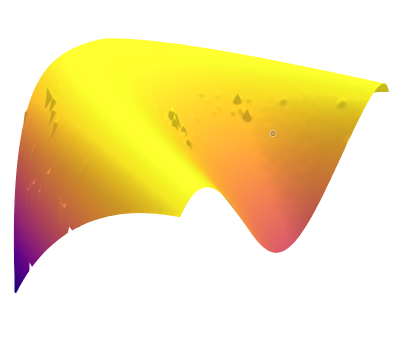}
    \caption{(d) Banana\label{fig:banana_contour}}
\end{subfigure}
\hfill
\centering 
\begin{subfigure}[b]{.24\columnwidth} 
    \scalebox{1}{\includegraphics[width=\columnwidth, clip]{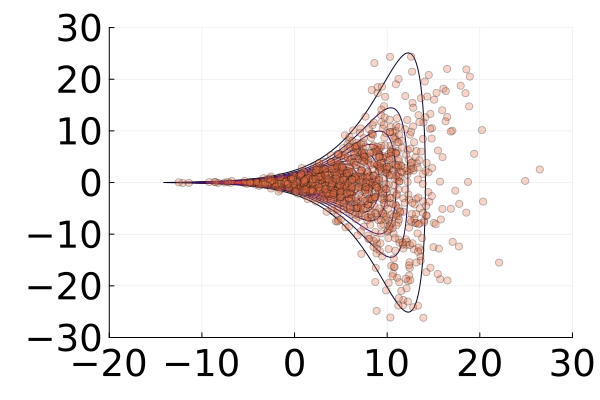}}
	\includegraphics[width=0.48\columnwidth]{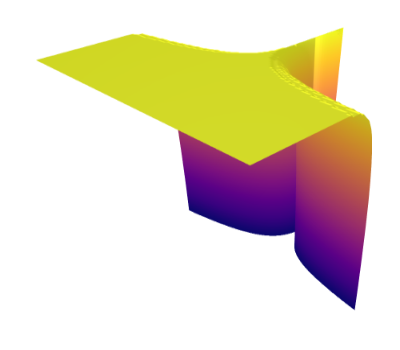}
	\includegraphics[width=0.48\columnwidth]{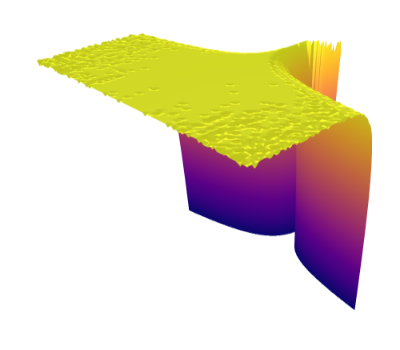}
    \caption{(e) Funnel\label{fig:funnel_contour}}
\end{subfigure}
\hfill
\centering
\begin{subfigure}[b]{0.24\columnwidth}
    \scalebox{1}{\includegraphics[width=\columnwidth, clip]{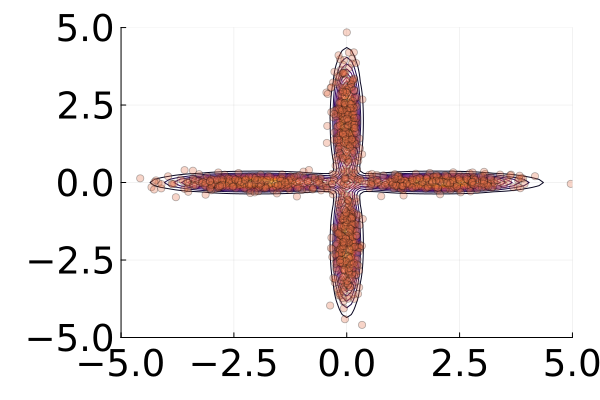}}
	\includegraphics[width=0.48\columnwidth]{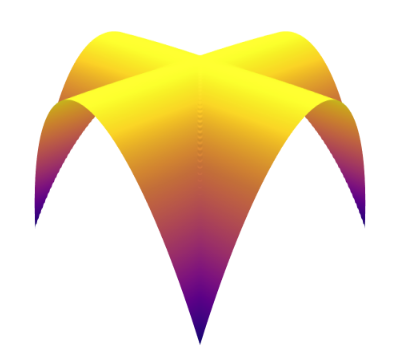}
	\includegraphics[width=0.48\columnwidth]{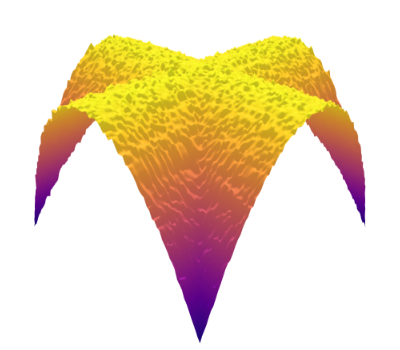}
    \caption{(f) Cross \label{fig:cross_contour}}
\end{subfigure}
\hfill
    \centering 
\begin{subfigure}[b]{.24\columnwidth} 
    \scalebox{1}{\includegraphics[width=\columnwidth, clip]{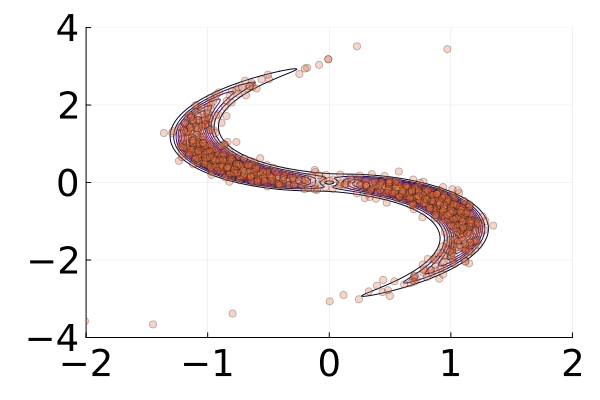}}
	\includegraphics[width=0.48\columnwidth]{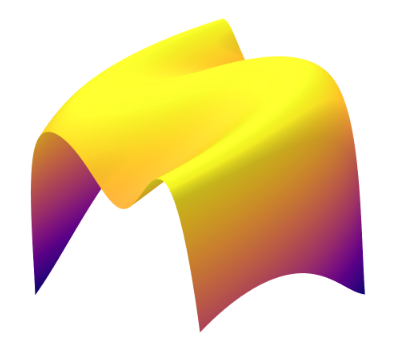}
	\includegraphics[width=0.48\columnwidth]{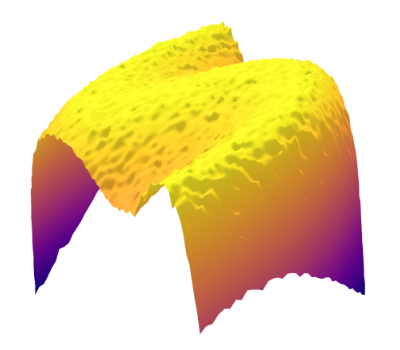}
    \caption{(g) Warp  \label{fig:warp_contour}}
\end{subfigure}
\caption{Marginal samples (first row) and pairs of exact and approximate joint log density (second row) 
for Gaussian (\ref{fig:gausshist}), mixture (\ref{fig:mixhist}), and Cauchy
(\ref{fig:cauchyhist}) targets. 
Marginal samples (third row), pairs of sliced exact and approximate joint log density (fourth row)
for banana (\ref{fig:banana_contour}), funnel
(\ref{fig:funnel_contour}), cross (\ref{fig:cross_contour}), and warped Gaussian
(\ref{fig:warp_contour}).}
\label{fig:vis_plots}
\end{figure}

In this section, we demonstrate the performance of our method
(\texttt{MixFlow}) on 7 synthetic targets and 7 real data targets.
See \cref{apdx:expt_detail} for the details of each target.
Both our synthetic and real data examples are designed to cover a
range of challenging features such as heavy tails, high dimensions,
multimodality, and weak identifiability.  
We compare posterior
approximation quality to several black-box normalizing flow methods (\texttt{NF}):
\texttt{PlanarFlow}, \texttt{RadialFlow}, and \texttt{RealNVP} with various
architectures \citep{Papamakarios21}. 
To make the methods comparable via the ELBO, we train all \texttt{NF}s on the same joint space as \texttt{MixFlow}.
We also compare the marginal sample quality of \texttt{MixFlow} against 
$5{,}000$ samples from \texttt{NUTS} and \texttt{NF}s. 
Finally, we compare sampling time with all competitors, and
effective sample size (ESS) per second with \texttt{HMC}. 
For all experiments, we use the standard Laplace distribution 
as the momentum distribution due to its numerical stability 
(see \cref{fig:stability} in \cref{apdx:expt_detail}). 
Additional comparisons to variational inference based
on uncorrected Hamiltonian annealing (\texttt{UHA}) \citep{Geffner21} and 
nonequilibrium orbit sampling (\texttt{NEO}) \citep[Algorithm 2]{Thin21b} 
may be found in \cref{apdx:expt_detail}.
All experiments were conducted on a machine with an AMD Ryzen 9 3900X and 32GB
of RAM. 
Code is available at \url{https://github.com/zuhengxu/Ergodic-variational-flow-code}.

\subsection{Qualitative assessment} \label{sec:visualization}
We begin with a qualitative examination of the \iid samples and the approximated targets produced
by \texttt{MixFlow} initialized at $q_0 = \distNorm(0,1)$ for three one-dimensional synthetic distributions:
a Gaussian, a mixture of Gaussians, and the standard Cauchy.
We excluded the pseudotime variable $u$ here in order to
visualize the full joint density of $(x,\rho)$ in 2 dimensions.
More details can be found in \cref{apdx:onedim}.
\cref{fig:gausshist,fig:mixhist,fig:cauchyhist} show histograms of $10{,}000$ \iid
$x$-marginal samples generated by \texttt{MixFlow} for each of the three targets, 
which nearly perfectly match the true target marginals.
\cref{fig:gausshist,fig:mixhist,fig:cauchyhist} also show that 
$\log q_{\lambda, N}$ is generally a good approximation of
the log target density. 

We then present similar visualizations on four more challenging synthetic target distributions:
the banana \citep{haario2001adaptive},
Neal's funnel \citep{neal2003slice}, a cross-shaped Gaussian mixture,
and a warped Gaussian.
All four examples have a 2-dimensional state $x\in\reals^2$, and hence $(x,\rho,u) \in \reals^5$.
In each example we set the initial distribution $q_0$ to be the mean-field Gaussian approximation. 
More details can be found in \cref{apdx:2d}.
\cref{fig:banana_contour,fig:funnel_contour,fig:cross_contour,fig:warp_contour} shows the scatter plots consisting of $1{,}000$ \iid 
$x$-marginal samples drawn from \texttt{MixFlow}, as well as 
the approximated \texttt{MixFlow} log density and exact log density 
sliced as a function of $x\in\reals^2$ for 
a single value of $(\rho, u)$ chosen randomly via $(\rho, u) \sim \distLaplace(0, I) \times \distUnif[0,1]$
(which is required for visualization, as $(x,\rho,u) \in \reals^5$). 
We see that, qualitatively, both the samples and approximated densities from \texttt{MixFlow} closely match the target. 
Finally, \cref{fig:2d_sample} in \cref{apdx:2d} provides a more comprehensive set of sample
histograms (showing the $x$-, $\rho$-, and $u$-marginals). 
These visualizations support our earlier theoretical analysis.

\subsection{Posterior approximation quality} 

\captionsetup[subfigure]{labelformat=empty}
\begin{figure*}[t!]
\centering 
\begin{subfigure}[b]{.12\columnwidth} 
    \scalebox{1}{\includegraphics[width=\columnwidth]{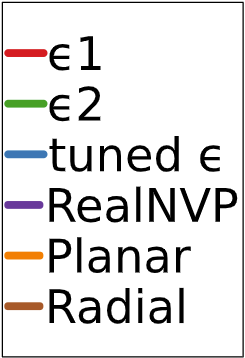}}
    \scalebox{1}{\includegraphics[width=\columnwidth]{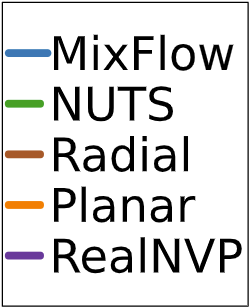}}
    \caption{}
\end{subfigure}
\hfill
\centering 
\addtocounter{subfigure}{-1}
\begin{subfigure}[b]{.26\columnwidth} 
    \scalebox{1}{\includegraphics[width=\columnwidth]{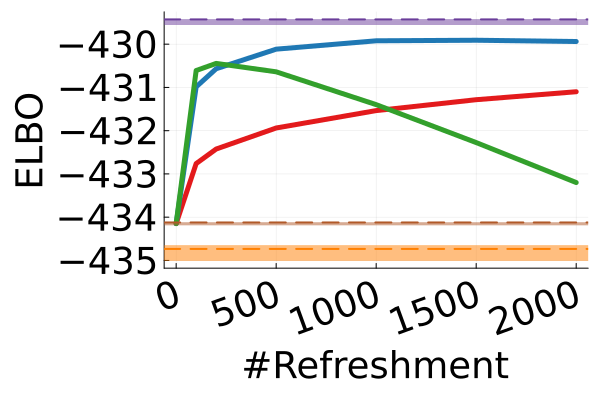}}
    \scalebox{1}{\includegraphics[width=\columnwidth]{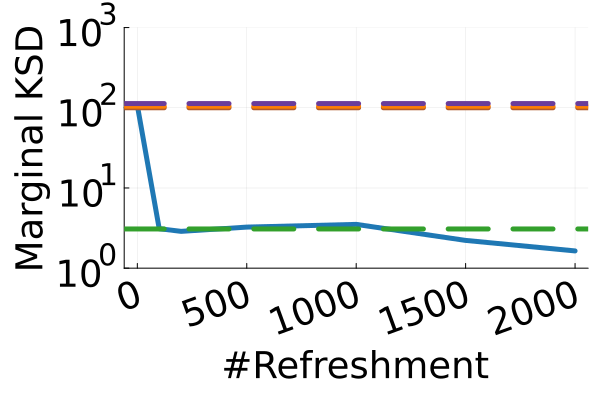}}
    \caption{(a) linear\\regression \label{fig:linreg_elbo}}
\end{subfigure}
\hfill
\centering 
\begin{subfigure}[b]{.26\columnwidth} 
    \scalebox{1}{\includegraphics[width=\columnwidth]{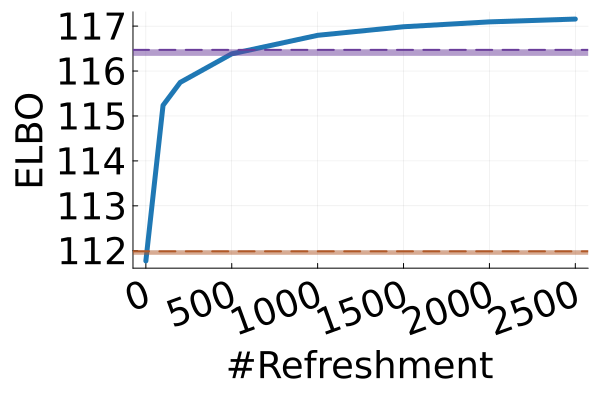}}
    \scalebox{1}{\includegraphics[width=\columnwidth]{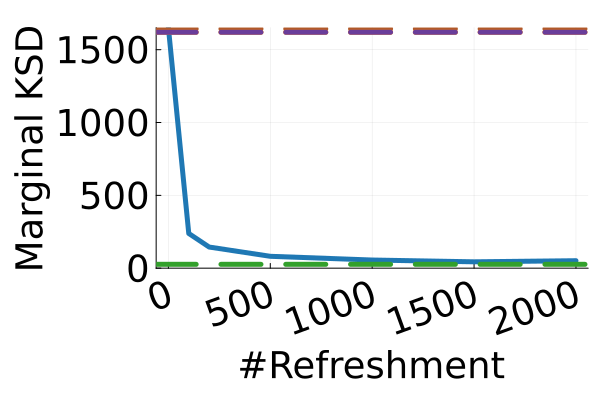}}
    \caption{(b) linear regression (heavy) \label{fig:lrh_elbo}}
\end{subfigure}
\hfill
\centering 
\begin{subfigure}[b]{.26\columnwidth} 
    \scalebox{1}{\includegraphics[width=\columnwidth]{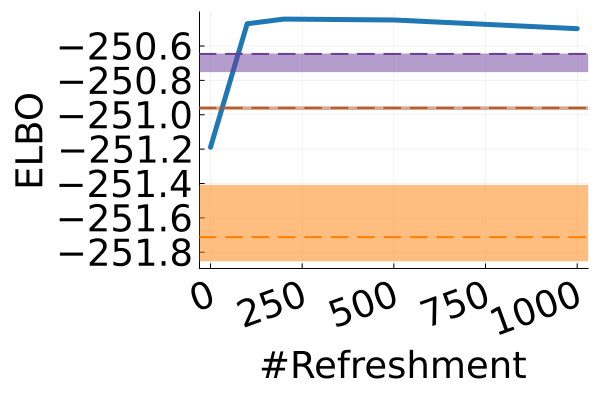}}
    \scalebox{1}{\includegraphics[width=\columnwidth]{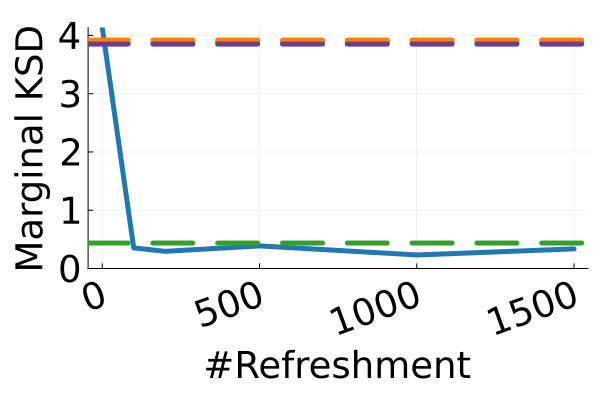}}
    \caption{(c) logistic\\ regression \label{fig:logreg_elbo}}
\end{subfigure}
\hfill
\centering
\begin{subfigure}[b]{.26\columnwidth} 
    \scalebox{1}{\includegraphics[width=\columnwidth]{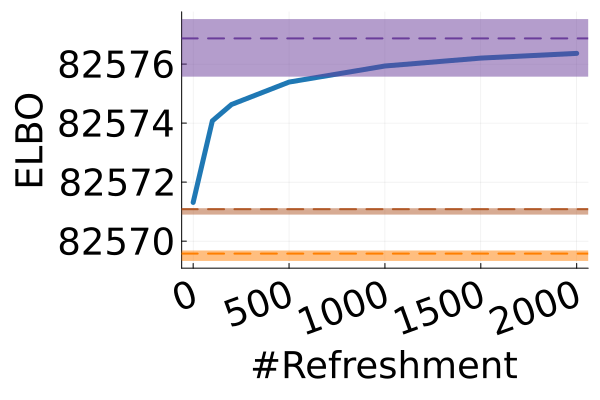}}
    \scalebox{1}{\includegraphics[width=\columnwidth]{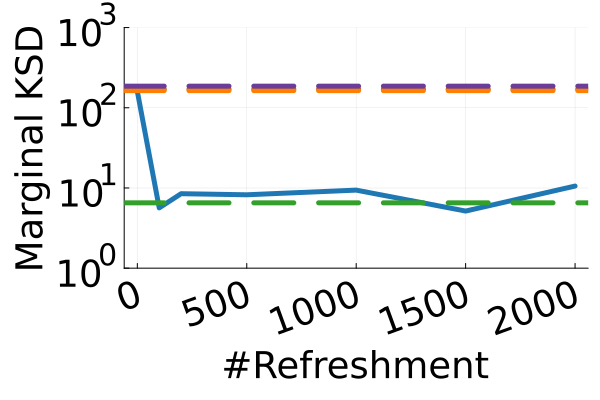}}
    \caption{(d) Poisson\\ regression \label{fig:poiss_elbo}}
\end{subfigure}
\hfill
\centering
\begin{subfigure}[b]{.26\columnwidth} 
    \scalebox{1}{\includegraphics[width=\columnwidth]{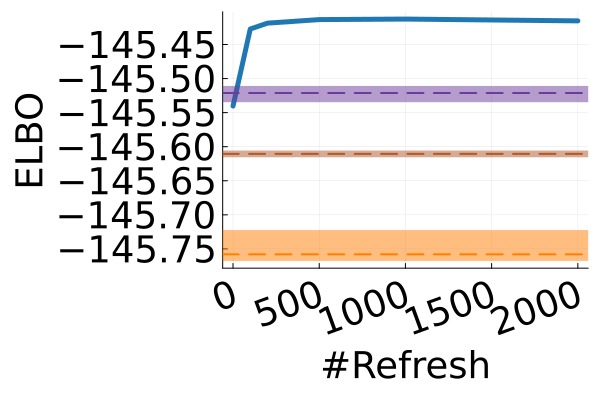}}
    \scalebox{1}{\includegraphics[width=\columnwidth]{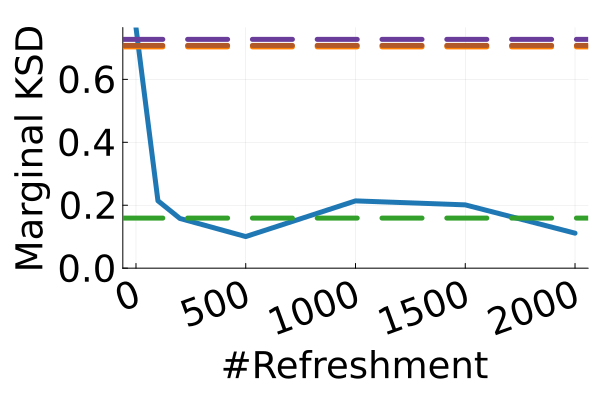}}
    \caption{(e) student t\\ regression \label{fig:tr_elbo}}
\end{subfigure}
\hfill
\centering
\begin{subfigure}[b]{.26\columnwidth} 
    \scalebox{1}{\includegraphics[width=\columnwidth]{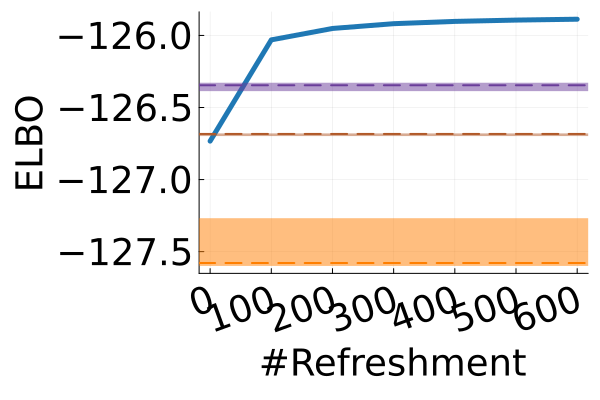}}
    \scalebox{1}{\includegraphics[width=\columnwidth]{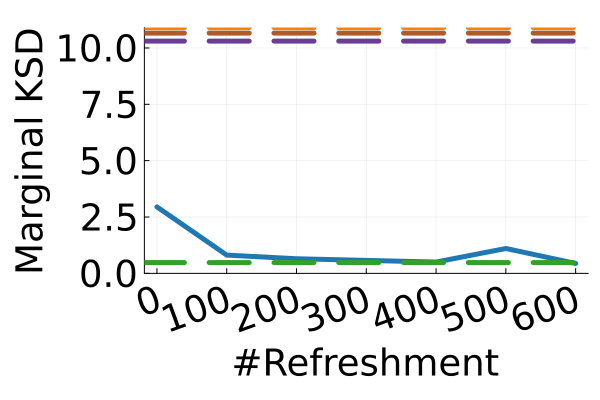}}
    \caption{(f) sparse\\ regression \label{fig:sp_elbo}}
\end{subfigure}
\hfill
\centering
\begin{subfigure}[b]{.26\columnwidth} 
    \scalebox{1}{\includegraphics[width=\columnwidth]{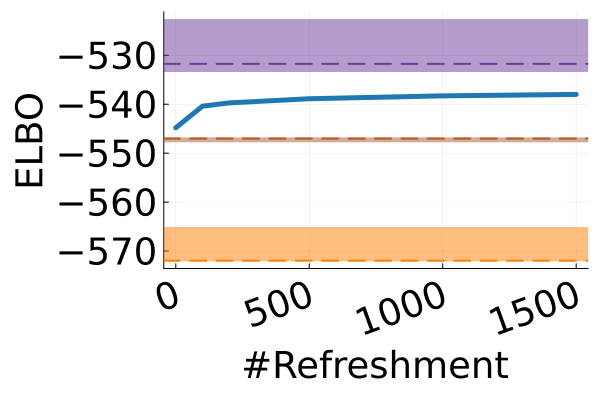}}
    \scalebox{1}{\includegraphics[width=\columnwidth]{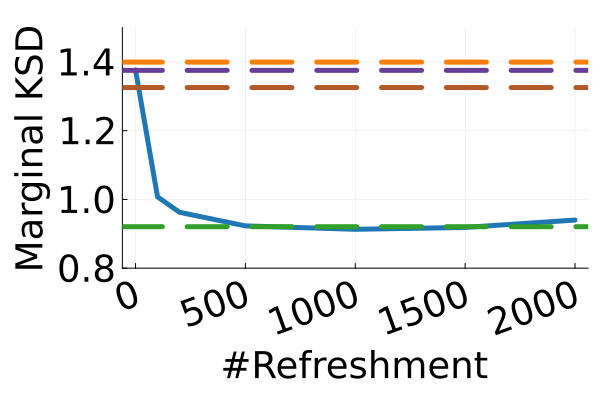}}
    \caption{(g) sparse regression (high dim) \label{fig:sp_big_elbo}}
\end{subfigure}
\hfill

\caption{ELBO and KSD comparison for real data examples. 
\cref{fig:linreg_elbo} displays the effect of step size for the linear regression problem: $\epsilon_1 = 0.0001, \epsilon_2
= 0.001$ and tuned $\epsilon = 0.0005$; see \cref{fig:real_elbo_tune} for step sizes for all other experiments.
Lines indicate the median, and error regions indicate $25^{th}$ to $75^{th}$ percentile from $5$ runs. 
\cref{fig:lrh_elbo} does not include ELBOs of \texttt{PlanarFlow} as values are
significantly worse than all other methods and are hard to visualize (see its ELBOs in \cref{fig:lrh_elbo_full}).}
\label{fig:real_metric}
\end{figure*}

Next, we provide a quantitative comparison of \texttt{MixFlow}, \texttt{NF}s, 
\texttt{NUTS} on 7 real data experiments outlined in \cref{apdx:realexpt}.
We tune each \texttt{NF} method under various settings
(\cref{tab:realnvp,tab:planar,tab:radial}), and present the best one for each
example.  ELBOs of \texttt{MixFlow} are estimated with \cref{alg:betterelbo},
averaging over $1{,}000$ independent trajectories. ELBOs of \texttt{NF}s are
based on $2{,}000$ samples.  To obtain an assessment for the target marginal
distribution itself (not the augmented target), we also compare methods using
the kernel Stein discrepancy (KSD) with an inverse quadratic (IMQ) kernel
\citep{Gorham17}.  \texttt{NUTS} and \texttt{NF}s use $5{,}000$ samples
for KSD estimation, while \texttt{MixFlow} is based on $2{,}000$ \iid draws
(\cref{alg:draw}). For KSD comparisons, all variational methods are tuned 
by maximizing the ELBO (\cref{fig:real_metric,fig:real_elbo_tune}).

\paragraph{Augmented target distribution}
\cref{fig:real_metric} displays the ELBO comparison.
First, \cref{fig:linreg_elbo} shows how
different step sizes $\epsilon$ and number of refreshments $N$ affects approximation quality. 
Overly large step sizes typically result in errors due to the use of discretized
Hamiltonian dynamics, while overly small step sizes result in a flow that makes slow
progress towards the target. \texttt{MixFlow} with a tuned step size 
generally shows a comparable joint ELBO value to the best
\texttt{NF} method, yielding a competitive target approximation. 
Similar comparisons and assessment of the effect of step size for the synthetic examples are presented in
\cref{fig:syn_2d,fig:higher_dim_metrics,fig:2d_more_nf}.
Note that in three examples (\cref{fig:linreg_elbo,fig:poiss_elbo,fig:sp_big_elbo}), the
tuned \texttt{RealNVP} ELBO exceeds \texttt{MixFlow} by a small amount; but this required expensive
architecture search and parameter tuning, and as we will describe next, \texttt{MixFlow} is
actually more reliable in terms of target marginal sample quality and density estimation.

\begin{figure}[t!]
\centering
\begin{subfigure}[b]{.49\columnwidth} 
    \scalebox{1}{\includegraphics[width=\columnwidth]{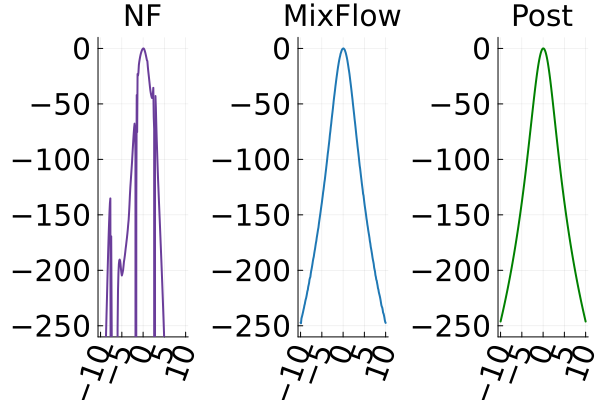}}
    \caption{(a) student t regression \label{fig:tr_lpdf}}
\end{subfigure}
\hfill
\centering
\begin{subfigure}[b]{.49\columnwidth} 
    \scalebox{1}{\includegraphics[width=\columnwidth]{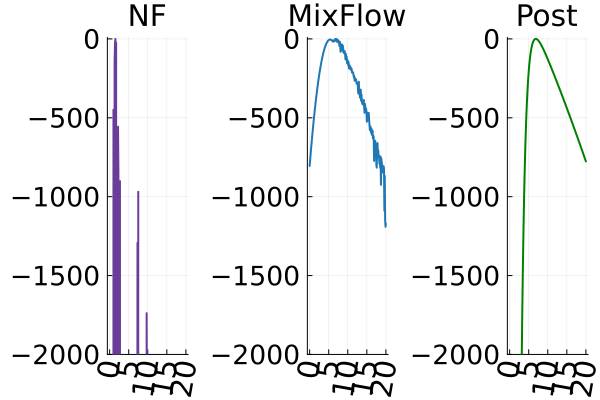}}
    \caption{(b) sparse regression (high dim) \label{fig:sp_big_lpdf}}
\end{subfigure}
\hfill

\caption{Sliced log conditional densities on
student-t regression (\cref{fig:tr_lpdf}) and high-dimensional sparse
regression (\cref{fig:sp_big_lpdf}). 
We visualize the log conditional density of the first coordinate by fixing 
other coordinates to value $0$. The \texttt{NF} methods are chosen to be the best performing 
ones from \cref{fig:tr_elbo,fig:sp_big_elbo}.
Since we only know the log posterior density up to an unknown
constant, we shift all lines to have maximal value $0$ for 
visualization.}
\label{fig:tr_metric}
\end{figure}

\paragraph{Original target distribution}
The second row of \cref{fig:real_metric} displays a comparison of KSD for the target distribution itself 
(instead of the augmented target). In particular, \texttt{MixFlow} produces 
comparable sample quality to that from \texttt{NUTS}---an exact MCMC method---and clearly outperforms
all of the \texttt{NF} competitors. The scatter plots of samples in \cref{fig:post_vis} confirm the
improvement in sample quality of \texttt{MixFlow} over variational competitors.
Further, \cref{fig:tr_metric} shows the (sliced) densities 
on two difficult real data examples: Bayesian student-t regression
(with a heavy-tailed posterior), and a high-dimensional sparse regression
(parameter dimension is $84$). This result demonstrates that the densities provided by \texttt{MixFlow} more
closely match those of the original target distribution than those of the best \texttt{NF}.  
Notice that \texttt{MixFlow} accurately
captures the skew and heavy tails of the exact target, while the \texttt{NF} density fails to
do so and contains many spurious modes. 

\subsection{Ease of tuning}
\begin{figure}[t!]
\centering 
\begin{subfigure}[b]{.49\columnwidth} 
    \scalebox{1}{\includegraphics[width=\columnwidth]{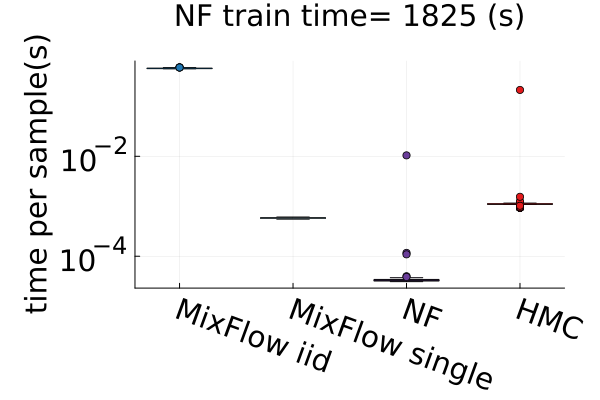}}
    \scalebox{1}{\includegraphics[width=\columnwidth]{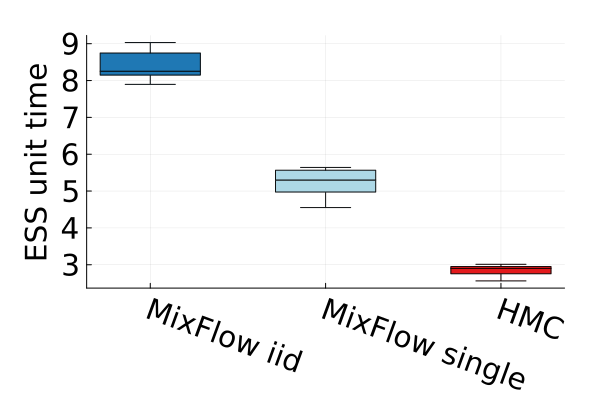}}
    \caption{(a) sparse regression (high dim)\label{fig:sp_big_time}}
\end{subfigure}
\hfill
\centering 
\begin{subfigure}[b]{.49\columnwidth} 
    \scalebox{1}{\includegraphics[width=\columnwidth]{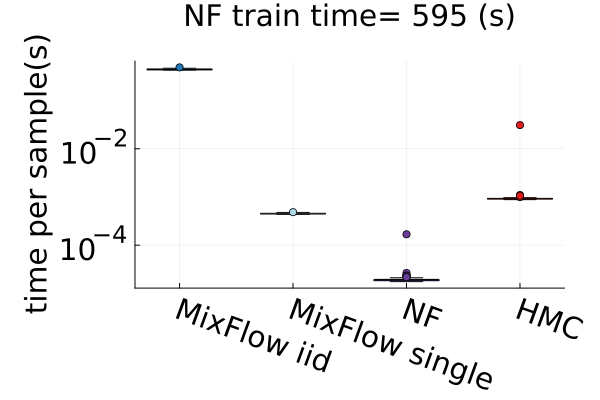}}
    \scalebox{1}{\includegraphics[width=\columnwidth]{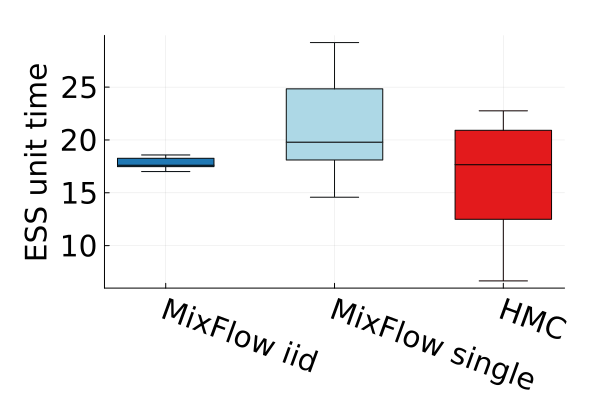}}
    \caption{(b) student t regression \label{fig:treg_time}}
\end{subfigure}
\hfill

\caption{Timing results (100 trials), showing sampling time (first row) and ESS per second (second row).}
\label{fig:real_timing}
\end{figure}

In order to tune \texttt{MixFlow}, we simply run a 1-dimensional 
parameter sweep for the step size $\epsilon$, and use a visual inspection of
the ELBO to set an appropriate number of flow steps $N$.
Tuning an \texttt{NF} requires optimizing its architecture,
number of layers, and its (typically many) parameters. 
Not only is this time consuming---in our experiments, 
tuning took 10 minutes to roughly 1 hour (\cref{fig:real_timing})---but 
the optimization can also behave in
unintuitive ways. For example, performance can be
heavily dependent on the number of flow layers, and
adding more layers does not necessarily improve quality. 
\cref{fig:2d_more_nf,tab:realnvp,tab:planar,tab:radial} show that using 
more layers does not necessarily help, and slows tuning considerably.
In the case of \texttt{RealNVP} specifically, tuning can be unstable,
especially for more complex models.
The optimizer often returns NaN values for flow parameters during training (see
\cref{tab:realnvp}). 
This instability has been noted in earlier work \citep[Sec.~3.7]{Dinh17}. 

\captionsetup[subfigure]{labelformat=empty}

\captionsetup[subfigure]{labelformat=empty}

\subsection{Time efficiency}
Finally, \cref{fig:real_timing} presents timing results for two of 
the real data experiments (additional comparisons in \cref{fig:2d_timing,fig:real_more_timing}).
In this figure, we use \texttt{MixFlow iid} to refer to \iid sampling from \texttt{MixFlow},
and \texttt{MixFlow single} to refer to collecting all intermediate states on a single trajectory. 
This result shows that the per sample time of \texttt{MixFlow single} is similar to \texttt{NUTS} and
\texttt{HMC} as one would expect.
\texttt{MixFlow iid} is the slowest because each
sample is generated by passing through the entire flow. The \texttt{NF} generates the fastest draws, but
recall that this comes at the cost of significant initial tuning time; 
in the time it takes \texttt{NF} to generate its first sample,
\texttt{MixFlow single} has generated millions of samples in \cref{fig:real_timing}. 
See \cref{apdx:2d_more} for a detailed discussion of this trade-off.

\cref{fig:real_timing,fig:real_more_timing} further show the computational
efficiency in terms of ESS per second on real data examples, which reflects the
autocorrelation between drawn samples. 
Results show that \texttt{MixFlow} produce comparable ESS per second to \texttt{HMC}.
\texttt{MixFlow single} behaves similarly to \texttt{HMC} as expected since the
pseudo-momentum refreshment we proposed (steps (2-3) of \cref{sec:hamiltonianmixflows})
resembles the momentum resample step of \texttt{HMC}. 
The ESS efficiency of \texttt{MixFlow iid} depends on the trade-off between a slower sampling time
and \iid nature of drawn samples. In these real data examples, \texttt{MixFlow iid}
typically produces a high ESS per second; but in the synthetic examples
(\cref{fig:2d_ess}), \texttt{MixFlow iid} is similar to the others.

%% file: conclusion.tex
\section{Conclusion}\label{sec:conclusion}
This work presented MixFlows, a new variational family
constructed from mixtures of pushforward maps. 
Experiments demonstrate a comparable sample quality to NUTS and more reliable posterior
approximations than standard normalizing flows.
A main limitation of our methodology is numerical stability; reversing the flow for long trajectories
can be unstable in practice.
Future work includes developing more stable momentum refreshment schemes
and extensions via involutive MCMC \citep{Neklyudov20,Spanbauer20,Neklyudov22}.

%% file: proofs.tex
\section{Proofs}\label{sec:proofs}

\bprfof{\cref{prop:smallervar}}
Because both estimates are unbiased, it suffices to show that
$\EE[f^2(X)] \geq \EE\left[\left( \frac{1}{N} \sum_{n =0}^{N-1} f(T^n X_0)\right)^2 \right]$,
which itself follows by Jensen's inequality: 
\[
\EE\left[\left( \frac{1}{N} \sum_{n =0}^{N-1} f(T^n X_0)\right)^2 \right]
\leq \EE\left[\frac{1}{N} \sum_{n =0}^{N-1} f^2(T^n X_0) \right]  = \EE[f^2(X)]. 
\]
\eprfof

\bprfof{\cref{thm:weakconvergence}}
Since setwise convergence implies weak convergence, we will focus on proving setwise convergence.
We have that $q_{\lambda,N}$ converges setwise to $\pi$ if and only if for all measurable bounded $f : \mcX \to \reals$,
\[
\EE f(X_N) \to \EE f(X), \qquad X_N \distas q_{\lambda,N}, \quad X \distas \pi.
\]
The proof proceeds by directly analyzing $\EE f(X_N)$:
\[
\EE f(X_N) &= \int f(x) q_{\lambda,N}(\dee x)\\
&= \frac{1}{N}\sum_{n=0}^{N-1}\int f(x) (T^n_\lambda q_0)(\dee x)\\
&= \frac{1}{N}\sum_{n=0}^{N-1}\int f(T^n_\lambda x) q_0(\dee x)\\
&= \int \frac{1}{N}\sum_{n=0}^{N-1}f(T^n_\lambda x) q_0(\dee x).
\]
Since $q_0 \ll \pi$, by the Radon-Nikodym theorem, there exists a density of $q_0$ with respect to $\pi$, so
\[
\EE f(X_N) &= \int \frac{1}{N}\sum_{n=0}^{N-1}f(T^n_\lambda x) \der{q_0}{\pi}(x) \pi(\dee x).
\]
By the pointwise ergodic theorem (\cref{thm:pointwiseergodic}), $f_N(x) = \frac{1}{N}\sum_{n=0}^{N-1}f(T^n_\lambda x)$ converges
pointwise $\pi$-\aev to $\int f\dee \pi$; and because $f$ is bounded, $f_N$ is uniformly bounded for all $N\in\nats$. Hence by the Lebesgue
dominated convergence theorem,
\[
\lim_{N\to\infty}\EE f(X_N) &= \lim_{N\to\infty}\int \frac{1}{N}\sum_{n=0}^{N-1}f(T^n_\lambda x) \der{q_0}{\pi}(x) \pi(\dee x)\\
&= \int \left(\int f\dee \pi\right)\der{q_0}{\pi}(x) \pi(\dee x)\\
&= \left(\int f\dee \pi\right) \cdot 1 = \EE f(X).
\]
\eprfof
\bthm[{Mean ergodic theorem in Banach spaces [\citealp{Yosida38,Kakutani38,Riesz38}; \citealp[Theorem 8.5]{Eisner15}]}]\label{thm:meanergodic}
Let $T$ be a bounded linear operator on a Banach space $E$, define the operator 
\[
A_N = \frac{1}{N}\sum_{n=0}^{N-1} T^n,
\]
and let
\[
\mathrm{fix}(T) = \left\{v \in E : Tv = v\right\} = \ker(I-T).
\]
Suppose that $\sup_{N\in\nats}\|A_N\| < \infty$ and that $\frac{1}{N} T^N v \to 0$ for all $v\in E$.
Then the subspace
\[
V = \left\{v \in E : \lim_{N\to\infty} A_N v \text{ exists}\right\}
\]
is closed, $T$-invariant, and decomposes into a direct sum of closed subspaces
\[
V = \mathrm{fix}(T) \oplus \overline{\mathrm{range}}(I-T).
\]
The operator $\left. T\right|_V$ on $V$ is mean ergodic. Furthermore, the operator
\[
A : V \to \mathrm{fix}(T) \qquad A v = \lim_{N\to\infty} A_N v
\]
is a bounded projection with kernel $\ker(A) = \overline{\mathrm{range}}(I-T)$ and $AT = A = TA$.
\ethm

\bprfof{\cref{thm:tvconvergence}}
We suppress $\lambda$ subscripts for brevity.
Consider the Banach space $\mcM(\pi)$ of signed finite measures dominated by $\pi$ 
endowed with the total variation norm $\|m\| = \sup_{A\subseteq \mcX} m(A) - \inf_{A\subseteq\mcX} m(A)$.
Then the pushforward $Tm$ of $m \in \mcM(\pi)$ under $T$ is dominated by $\pi$
since
\[
\pi(A) = 0 \implies \pi(T^{-1}(A)) = 0 \implies m(T^{-1}(A)) = 0.
\]
Note the slight abuse of notation involving the same symbol for $T : \mcX\to\mcX$ and the associated operator.
Hence $T:\mcM(\pi) \to \mcM(\pi)$ is a linear operator on $\mcM(\pi)$.
Further,
\[
\|Tm\| = \sup_{A\subseteq \mcX} m(T^{-1}(A)) - \inf_{A\subseteq\mcX} m(T^{-1}(A)) \leq \sup_{A\subseteq \mcX} m(A)- \inf_{A\subseteq\mcX} m(A) = \|m\|,
\]
and so $T$ is bounded with $\|T\| \leq 1$. Hence $\frac{1}{N}T^N m \to 0$ for all $m\in \mcM(\pi)$,
and if we define the operator
\[
\forall N\in\nats, \quad A_N = \frac{1}{N}\sum_{n=0}^{N-1} T^n,
\]
we have that $\sup_{N\in\nats}\|A_N\| \leq 1$. Therefore by
the mean ergodic theorem in Banach spaces [\citealp{Yosida38,Kakutani38,Riesz38}; \citealp[Theorem 8.5]{Eisner15}],
we have that 
\[
A : V \to \mathrm{fix}(T) \quad A m = \lim_{N\to\infty} A_N m
\]
where $V = \left\{m \in \mcM(\pi) : \lim_{N\to\infty} A_N m \text{ exists}\right\}$.
\citet[Theorem 8.20]{Eisner15} guarantees that $V = \mcM(\pi)$ as long as 
the weak limit of $A_N m$ exists for each $m \in \mcM(\pi)$.
Note that since $\mcM(\pi)$ and $L^1(\pi)$ are isometric (via the map $m \mapsto \der{m}{\pi}$),
and $\pi$ is $\sigma$-finite, the dual of $\mcM(\pi)$ 
is the set of linear functionals
\[
m \mapsto \int \phi(x) m(\dee x), \qquad \phi \in L^\infty(\pi).
\]
So therefore we have that $V = \mcM(\pi)$ if for each $m\in \mcM(\pi)$,
there exists a $g \in \mcM(\pi)$ such that
\[
\forall \phi \in L^\infty(\pi), \qquad \int \phi(x) (A_N m)(\dee x) \to \int \phi(x) g(\dee x).
\]
The same technique using a transformation of variables and the pointwise
ergodic theorem from the proof of \cref{thm:weakconvergence} provides the desired weak convergence.
Therefore we have that $A_N m$ converges in total variation to $\mathrm{fix}(T)$ for all $m\in\mcM(\pi)$,
and hence $A_N q_0$ converges in total variation to $\mathrm{fix}(T)$.
Furthermore \cref{thm:weakconvergence} guarantees weak convergence of $A_N q_0$ to $\pi$,
and $\pi\in\mathrm{fix}(T)$, so thus $\tvd{q_{\lambda,N}}{\pi} \to 0$.
\eprfof

\bprfof{\cref{thm:approxtvconvergence}}
We suppress $\lambda$ subscripts for brevity.
By the triangle inequality,
\[
\tvd{\hq_N}{\pi}
&\leq 
\tvd{\hq_N}{q_N}
+
\tvd{q_N}{\pi}.
\]
Focusing on the error term, we have that
\[
\tvd{\hq_N}{q_N}
&=\tvd{\frac{1}{N}\sum_{n=1}^{N-1}\hT^n q_0}{\frac{1}{N}\sum_{n=1}^{N-1}T^n q_0}\\
&=\frac{N-1}{N}\tvd{\frac{1}{N-1}\sum_{n=1}^{N-1}\hT^n q_0}{\frac{1}{N-1}\sum_{n=1}^{N-1}T^n q_0}\\
&=\frac{N-1}{N}\tvd{\hT\frac{1}{N-1}\sum_{n=1}^{N-1}\hT^{n-1} q_0}{T\frac{1}{N-1}\sum_{n=1}^{N-1}T^{n-1} q_0}\\
&=\frac{N-1}{N}\tvd{\hT\frac{1}{N-1}\sum_{n=0}^{N-2}\hT^{n} q_0}{T\frac{1}{N-1}\sum_{n=0}^{N-2}T^{n} q_0}\\
&=\frac{N-1}{N}\tvd{\hT \hq_{N-1}}{T q_{N-1}}\\
&=\frac{N-1}{N}\tvd{\hq_{N-1}}{\hT^{-1}T q_{N-1}},
\]
where the last equality is due to the fact that $\hT$ is a bijection.
The triangle inequality yields
\[
\tvd{\hq_N}{q_N}
&\leq \frac{N-1}{N}\left(\tvd{\hq_{N-1}}{q_{N-1}} + \tvd{q_{N-1}}{\hT^{-1}T q_{N-1}}\right)\\ 
&= \frac{N-1}{N}\left(\tvd{\hq_{N-1}}{q_{N-1}} + \tvd{\hT q_{N-1}}{T q_{N-1}}\right).
\]
Then iterating that technique yields 
\[ \label{eq:approxtverr}
\tvd{\hq_N}{q_N}
\leq \sum_{n = 1}^{N-1} \frac{n}{N} \tvd{\hT q_n}{T q_n},
\]
which completes the proof.
\eprfof

\bprfof{\cref{cor:approxtvconvergence}}
Examining the total variation in its $L^1$ distance formulation yields that for $n = 1, \dots, N-1$,
\[
    \tvd{\hT q_n}{T q_n} 
    &=  \int \left|\frac{\hT q_n(x)}{T q_n(x)} -1\right| Tq_n(\dee x) \\ 
    &= \int \left|\exp\left\{ \log q_n(\hT^{-1} x) - \log q_n(T^{-1}x) 
    + \log J(T^{-1}x)-\log \hJ(\hT^{-1}x)\right\} -1 \right|  Tq_n(\dee x) .
\]
By assumption,
\[
\left|\log q_n(\hT^{-1}x) - \log q_n(T^{-1}x)\right| \leq \ell\left\|\hT^{-1}x - T^{-1}x\right\| \leq \ell\epsilon,
\]
and
\[
\left|\log J(T^{-1}x)-\log \hJ(\hT^{-1}x)\right| &\leq
\left|\log J(T^{-1}x)-\log J(\hT^{-1}x)\right| + \left|\log J(\hT^{-1}x) - \log \hJ(\hT^{-1}x)\right|\\
&\leq \ell\left\|\hT^{-1}x - T^{-1}x\right\| + \epsilon\\
&\leq (\ell+1)\epsilon.
\]
Combining these two bounds yields
\[
    \tvd{\hT q_n}{T q_n} 
    &\leq \exp((2\ell+1)\epsilon) - 1 \leq (2\ell+1)\epsilon\exp((2\ell+1)\epsilon).
\]
Therefore, by \cref{eq:approxtverr}, we obtain that
\[
\tvd{\hq_N}{q_N}
\leq \sum_{n = 1}^{N-1} \frac{n}{N} \tvd{\hT q_n}{T q_n} \leq
\frac{N-1}{2}e^{(2\ell+1)\eps} \cdot (2\ell+1)\eps \leq N\epsilon (\ell+1) e^{(2\ell+1)\epsilon}.
\]
Finally, combining the above with \cref{thm:approxtvconvergence} yields the
desired result.
\eprfof

%% file: apdx_hamflow.tex
\section{Memory efficient ELBO estimation} \label{sec:efficient_elbo}

\begin{algorithm}[tb]
\caption{$\texttt{DensityTriple}(x)$: Evaluate $T_\lambda^{-N+1}(x)$, $q_{\lambda,N}(x)$, and $\prod_{j=1}^{N-1} J_\lambda(T_\lambda^{-j}x) $ with $O(N)$-time, $O(1)$-memory}\label{alg:densityandmore}
\begin{algorithmic}
\STATE{\bfseries Input:} location $x$, reference distribution $q_0$, flow map $T_\lambda$, Jacobian $J_\lambda$, number of steps $N$
\STATE $L \gets 0$
\STATE $w \gets q_0(x)$
\FOR{$n=1, \dots, N-1$}
\STATE $x \gets T^{-1}_\lambda(x)$
\STATE $L \gets L + \log J_\lambda(x)$
\STATE $w \gets \texttt{LogSumExp}(w, \log q_0(x) - L)$
\ENDFOR
\STATE $w \gets w - \log N$
\STATE{ \bfseries Return:} $x$, $\exp(w)$, $\exp(L)$
\end{algorithmic}
\end{algorithm}

\begin{algorithm}[tb]
\caption{$\texttt{EstELBO}(\lambda,N)$: Estimate the ELBO for $q_{\lambda,N}$ in $O(N)$-time, $O(1)$-memory. } \label{fig:efficientelbo}
\begin{algorithmic}
\STATE{\bfseries Input:} reference $q_0$, unnormalized target $p$, flow map $T_\lambda$, Jacobian $J_\lambda$, number of flow steps $N$
\STATE $x \gets \texttt{Sample}(q_0)$
\STATE $x', z, J \gets \texttt{DensityTriple}(x)$
\STATE $f \gets \log p(x)$
\STATE $g \gets \log z$
\FOR{$n=1,\dots,N-1$}
\STATE $\bq \gets \frac{1}{N} q_0(x')/J $
\STATE $J_{n-1} \gets J_{\lambda}(x)$
\STATE $ z \gets \left(z - \bq\right)/J_{n-1}$ 
\STATE $ x \gets T_\lambda(x) $
\STATE $ z \gets z + \frac{1}{N}q_0(x)$
\STATE $ f \gets f + \log p(x)$
\STATE $g \gets g + \log z$
\IF{$n < N-1$}
\STATE $J \gets J \cdot J_{n-1}/ J_\lambda(x')$
\ENDIF
\STATE $ x' \gets T_\lambda(x') $
\ENDFOR
\STATE{\bfseries Return:} $\widehat{\mathrm{ELBO}}(\lambda,N) \gets \frac{1}{N} (f - g)$
\STATE $\widehat{\mathrm{ELBO}}(\lambda,N)$
\end{algorithmic}
\end{algorithm}

\section{Hamiltonian flow pseudocode}\label{sec:hamflow_code}

\begin{algorithm}[tb]
\caption{Compute $T_{\lambda,\epsilon}$ and its Jacobian $J_{\lambda,\epsilon}$ for Hamiltonian flow with leapfrog integrator}\label{alg:hamflow}
\begin{algorithmic}
\STATE{\bfseries Input:} initial state $x\in\mcX$, $\rho \in \reals^d$, $u\in[0,1]$, step size $\epsilon$, shift $\xi\in\reals$, pseudorandom shift $z(\cdot, \cdot)$ 
\STATE $x_0, \rho_0 \gets x, \rho$
\FOR{$\ell=1,\dots,L$}
\STATE $x_\ell, \rho_\ell \gets \hH_\eps(x_{\ell-1}, \rho_{\ell-1})$
\ENDFOR
\STATE $x', \rho' \gets x_L, \rho_L$
\STATE $u' \gets u + \xi \mod 1$
\FOR{$i=1,\dots,d$}
\STATE $\rho''_i \gets R^{-1}(R(\rho_i') + z(x_i', u') \mod 1)$
\ENDFOR
\STATE $J \gets m(\rho')/m(\rho'')$
\STATE{\bfseries Return:} $(x', \rho'', u'), J$
\end{algorithmic}
\end{algorithm}

%% file: apdx_extensions.tex
\section{Extensions}\label{sec:extensions}

\paragraph{Tunable reference}
So far we have assumed that the reference distribution $q_0$ for the flow is fixed. 
Given that $q_0$ is often quite far from the target $\pi$, this forces the variational flow
to spend some of its steps just moving the bulk of the mass to $\pi$.
But this can be accomplished much easier with, say, a simple linear transformation that efficiently
allows large global moves in mass.
For example, if $q_0 = \mathcal{N}(0,I)$, we can include a map $M_\theta$
\[
q_{\lambda,N} &= \frac{1}{N}\sum_{n=0}^{N-1} T_\lambda^n M_\theta q_0,
\]
where $M_\theta(x) = \theta_1 x + \theta_2$, where $\theta_1\in\reals^{d\times d}$ and $\theta_2\in\reals^d$.
Note that it is possible to optimize the reference and flow jointly, or to
optimize the reference distribution by itself first and then use that fixed
reference in the flow optimization.

\paragraph{Automated burn-in}
A common practice in MCMC is to throw away a first fraction of the states in the sequence to
ensure that the starting sample is in a high probability region of the target
distribution, thus reducing the bias from initialization (``burn-in''). 
Usually one needs a diagnostic to check when burn-in is completed.
In the case of \texttt{MixFlow}, we can monitor the burn-in phase in a
principled way by evaluating the ELBO.  Once the flow is trained, the
variational distribution with $M$ burn-in samples is simply
\[
q_{\lambda,M,N} &= \frac{1}{N-M}\sum_{n=M}^{N-1} T_\lambda^n q_0,
\]
We can easily optimize this by estimating the ELBOs for $M = 1, \dots, N$. 

\paragraph{Mixtures of MixFlows}
One can build multiple MixFlows starting from multiple different initial reference distributions; when
the posterior is multimodal, it may be the case that some of these MixFlows converge to different
modes but do not mix across modes. 
In this scenario, it can be helpful to average several MixFlows, i.e., build an approximation of the form
\[
q^\star_{\lambda, N} = \sum_{k = 1}^K w_k (q_{\lambda, N})_{k}, \quad \sum_{k = 1}^K w_k = 1.
\]
Because each component flow provides access to \iid samples and density evaluation,
\texttt{MixFlow} provides the ability to optimize the weights by maximizing the ELBO (i.e., minimizing the KL divergence).

%% file: apdx_expt.tex
\section{Additional experimental details} \label{apdx:expt_detail}
In this section, we provide details for each experiment presented in the main text,
as well as additional results regarding numerical stability and a high-dimensional synthetic experiment. 
Aside from the univariate synthetic example, all examples include a pseudotime shift step with
$\xi = \pi/16$, and a momentum refreshment with $z(x,u) = 0.5\sin(2x + u)+ 0.5$.
For the kernel Stein discrepancy, we use a 
IMQ kernel $k(x, y) = (c^2 + \|x - y\|_2^2)^\beta$ with
$\beta = -0.5, c = 1$, the same setting as in \citep{Gorham17}. 

For all experiments, unless otherwise stated, \texttt{NUTS} uses $20{,}000$ steps 
for adaptation, targeting at an average acceptance ratio $0.7$, and generates $5{,}000$ samples for KSD estimation. 
The KSD for \texttt{MixFlow} is estimated using $2{,}000$ samples. 
The KSD estimation for \texttt{NEO} is based on $5{,}000$ samples generated 
by a \emph{tuned} \texttt{NEO} after $20{,}000$ burn-in steps.
We adopted two tuning strategies for \texttt{NEO}: (1) choosing among the combinations of 
several fixed settings of discretization step ($ \epsilon = 0.2, 0.5, 1.0$), friction parameter $\gamma$ of nonequilibrium Hamiltonian dynamics ($\gamma = 0.2, 0.5, 1.0$), integration steps ($K = 10, 20$), and mass matrix of momentum distribution ($M = I$); 
(2) fixing the integration steps ($K = 10, 20$) and friction parameter ($\gamma  = 0.2, 0.5, 1.0$), and 
using windowed adaptation \cite{carpenter2017stan} to
adapt the mass matrix and integration step size, targeting at an average acceptance ratio at $0.7$.  
The optimal setting of \texttt{NEO} is considered to be the one that produces lowest average
marginal KSD over $3$ runs with no NaN values encountered. 
Performance of \texttt{NEO} across various settings for each example is summarized in \cref{tab:aggregated_neo}.
In each of \texttt{NEO} MCMC transition, we run $10$ deterministic orbits and computes
corresponding normalizing constant estimates \emph{in parallel}. 
Aside from \texttt{NEO}, all other methods are deployed using a single processor.

As for \texttt{NUTS}, we use the Julia package \texttt{AdvancedHMC.jl} 
\citep{xu2020advancedhmc} with all default settings. 
\texttt{NEO} adaptation is also conducted using  \texttt{AdvancedHMC.jl} with
the number of simulation steps set to the number of integration steps of \texttt{NEO}.
The ESS is computed using the R package \texttt{mcmcse}, which is based  
on batch mean estimation. 
We implement all \texttt{NF}s
using the Julia package \texttt{Bijectors.jl} \citep{fjelde2020bijectors}.
The flow layers of \texttt{PlanarFlow} and \texttt{RadialFlow}, as well as 
the coupling layers of \texttt{RealNVP} are implemented in \texttt{Bijectors.jl}. 
The affine coupling functions of \texttt{RealNVP}---a scaling function and a shifting function---are 
both parameterized using fully connected neural networks of three layers, of which 
the activation function is LeakyReLU and number of hidden units is by default the 
same dimension as the target distribution, unless otherwise stated. 

\subsection{Univariate synthetic examples} \label{apdx:onedim}

The three target distributions tested in this experiment were
\bitems
\item normal: $\distNorm(2, 2^2)$,
\item synthetic Gaussian mixture: $0.5\distNorm(-3, 1.5^2) +  0.3\distNorm(0, 0.8^2) + 0.2\distNorm(3, 0.8^2)$, and
\item Cauchy: $\distCauchy(0, 1)$.
\eitems

For all three examples, we use a momentum refreshment without
introducing the pseudotime variable; this enables us to plot the full joint density of $(x, \rho) \in \reals^2$:
\[
    \rho'' \gets R_\text{Lap}^{-1}(R_\text{Lap}(\rho') + (\sin(2x') + 1)/2  \mod 1).
\]
For the three examples, we used the leapfrog stepsize $\eps =0.05$ 
and run $L=50$ leapfrogs between each refreshment.  
For both the Gaussian and Gaussian mixture targets, we use $100$ refreshments. In the case of the Cauchy, 
 we used $1{,}000$ refreshments.        

\subsection{Multivariate synthetic examples} \label{apdx:2d}
The three target distributions tested in this experiment were
\bitems 
\item  the banana distribution \citep{haario2001adaptive}:
\[
y = \begin{bmatrix}y_1 \\ y_2\end{bmatrix} \distas 
\distNorm\left( 0, \begin{bmatrix}100 & 0 \\ 0 & 1\end{bmatrix} \right), \quad 
x = \begin{bmatrix} y_1 \\ y_2 + by_1^2 - 100b \end{bmatrix}, \quad b = 0.1;
\]
\item  Neals' funnel \citep{neal2003slice}:
\[
x_1 \distas \distNorm\left( 0, \sigma^2 \right), \quad 
x_2 \given x_1 \distas \distNorm\left( 0, \exp\left(\frac{x_1}{2}\right) \right), \quad \sigma^2 = 36;
\]
\item  a cross-shaped distribution: in particular, a Gaussian mixture of the form
\[
x &\distas \frac{1}{4}\distNorm\left( \begin{bmatrix} 0 \\ 2 \end{bmatrix}, 
				\begin{bmatrix} 0.15^2 & 0 \\ 0 & 1 \end{bmatrix} \right) + 
	\frac{1}{4}\distNorm\left( \begin{bmatrix} -2 \\ 0 \end{bmatrix}, 
				\begin{bmatrix} 1 & 0 \\ 0 & 0.15^2  \end{bmatrix} \right)\\ 
	&+ \frac{1}{4}\distNorm\left( \begin{bmatrix} 2 \\ 0 \end{bmatrix}, 
	\begin{bmatrix} 1 & 0 \\ 0 & 0.15^2  \end{bmatrix} \right) + 
	\frac{1}{4}\distNorm\left( \begin{bmatrix} 0 \\ -2 \end{bmatrix}, 
	\begin{bmatrix} 0.15^2 & 0 \\ 0 & 1 \end{bmatrix} \right);
\]
\item  and a warped Gaussian distribution 
\[
y = \begin{bmatrix}y_1 \\ y_2\end{bmatrix} \distas 
\distNorm\left( 0, \begin{bmatrix} 1 & 0 \\ 0 & 0.12^2 \end{bmatrix} \right), 
\quad x = \begin{bmatrix} \|y\|_2\cos\left( \atantwo\left( y_2, y_1 \right) - \frac{1}{2}\|y\|_2 \right) \\ 
\|y\|_2\sin\left( \atantwo\left( y_2, y_1 \right) - \frac{1}{2}\|y\|_2 \right) \end{bmatrix},
\]
where $\atantwo(y,x)$ is the angle, in radians, between the positive $x$ axis
and the ray to the point $(x,y)$.
\eitems

\begin{figure}[t!]
    \begin{subfigure}{\columnwidth}
		\includegraphics[width=0.24\columnwidth]{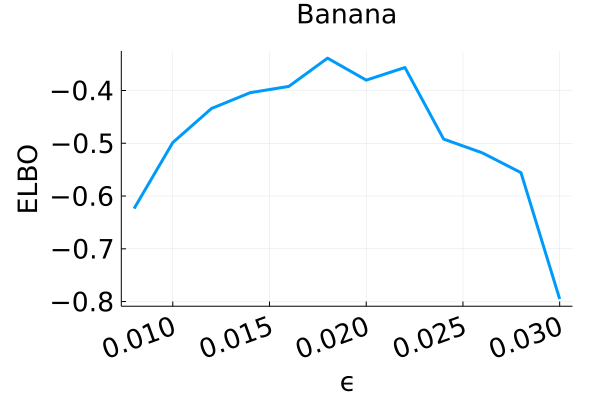}
		\includegraphics[width=0.24\columnwidth]{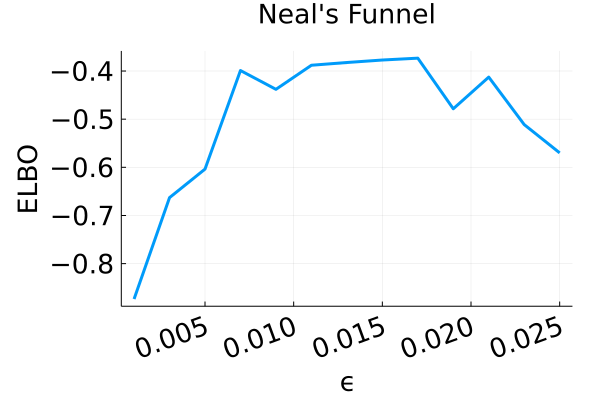}
		\includegraphics[width=0.24\columnwidth]{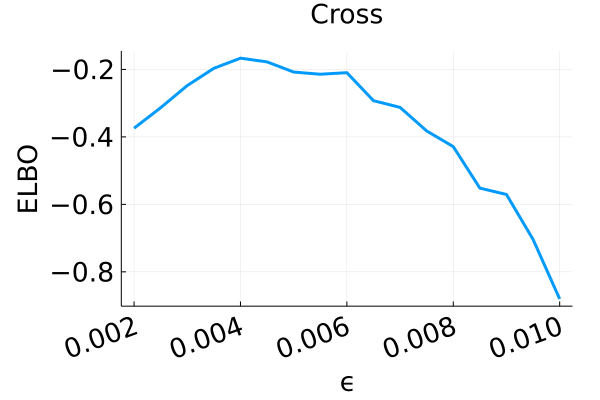}
		\includegraphics[width=0.24\columnwidth]{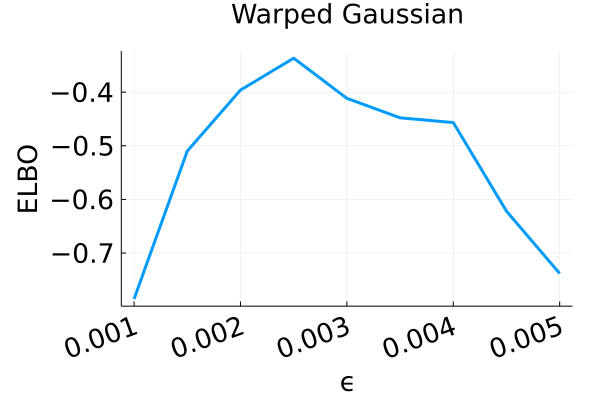}
		\caption{(a): ELBO versus step size $\epsilon$. }\label{fig:2d_tune_elbo}
	\end{subfigure}
    \begin{subfigure}{\columnwidth}
		\includegraphics[width=0.24\columnwidth]{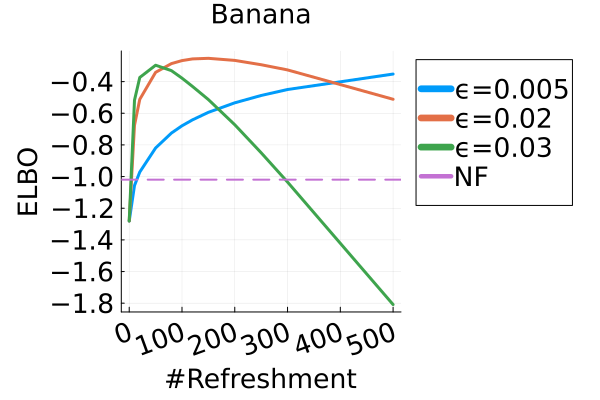}
		\includegraphics[width=0.24\columnwidth]{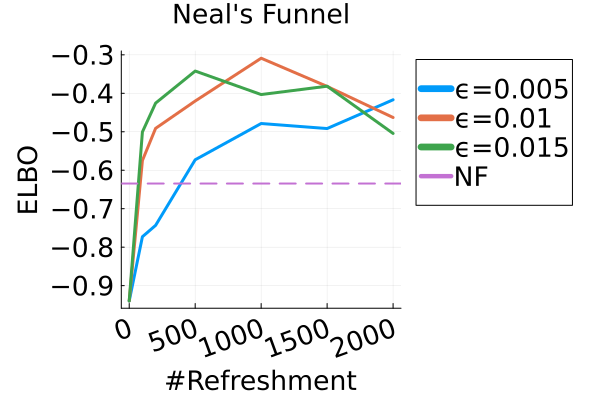}
		\includegraphics[width=0.24\columnwidth]{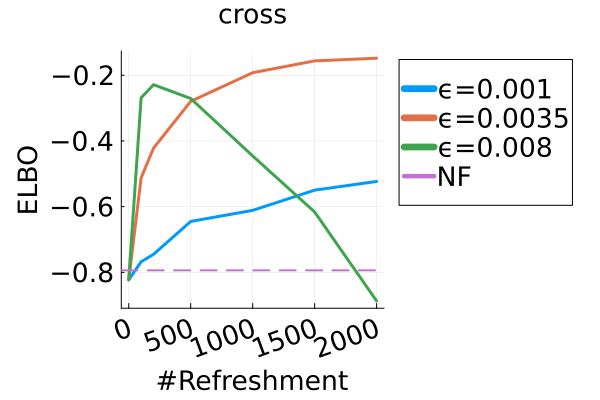}
		\includegraphics[width=0.24\columnwidth]{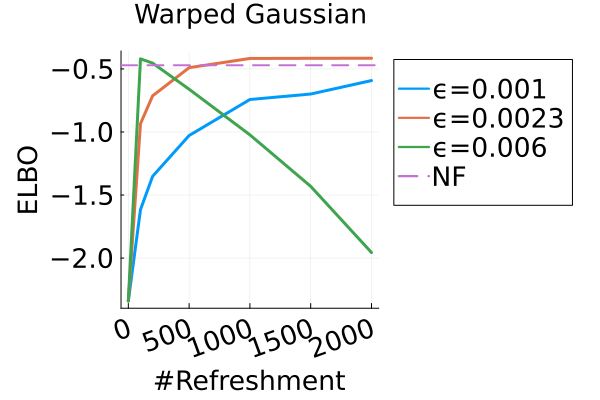}
		\caption{(b): ELBO of \texttt{MixFlow} versus number of refreshments $N$ over different step sizes, and comparision to tuned \texttt{PlanarFlow}}\label{fig:2d_elbo}
	\end{subfigure}

\caption{\texttt{MixFlow} tuning for the four multivariate synthetic examples}\label{fig:syn_2d}
\end{figure}

We used flows with $500$ and $2000$ refreshments for the banana distribution, Neal's Funnel respectively, and
a flow with $1000$ refreshments for both cross distribution and warped Gaussian.
Between each refreshment we used 200 leapfrog steps for the banana distribution, 60 for the cross distribution,
and 80 for the funnel and warped Gaussian. Note that in all four examples, we individually
tuned the step size $\eps$ by maximizing the estimated ELBO, as shown in \cref{fig:2d_tune_elbo}. 
\cref{fig:2d_elbo} also demonstrates how the ELBO varies versus the number of refreshments $N$.
For small step sizes $\epsilon$, the discretized Hamiltonian dynamics approximates the continuous dynamics,
and the ELBO generally increases with $N$ indicating convergence to the target. For larger step sizes $\epsilon$,
the ELBO increases to a peak and then decreases, indicating that the discretized dynamics do not exactly target
the desired distribution.

\begin{figure*}[t!]
    \centering 
\begin{subfigure}[b]{.24\textwidth} 
    \scalebox{1}{\includegraphics[width=\textwidth]{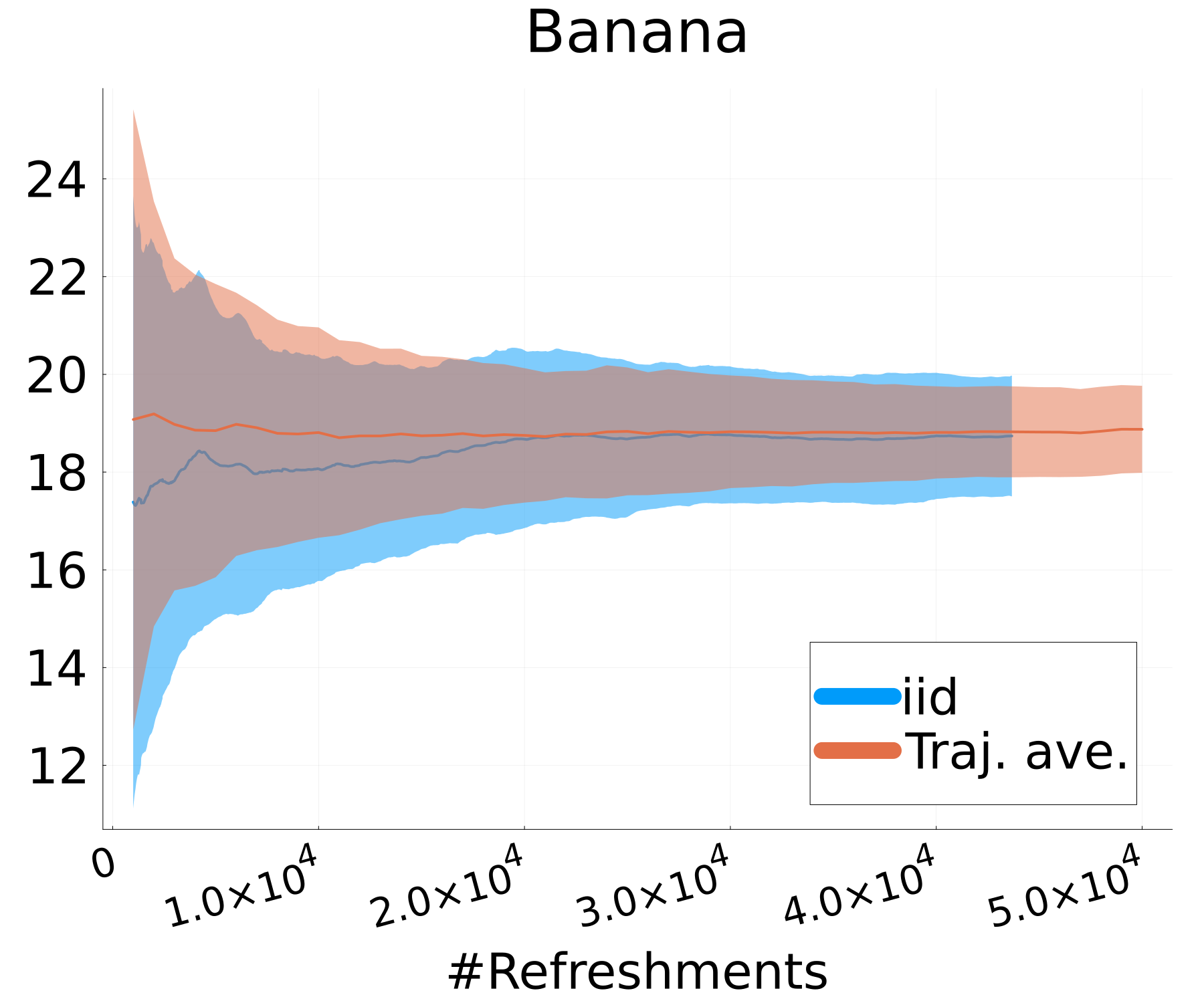}}
\end{subfigure}
\hfill
\centering 
\begin{subfigure}[b]{.24\textwidth} 
    \scalebox{1}{\includegraphics[width=\textwidth]{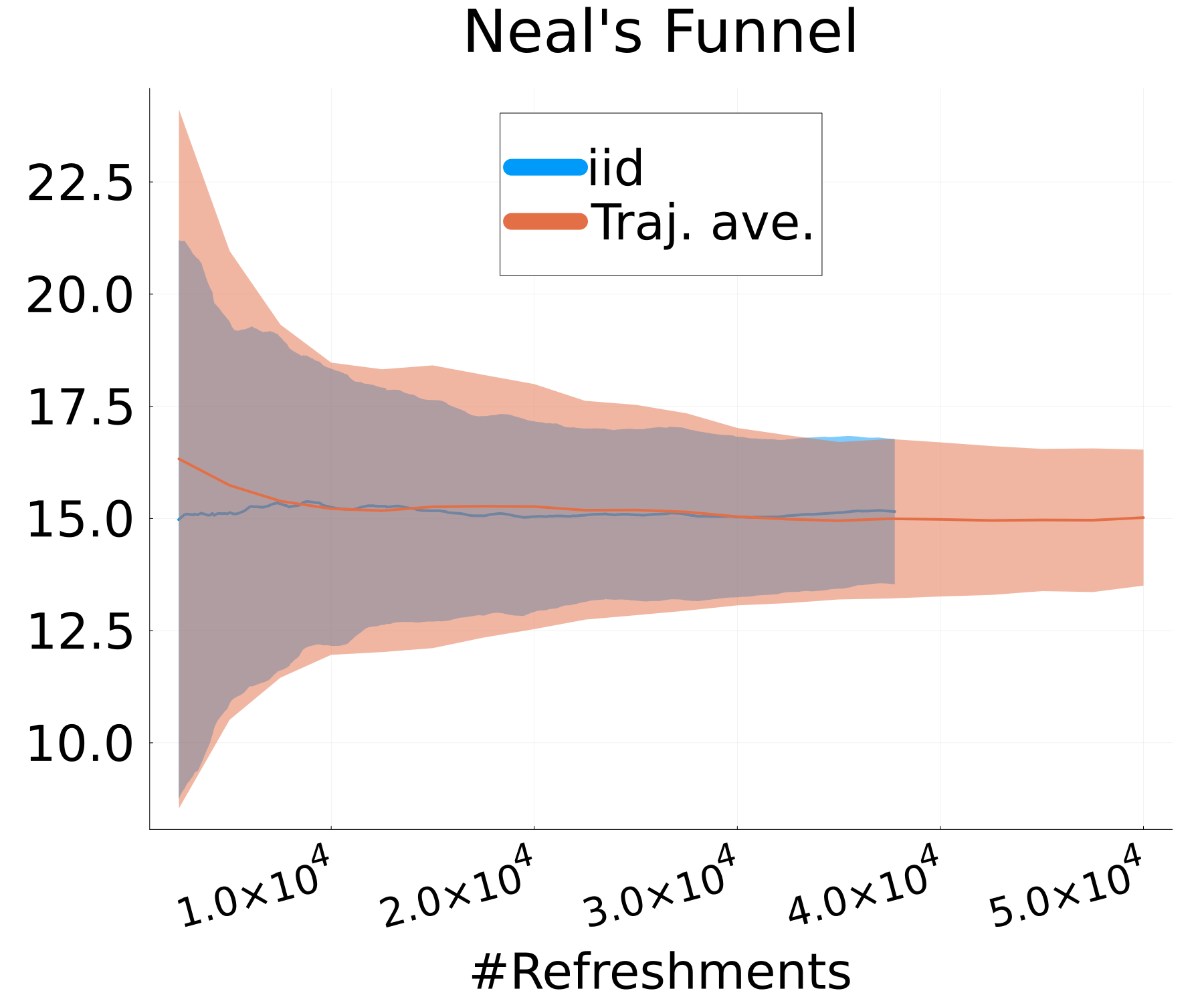}}
\end{subfigure}
\hfill
\begin{subfigure}[b]{.24\textwidth} 
    \scalebox{1}{\includegraphics[width=\textwidth]{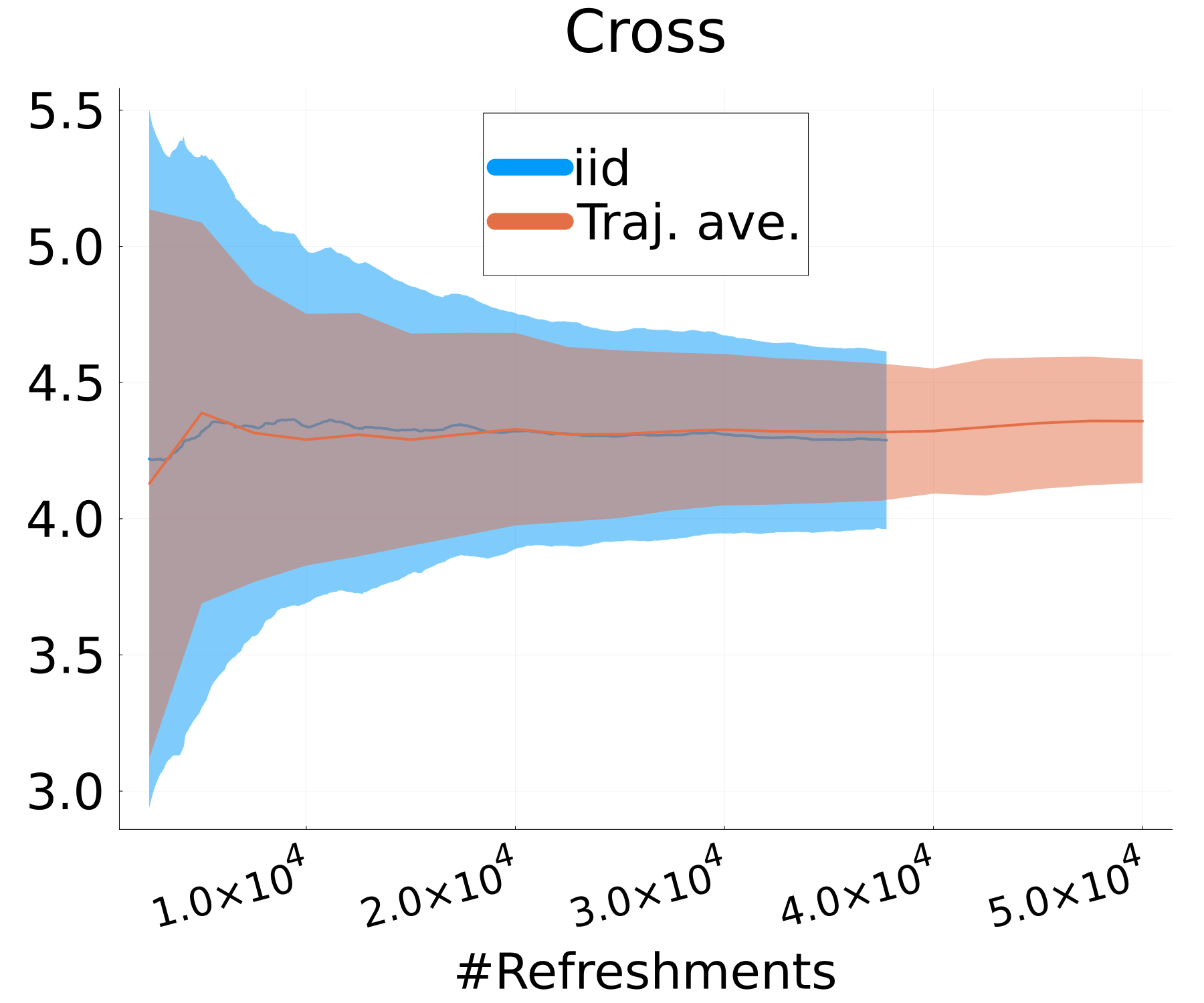}}
\end{subfigure}
\hfill
\centering 
\begin{subfigure}[b]{.24\textwidth} 
    \scalebox{1}{\includegraphics[width=\textwidth]{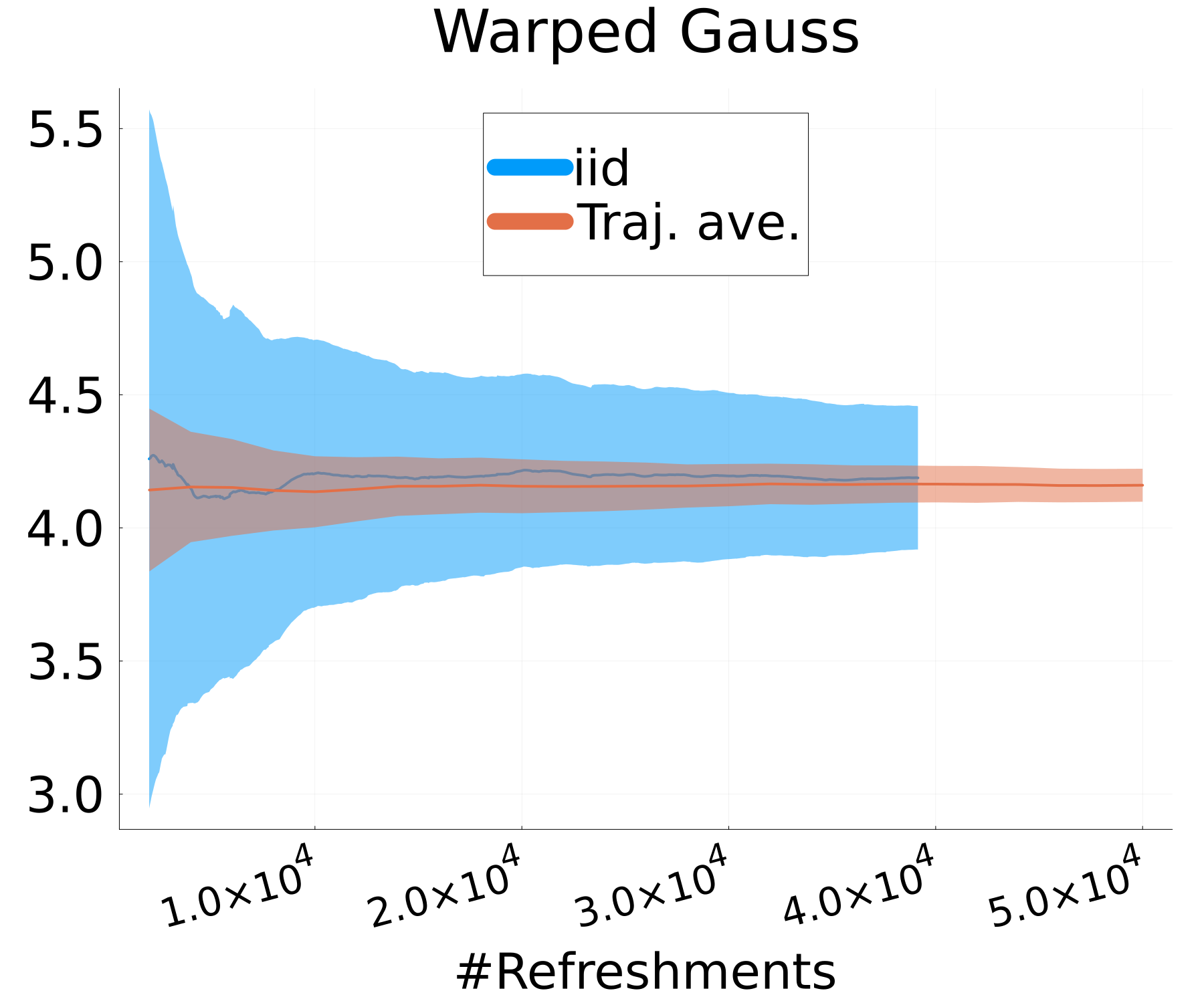}}
\end{subfigure}

\caption{
Comparison of Monte Carlo estimates of $\EE[f(X)], X \sim q_N, f(x) = \|x\|_1$ based on
individual \iid draws (blue) and trajectory-averaged estimates in \cref{eq:traj_average} (orange)
on four synthetic examples. The vertical axis
indicates the estimate of $\EE[f]$, and
the horizontal axis indicates the total number of flow transformations evaluated, i.e., total computational cost.
Note that in each example, $N$ is fixed; the number of refreshments on the horizontal axis
increases because we average over increasingly many draws $X \distiid q_N$ (blue) 
or increasingly many trajectory averages (orange).
We run $10$ trials to assess the quality of each estimate: lines 
indicate the mean, and error regions indicate standard deviation.}
\label{fig:var_compare} 
\end{figure*}

\cref{fig:var_compare} presents a comparison of the uncertainty involved in
estimating the expectations of a test function $f$ for both \texttt{MixFlow iid}
and \texttt{MixFlow single}. Specifically, it examines the streaming
estimation of $\EE[f(X)], X\sim q_N$, where $f(x) = \|x\|_1$, based on samples
generated from \texttt{MixFlow iid} and \texttt{MixFlow single} under $50{,}000$
flow map evaluations over 10 independent trials. Note that we assess computational cost via the 
number of flow map evaluations, not by the number of draws, because
the cost per draw is random in \texttt{MixFlow iid} (due to $K\distas \distUnif\{0, \dots, N-1\}$),
while the cost per draw is fixed to $N$ evaluations for the trajectory estimate in \texttt{MixFlow single}.
The results indicate that,
given equivalent computational resources, the trajectory-averaged estimates
generally exhibit lower variances between trials than the naïve \iid Monte
Carlo estimate. This observation validates \cref{prop:smallervar}.

\begin{figure*}[t!]
    \centering 
\begin{subfigure}[b]{.24\textwidth} 
    \scalebox{1}{\includegraphics[width=\textwidth]{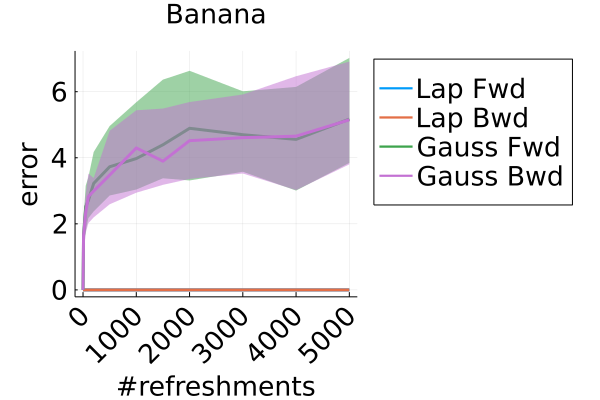}}
\end{subfigure}
\hfill
\centering 
\begin{subfigure}[b]{.24\textwidth} 
    \scalebox{1}{\includegraphics[width=\textwidth]{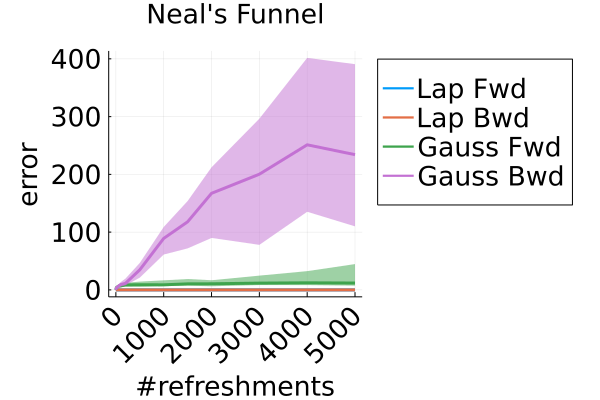}}
\end{subfigure}
\hfill
\begin{subfigure}[b]{.24\textwidth} 
    \scalebox{1}{\includegraphics[width=\textwidth]{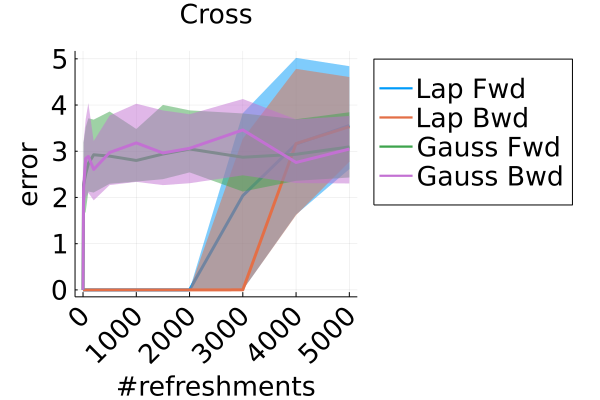}}
\end{subfigure}
\hfill
\centering 
\begin{subfigure}[b]{.24\textwidth} 
    \scalebox{1}{\includegraphics[width=\textwidth]{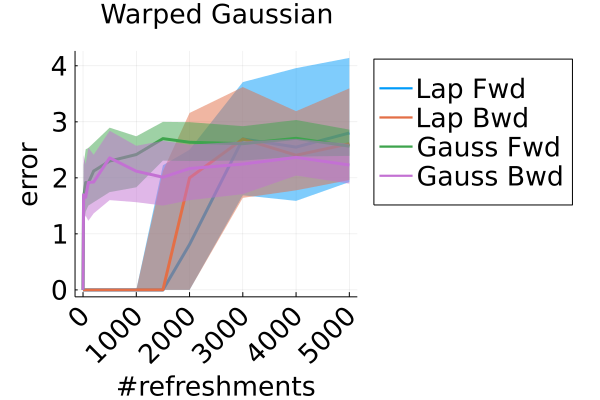}}
\end{subfigure}

\caption{Stability of composing $T^{-K}_\lambda \circ T^{K}_\lambda$
(\texttt{Fwd}) and $T^{K}_\lambda \circ T^{-K}_\lambda$ (\texttt{Bwd}) for the
four multivariate experiments with flows constructed using Gaussian
(\texttt{Gauss}) and Laplace (\texttt{Lap}) momentum distributions.  The
vertical axis shows the 2-norm error of reconstructing $(x,\rho,u)$ sampled
from $q_0$; the horizontal axis shows increasing numbers of refreshments $K$.
The lines indicate the median, and error regions indicate 25$^\text{th}$ to 75$^\text{th}$ percentile
for 100 independent samples.}
\label{fig:stability}
\end{figure*}

\begin{figure*}[t!]
    \centering 
\begin{subfigure}[b]{.24\textwidth} 
    \scalebox{1}{\includegraphics[width=\textwidth]{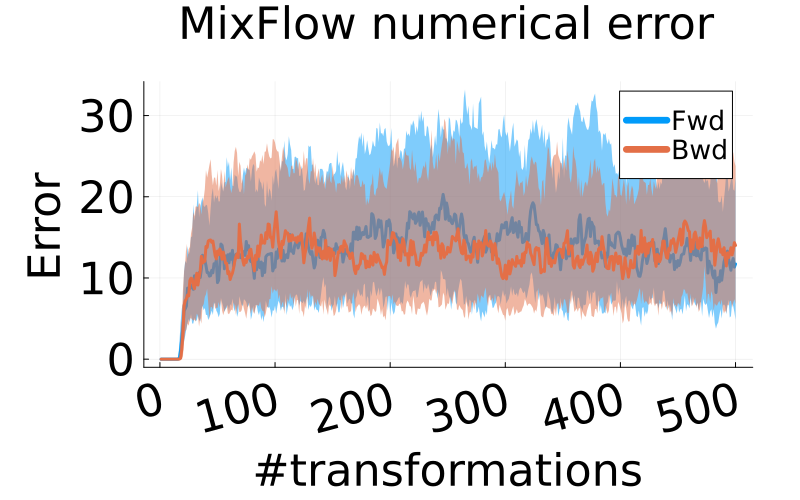}}
    \caption{(a): Banana}
\end{subfigure}
\hfill
\centering 
\begin{subfigure}[b]{.24\textwidth} 
    \scalebox{1}{\includegraphics[width=\textwidth]{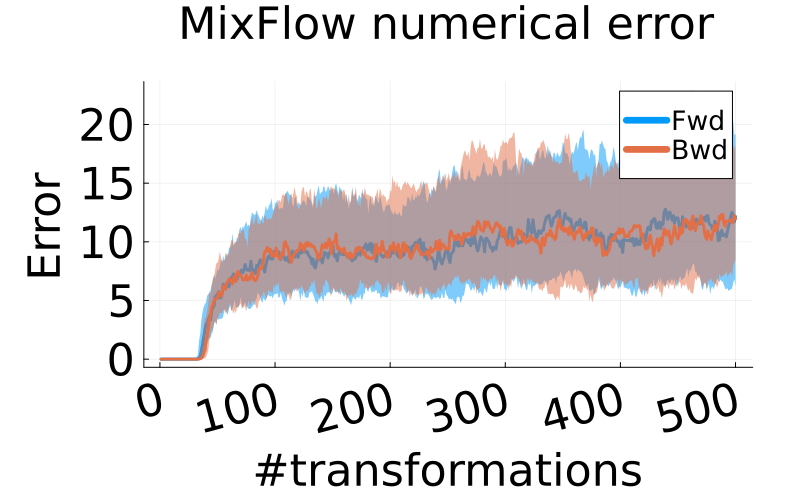}}
    \caption{(b): Neal's Funnel}
\end{subfigure}
\hfill
\begin{subfigure}[b]{.24\textwidth} 
    \scalebox{1}{\includegraphics[width=\textwidth]{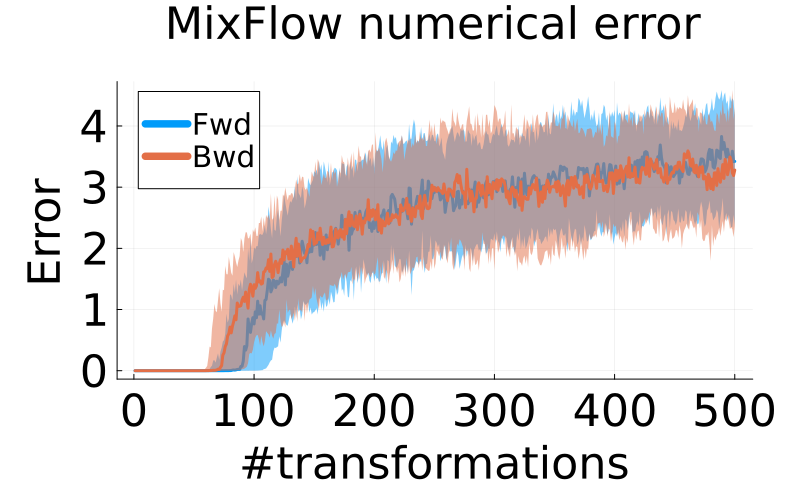}}
    \caption{(c): Cross}
\end{subfigure}
\hfill
\centering 
\begin{subfigure}[b]{.24\textwidth} 
    \scalebox{1}{\includegraphics[width=\textwidth]{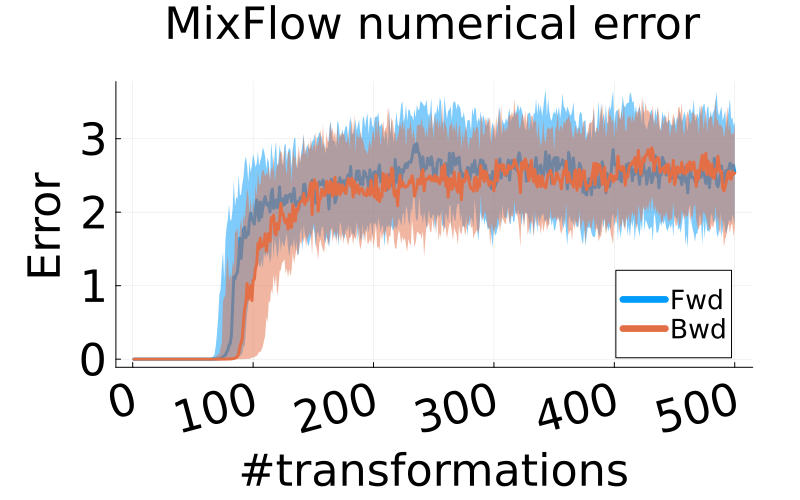}}
    \caption{(d): Warped Gaussian}
\end{subfigure}

\caption{
Large numerical errors exhibited by Hamiltonian MixFlow with Gaussian momentum on synthetic examples.
Figure shows forward (\texttt{fwd}) error $\|T^k x - \hT^k x\|$ and backward (\texttt{bwd}) error $\|T^{-k} x - \hB^{k}x\|$
comparing $k$ transformations of the forward approximate/exact maps $\hT \approx T$
and backward approximate/exact maps $\hB \approx T^{-1}$.
For the \texttt{exact} maps we use a 256-bit \texttt{BigFloat}
representation, and for the \texttt{numerical} approximations we use 64-bit \texttt{Float} representation.
The lines indicate the median, and error regions indicate 25$^\text{th}$ to
75$^\text{th}$ percentile over 100 initialization draws from the reference
distribution $q_0$.}
\label{fig:norm_flow_err}
\end{figure*}

\cref{fig:stability} shows why we opted to use a Laplace momentum distribution as opposed to a Gaussian momentum.
In particular, the numerical error of composing $T_\lambda^K \circ T_\lambda^{-K}$ and $T_\lambda^{-K} \circ T_\lambda^K$ 
(denoted as ``Bwd'' and ``Fwd'' in the legend, respectively) indicate that the flow using the Laplace momentum is 
more reliably invertible for larger numbers of refreshments than the flow with a Gaussian momentum. 
\cref{fig:norm_flow_err} uses a high-precision (256 bit) floating point representation to further 
illustrate the rapid escalation of numerical
error when evaluating forward and backward trajectories on Gaussian Hamiltonian
\texttt{MixFlows}. For all four synthetic examples, after approximately 100 flow
transformations, \texttt{MixFlows} with Gaussian momentum exhibits an error on the 
scale of the target distribution itself (per the contour plots (d)-(g) in \cref{fig:vis_plots}).
This may be due to the fact that the normal has very light tails; for large momentum values in the right tail,
the CDF is $\approx 1$, which gets rounded to exactly 1 in floating point
representation. Our implementation of the CDF and inverse CDF was fairly na\"ive, 
so it is possible that stability could be improved with more care taken to prevent rounding error.
We leave a more careful examination of this error to future work; for this work, using
a Laplace distribution momentum sufficed to maintain a low numerical error.

\begin{figure*}[t!]
    \centering 
\begin{subfigure}[b]{0.75\textwidth} 
    \scalebox{1}{\includegraphics[width=\textwidth,trim=0 0 0 30, clip]{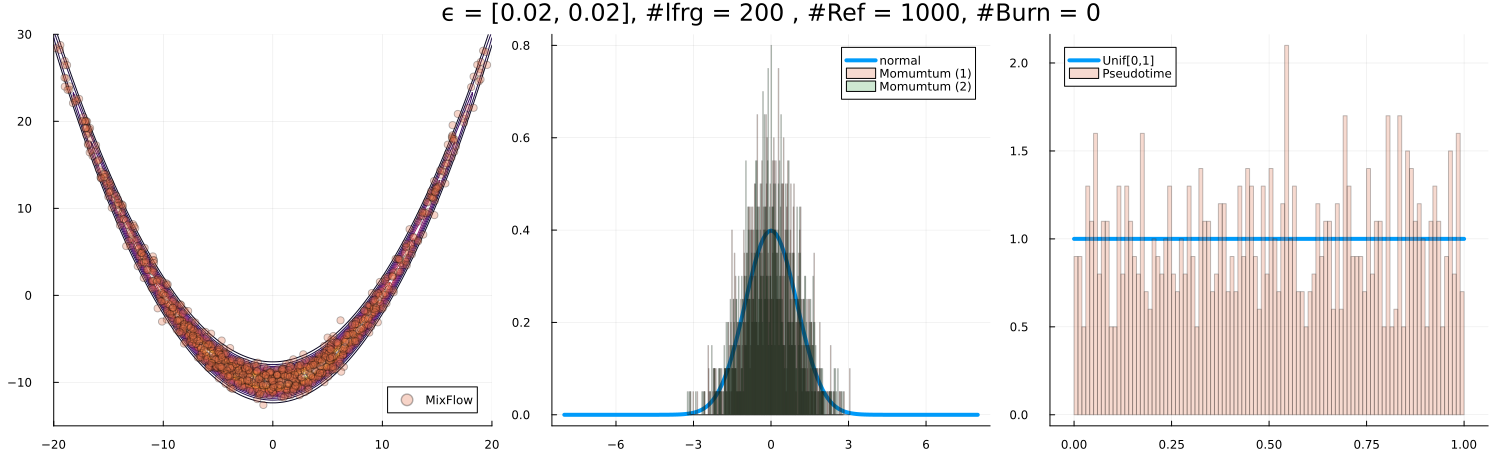}}
    \caption{(a) Banana distribution} \label{fig:banana_sample}
\end{subfigure}
\hfill
\centering 
\begin{subfigure}[b]{0.75\textwidth} 
    \scalebox{1}{\includegraphics[width=\textwidth,trim=0 0 0 30, clip]{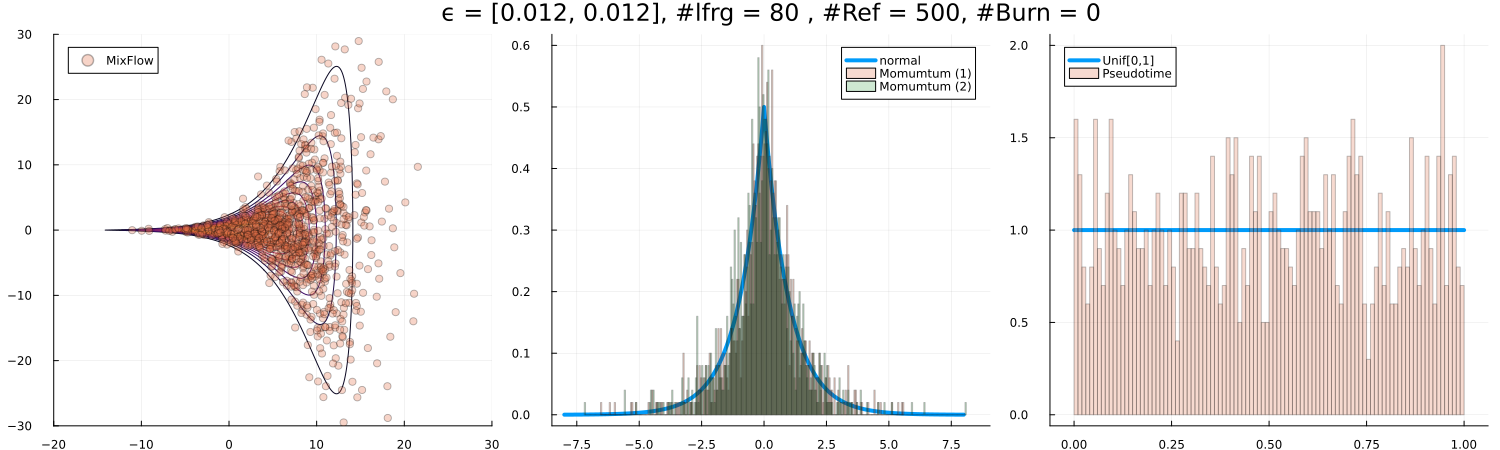}}
    \caption{(b) Neal's funnel} \label{fig:funnel_sample}
\end{subfigure}
\hfill
\centering
\begin{subfigure}[b]{0.75\textwidth}
    \scalebox{1}{\includegraphics[width=\textwidth,trim=0 0 0 30, clip]{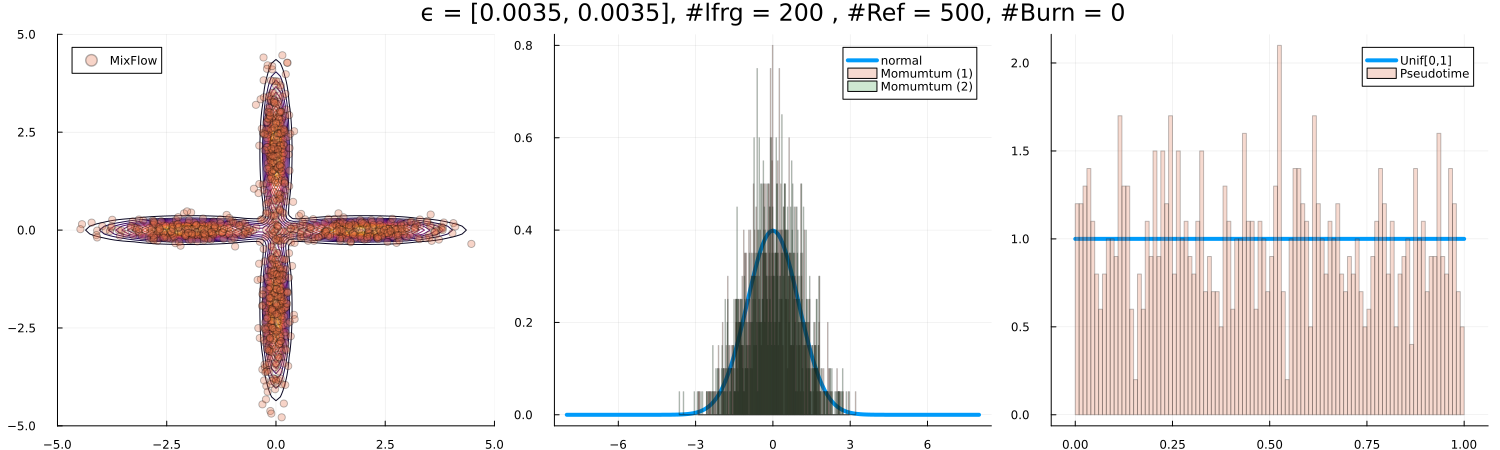}}
    \caption{(c) Cross distribution}\label{fig:cross_sample}
\end{subfigure}
\hfill
    \centering 
\begin{subfigure}[b]{0.75\textwidth} 
    \scalebox{1}{\includegraphics[width=\textwidth,trim=0 0 0 30, clip]{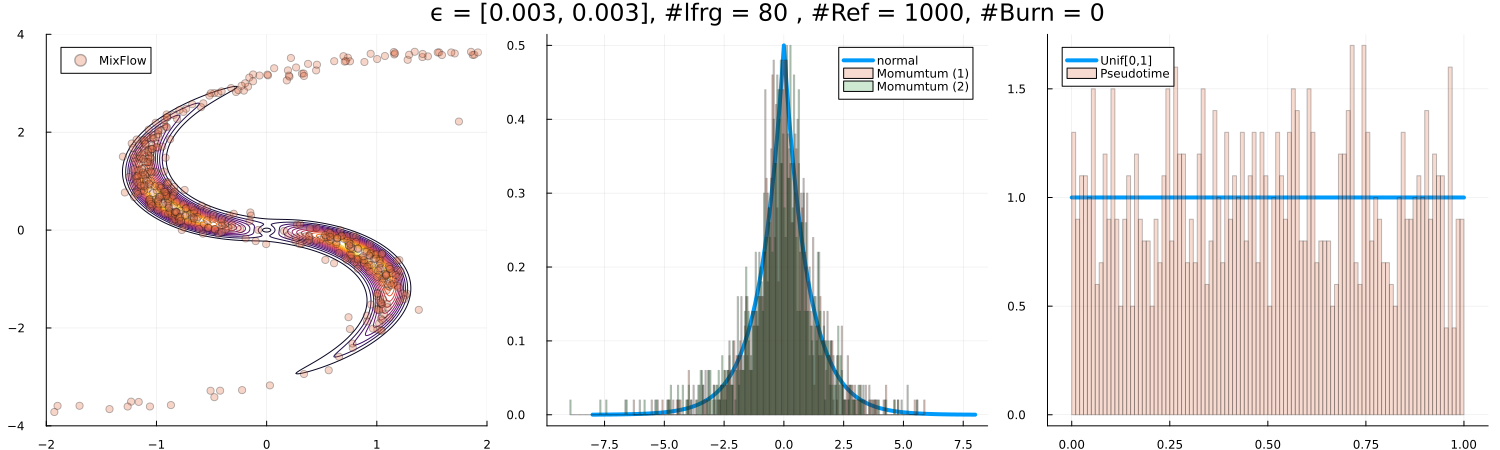}}
    \caption{(d) Warped Gaussian}\label{fig:warp_sample}
\end{subfigure}

\caption{Scatter plots and histograms for the $x$-, $\rho$-, and $u$-marginals in the four synthetic experiments.}
\label{fig:2d_sample}
\end{figure*}

Finally, \cref{fig:2d_sample} provides a more comprehensive set of sample
histograms for these experiments, showing the $x$-, $\rho$-, and $u$-marginals.
It is clear that \texttt{MixFlow} generates samples for each variable reliably
from the target marginal.

\subsection{Higher-dimensional synthetic experiment} \label{apdx:highdim}
\begin{figure*}[t!]
    \centering 
\begin{subfigure}[b]{.24\textwidth} 
    \scalebox{1}{\includegraphics[width=\textwidth]{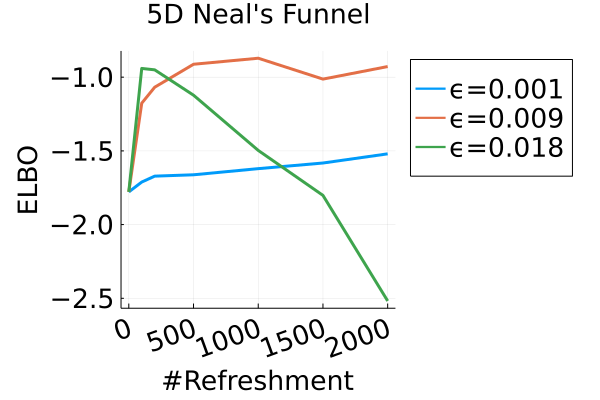}}
    \caption{(a) \label{fig:5d_elbo}}
\end{subfigure}
\hfill
\centering 
\begin{subfigure}[b]{.24\textwidth} 
    \scalebox{1}{\includegraphics[width=\textwidth]{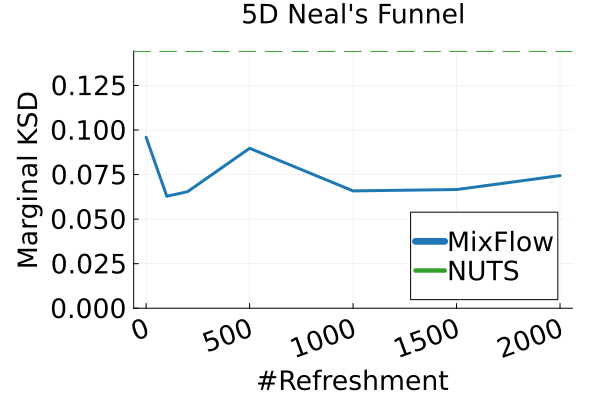}}
    \caption{(b)\label{fig:5d_ksd}}
\end{subfigure}
\hfill
\begin{subfigure}[b]{.24\textwidth} 
    \scalebox{1}{\includegraphics[width=\textwidth]{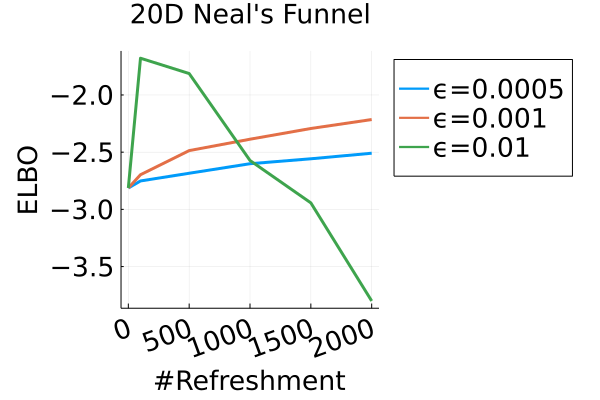}}
    \caption{(c) \label{fig:20d_elbo}}
\end{subfigure}
\hfill
\centering 
\begin{subfigure}[b]{.24\textwidth} 
    \scalebox{1}{\includegraphics[width=\textwidth]{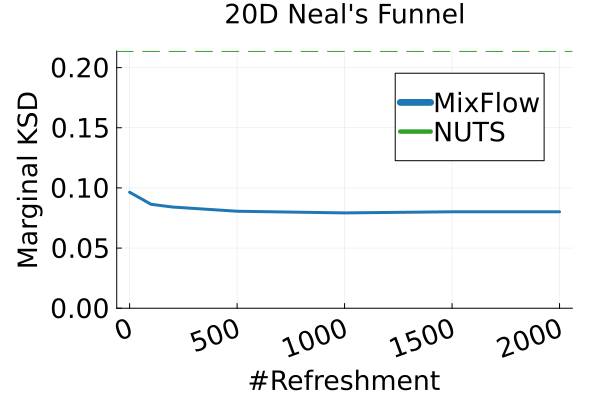}}
    \caption{(d)\label{fig:20d_ksd}}
\end{subfigure}

\caption{ELBO versus step size (\ref{fig:5d_elbo},\ref{fig:20d_elbo}) and KSD comparison with \texttt{NUTS} (\ref{fig:5d_ksd},\ref{fig:20d_ksd}) 
for the 5- and 20-dimensional Neal's funnel examples.}
\label{fig:higher_dim_metrics}
\end{figure*}

We also tested two higher dimensional Neal's funnel target distributions of $x \in \reals^d$ where $d=5, 20$.
In particular, we used the target distribution
\[
x_1 \distas \distNorm\left( 0, \sigma^2 \right), \quad\text{and}\quad \forall i \in \{2, \cdots, d\},\,\, x_i \given x_1 \distiid \distNorm\left( 0, \exp\left(\frac{x_1}{2}\right) \right), 
\]
with $\sigma^2 = 36$.
We use $80$ leapfrogs between refreshments and $\eps = 0.0009$ when $d = 5$, and 
$100$ leapfrogs between refreshments and  $\eps = 0.001$ when $d = 20$
(\cref{fig:5d_elbo,fig:20d_elbo} shows the ELBO comparison used to tune the step size $\epsilon$).
\cref{fig:5d_ksd,fig:20d_ksd} confirm via the KSD that the method performs as well 
as \texttt{NUTS} in higher-dimensional cases. 

\captionsetup[subfigure]{labelformat=empty}

\captionsetup[subfigure]{labelformat=empty}

\captionsetup[subfigure]{labelformat=empty}

\captionsetup[subfigure]{labelformat=empty}

\captionsetup[subfigure]{labelformat=empty}

\subsection{Additional experiments for synthetic examples} \label{apdx:2d_more}

\captionsetup[subfigure]{labelformat=empty, width=\columnwidth}
\begin{figure}[t!]
\centering 
\begin{subfigure}[b]{.48\columnwidth} 
	\includegraphics[width=0.32\columnwidth]{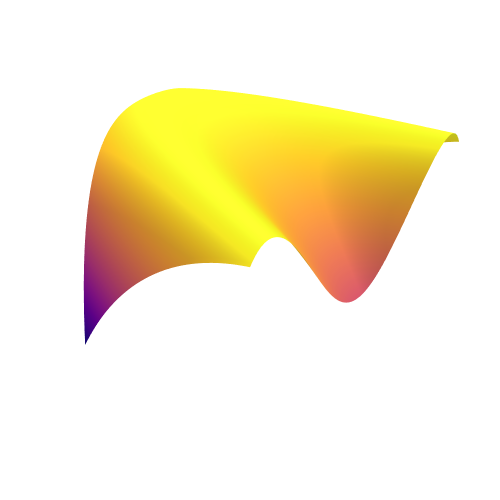}
	\includegraphics[width=0.32\columnwidth]{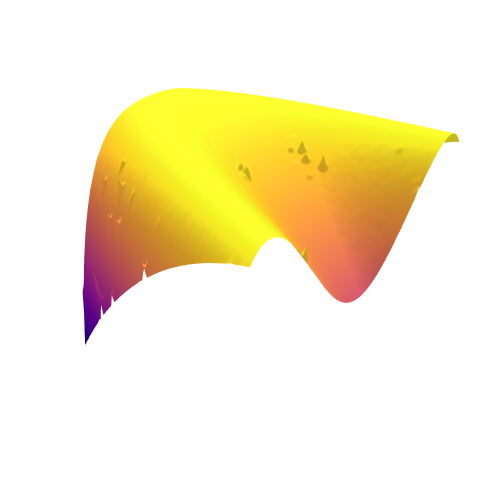}
    \includegraphics[width=0.32\columnwidth]{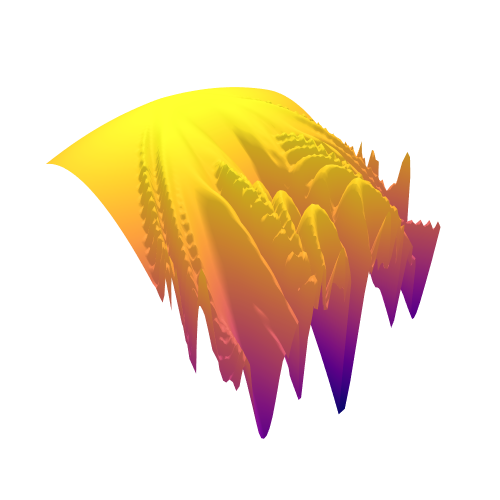}
    \caption{(a) Banana. From left to right: True, \texttt{MixFlow}, and \texttt{NEO}\label{fig:neo_banana_density}}
\end{subfigure}
\hfill
\centering 
\begin{subfigure}[b]{.48\columnwidth} 
	\includegraphics[width=0.32\columnwidth]{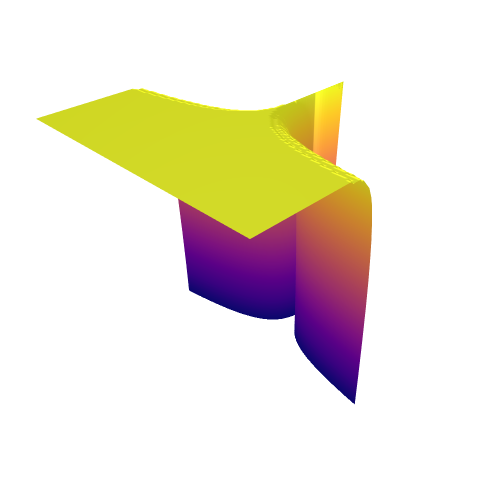}
	\includegraphics[width=0.32\columnwidth]{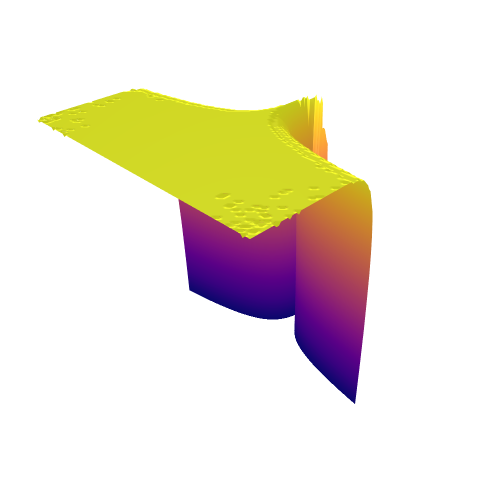}
    \includegraphics[width=0.32\columnwidth]{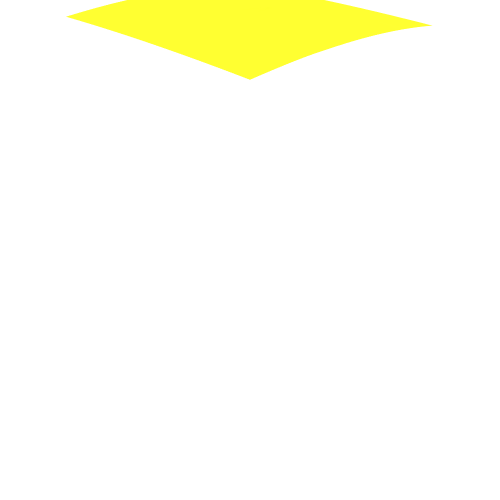}
    \caption{(b) Neal's Funnel. From left to right: True, \texttt{MixFlow}, and \texttt{NEO}\label{fig:neo_funnel_density}}
\end{subfigure}
\hfill
\centering
\begin{subfigure}[b]{0.48\columnwidth}
	\includegraphics[width=0.32\columnwidth]{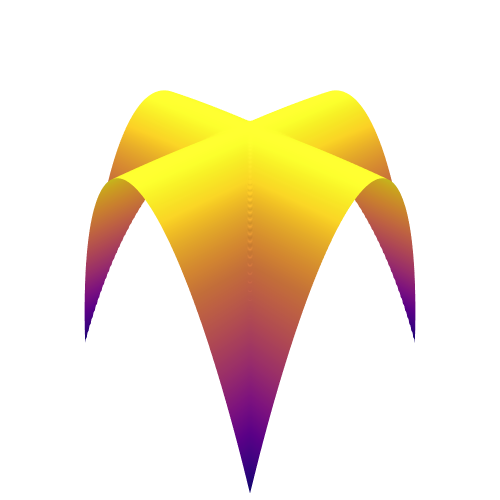}
	\includegraphics[width=0.32\columnwidth]{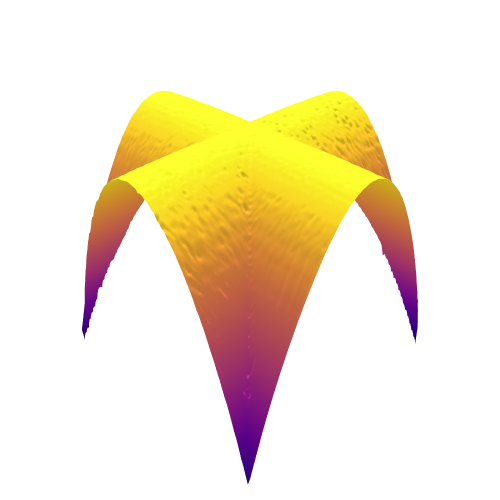}
	\includegraphics[width=0.32\columnwidth]{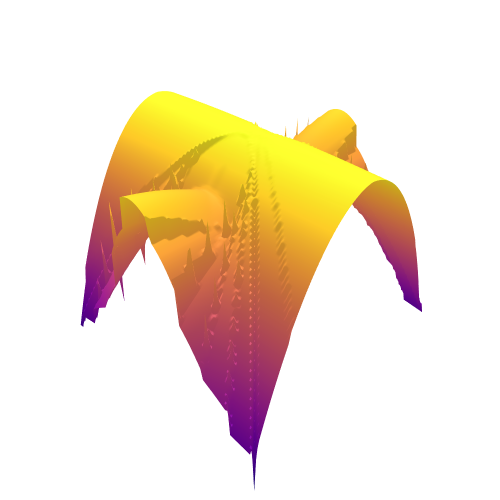}
    \caption{(c) Cross. From left to right: True, \texttt{MixFlow}, and \texttt{NEO} \label{fig:neo_cross_density}}
\end{subfigure}
\hfill
    \centering 
\begin{subfigure}[b]{.48\columnwidth} 
	\includegraphics[width=0.32\columnwidth]{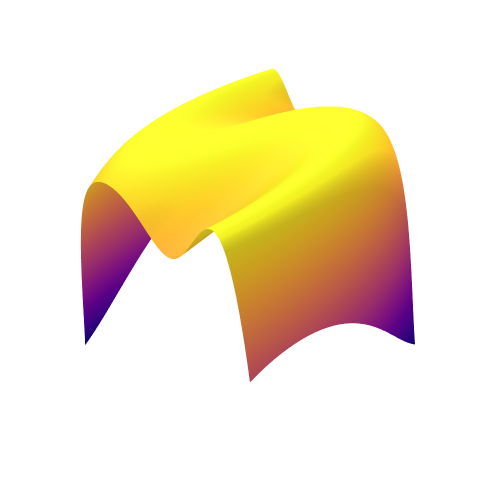}
	\includegraphics[width=0.32\columnwidth]{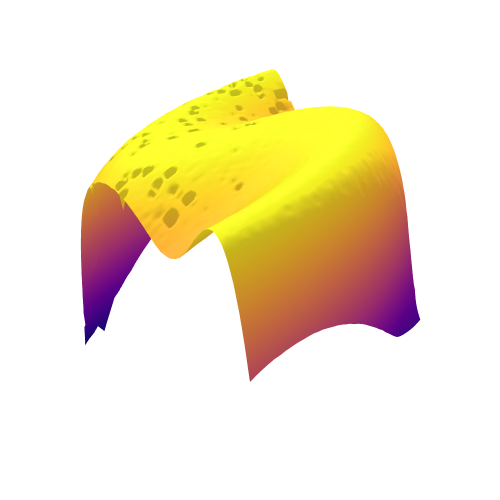}
	\includegraphics[width=0.32\columnwidth]{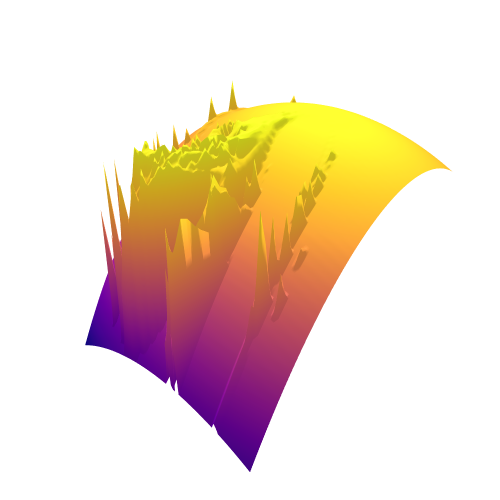}
    \caption{(d) Warped Gauss. From left to right: True, \texttt{MixFlow}, and \texttt{NEO}  \label{fig:neo_warp_density}}
\end{subfigure}
\caption{Visualization of sliced exact (left) and \texttt{MixFlow}-approximated (middle) joint log density  (middle),
and sliced joint log density of tuned \texttt{NEO} importance sampling proposal (right) 
for banana (\ref{fig:neo_banana_density}), funnel (\ref{fig:neo_funnel_density}), 
cross (\ref{fig:neo_cross_density}), and warped Gaussian (\ref{fig:neo_warp_density}).}
\label{fig:neo_density}
\end{figure}

\begin{figure}[t!]
\centering 
\includegraphics[width=0.24\columnwidth]{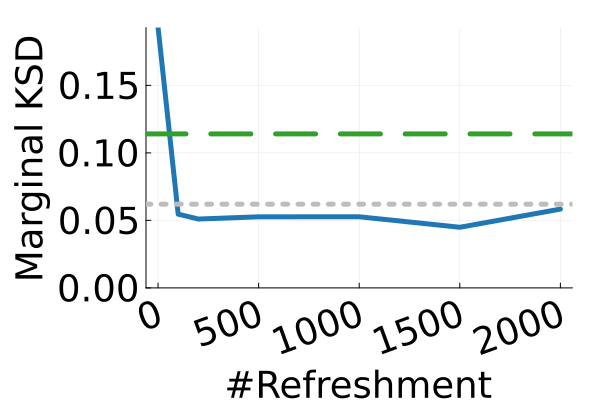}
\includegraphics[width=0.24\columnwidth]{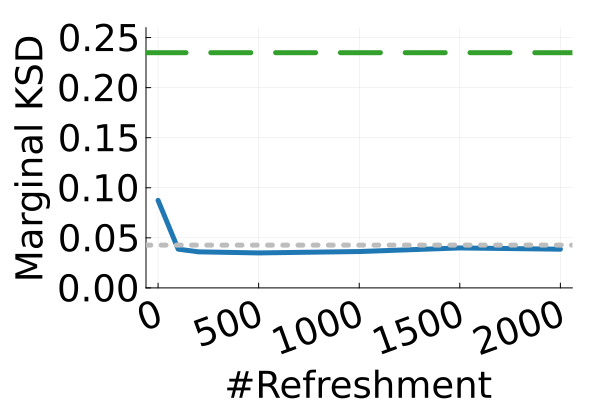}
\includegraphics[width=0.24\columnwidth]{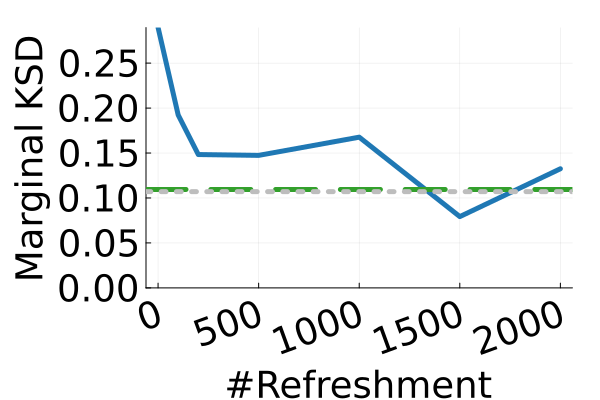}
\includegraphics[width=0.24\columnwidth]{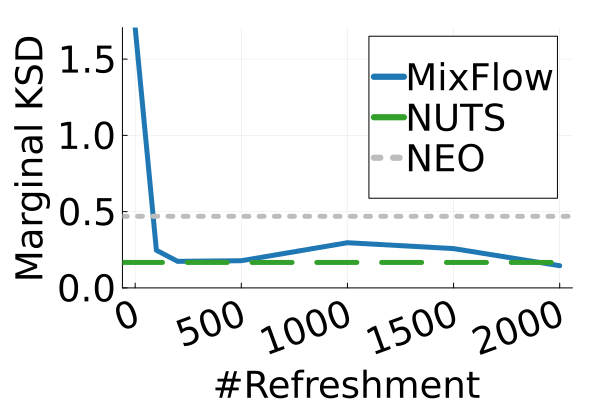}
\caption{Marginal KSD of tuned \texttt{MixFlow} versus number of refreshments $N$, and comparison to \texttt{NUTS} and \texttt{NEO}}\label{fig:2d_ksd}
\end{figure}

\begin{figure}[t!]
\includegraphics[width=0.24\columnwidth]{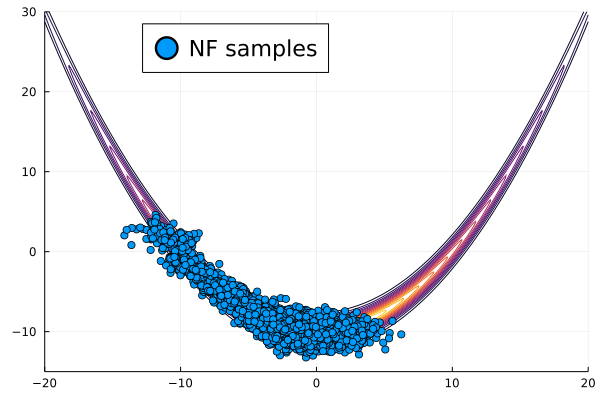}
\includegraphics[width=0.24\columnwidth]{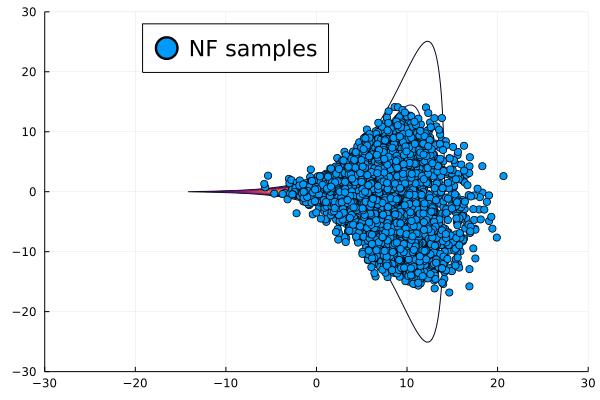}
\includegraphics[width=0.24\columnwidth]{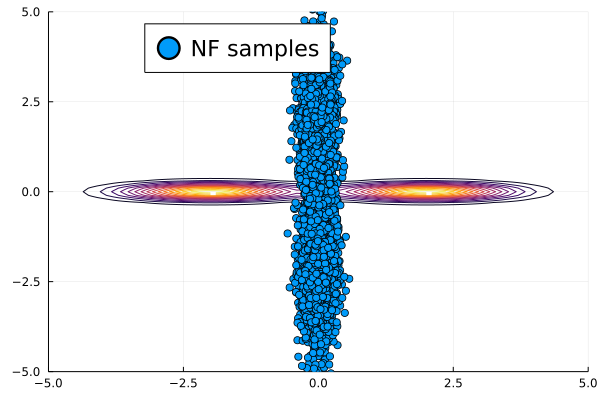}
\includegraphics[width=0.24\columnwidth]{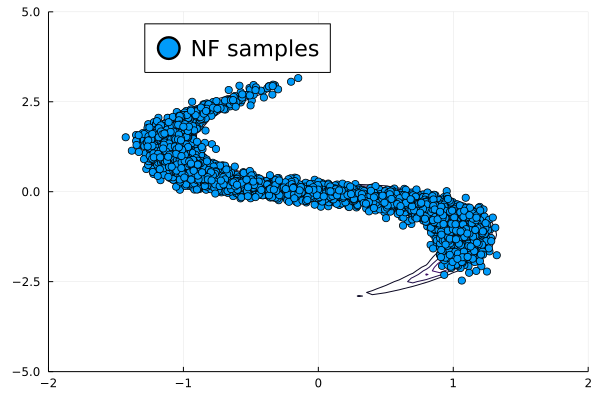}
\caption{Scatter plot of $5000$ \iid samples generated from trained \texttt{NF}.}\label{fig:nf_scatter}
\end{figure}

\begin{figure}[t!]
    \begin{subfigure}{\columnwidth}
		\includegraphics[width=0.24\columnwidth]{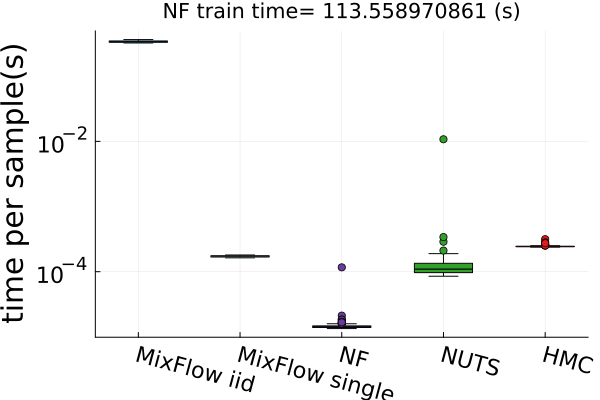}
		\includegraphics[width=0.24\columnwidth]{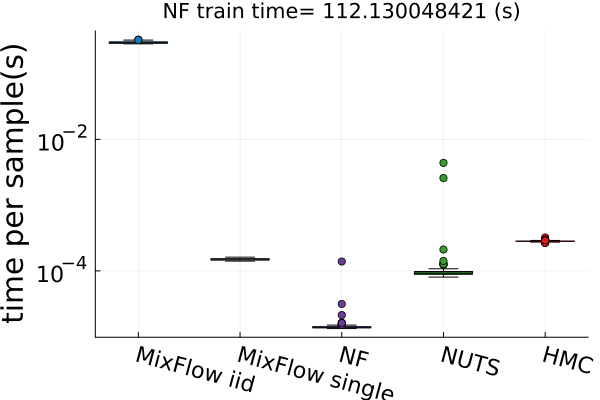}
		\includegraphics[width=0.24\columnwidth]{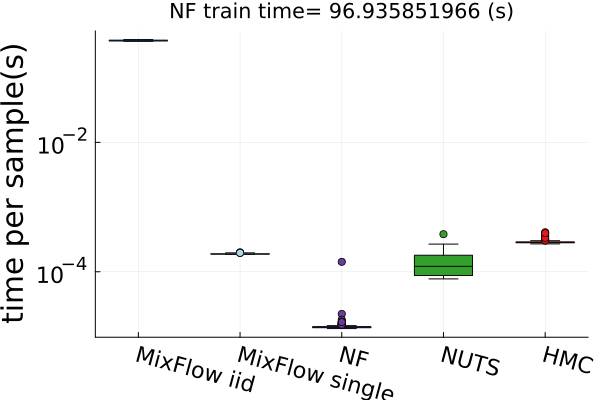}
		\includegraphics[width=0.24\columnwidth]{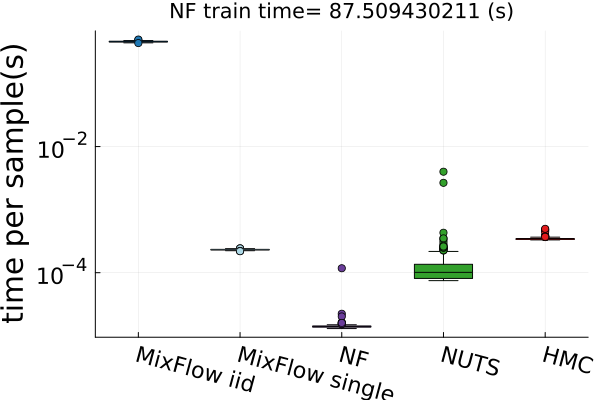}
		\caption{(a): Sampling time comparison to \texttt{NF},\texttt{NUTS}, and \texttt{HMC} (100 trials). }\label{fig:time_per_sample}
	\end{subfigure}
    \begin{subfigure}{\columnwidth}
		\includegraphics[width=0.24\columnwidth]{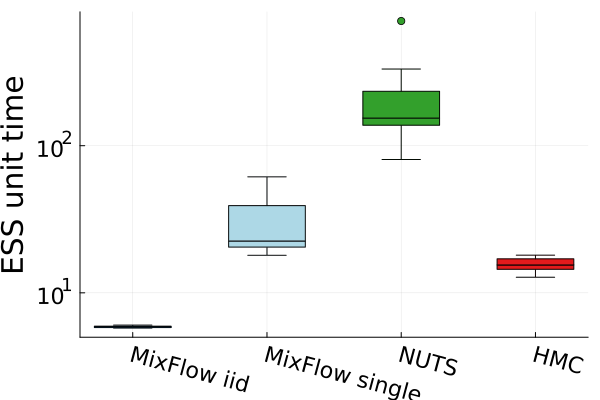}
		\includegraphics[width=0.24\columnwidth]{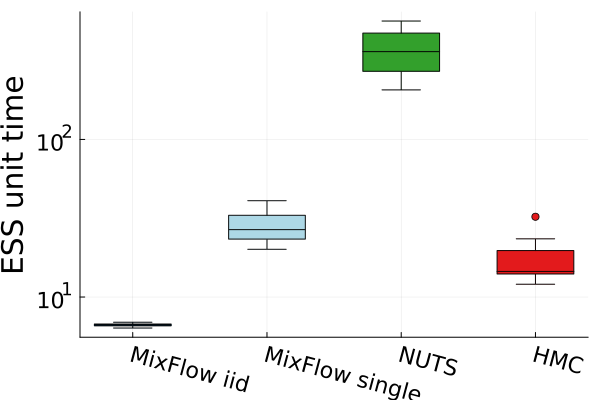}
		\includegraphics[width=0.24\columnwidth]{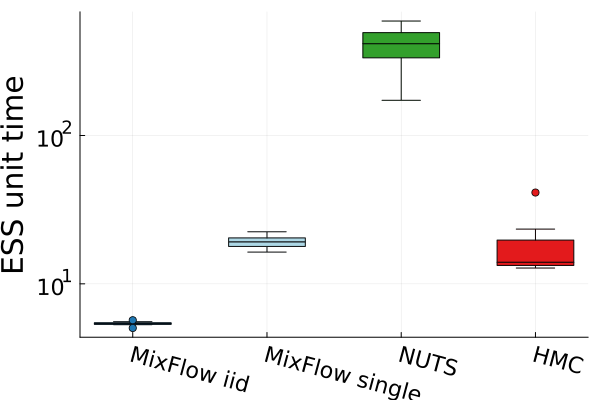}
		\includegraphics[width=0.24\columnwidth]{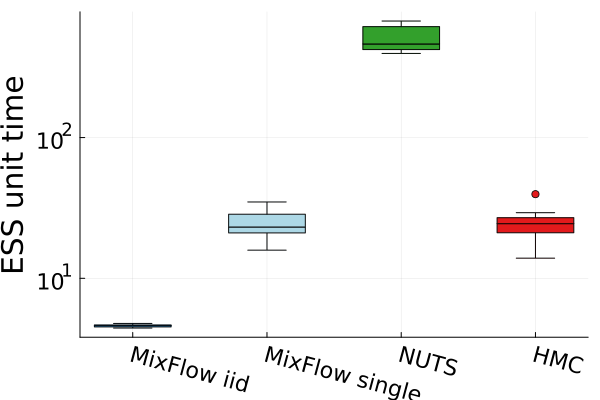}
		\caption{(b): Per second ESS comparison to \texttt{NUTS} and \texttt{HMC} (100 trials).}\label{fig:2d_ess}
	\end{subfigure}
\caption{Time results for four multivariate synthetic examples: from left to right, each column corresponds to Banana, Neal's funnel, Cross, and warped Gaussian respectively.}\label{fig:2d_timing}
\end{figure}

\captionsetup[subfigure]{labelformat=empty}
\begin{figure*}[t!]
\centering 
\begin{subfigure}[b]{.24\textwidth}
    \centering 
    \includegraphics[width=\textwidth]{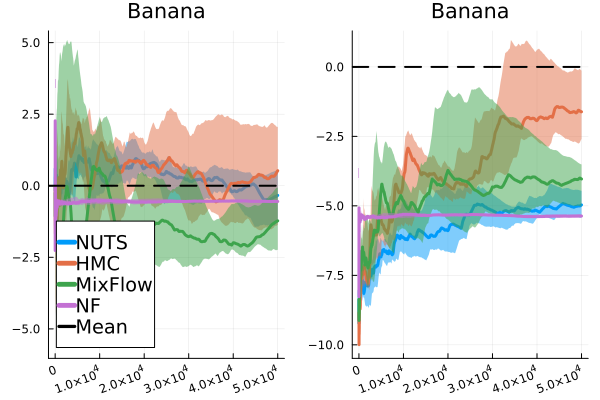}
    \caption{(a)}
    \label{fig:banana_mean}
\end{subfigure}
\hfill
\centering 
\begin{subfigure}[b]{.24\textwidth} 
    \centering 
    \includegraphics[width=\textwidth]{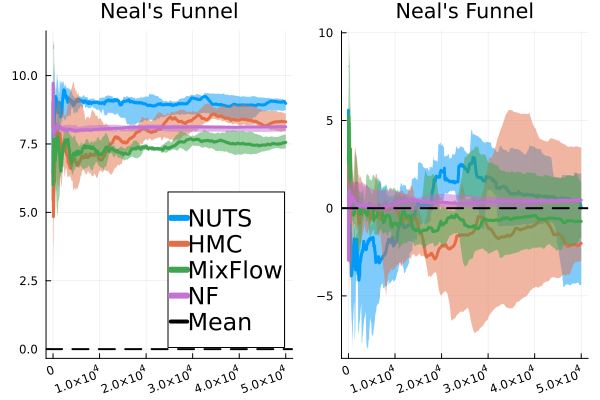}
    \caption{(b)}
    \label{fig:funnel_mean}
\end{subfigure}
\hfill
\centering
\begin{subfigure}[b]{.24\textwidth} 
    \centering 
    \includegraphics[width=\textwidth]{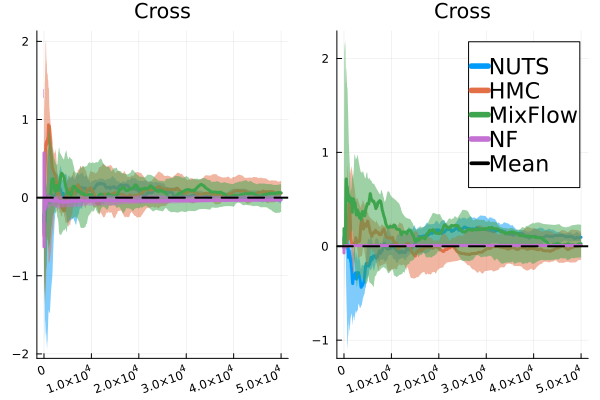}
    \caption{(c)}
    \label{fig:cross_mean}
\end{subfigure}
\hfill
\centering
\begin{subfigure}[b]{.24\textwidth} 
    \centering 
    \includegraphics[width=\textwidth]{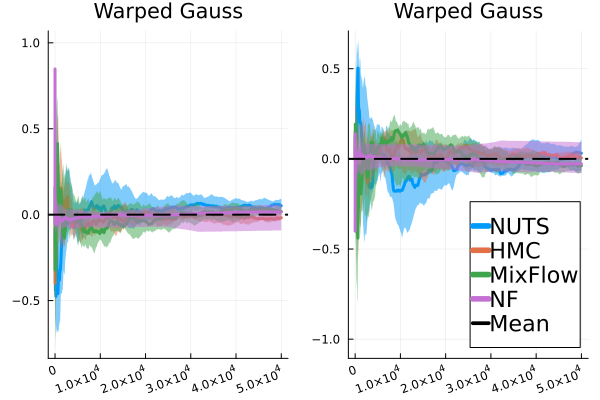}
    \caption{(d)}
    \label{fig:warp_mean}
\end{subfigure}
\hfill
\centering 
\begin{subfigure}[b]{.24\textwidth}
    \centering 
    \includegraphics[width=\textwidth]{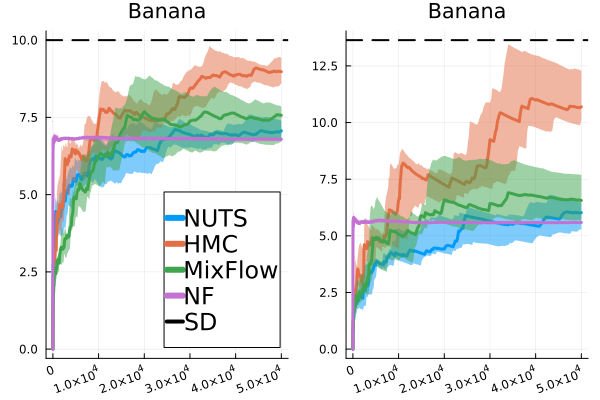}
    \caption{(e)}
    \label{fig:banana_var}
\end{subfigure}
\hfill
\centering 
\begin{subfigure}[b]{.24\textwidth} 
    \centering 
    \includegraphics[width=\textwidth]{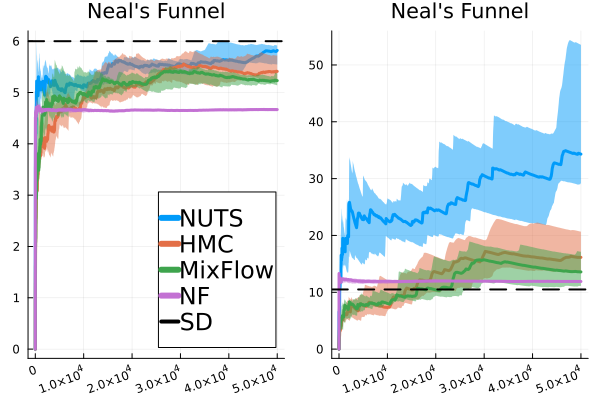}
    \caption{(f)}
    \label{fig:funnel_var}
\end{subfigure}
\hfill
\centering
\begin{subfigure}[b]{.24\textwidth} 
    \centering 
    \includegraphics[width=\textwidth]{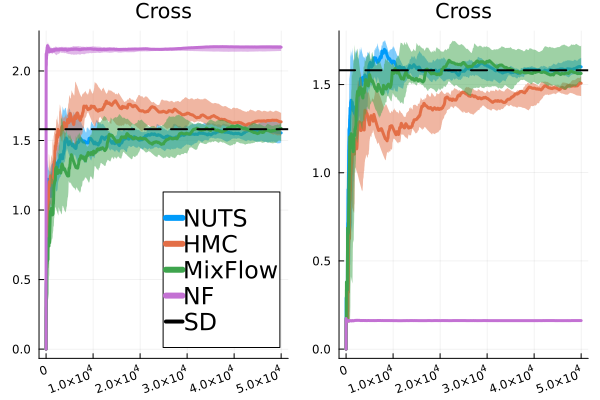}
    \caption{(g)}
    \label{fig:cross_var}
\end{subfigure}
\hfill
\centering
\begin{subfigure}[b]{.24\textwidth} 
    \centering 
    \includegraphics[width=\textwidth]{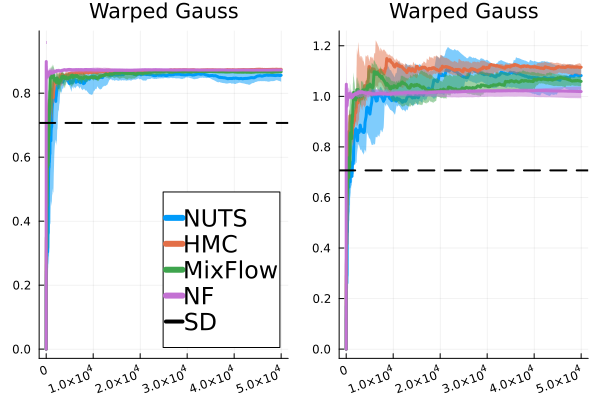}
    \caption{(h)}
    \label{fig:warp_var}
\end{subfigure}
\caption{Streaming mean and standard deviation (SD) estimation ($50,000$ samples) for four $2$-dimensional
synthetic distributions. For each distribution, the two plots on the top row correspond to the two marginal means,
and the two plots on the bottom row correspond to the two marginal SDs. Each plot shows the evolution of coordinate-wise mean/SD
estimates, using samples from \texttt{MixFlow single} (green), \texttt{NUTS}
(blue), \texttt{HMC} (red), and \texttt{NF} (purple); black dashed line
indicates the true target mean/SD, which is estimated using $50,000$ samples from
the actual synthetic target distribution. The lines indicate the median, and
error regions indicate $25^\text{th}$ to $75^\text{th}$ percentile from $10$ runs.}
\label{fig:running_estimation}
\end{figure*}

In this section, we provide additional comparisons of \texttt{MixFlow} 
against \texttt{NUTS}, \texttt{HMC}, \texttt{NEO} and a generic normalizing flow method---planar flow (\texttt{NF}) \citep{rezende2015variational} on all four synthetic examples in \cref{apdx:2d}. 
For this set of experiments, all of the settings for \texttt{MixFlow} are the same as outlined in 
\cref{apdx:2d}. 
\texttt{HMC} uses the same leapfrog step size and number of 
leapfrogs steps (between refreshments) as \texttt{MixFlow}. 
For \texttt{NF}, $5$ Planar layers \citep{rezende2015variational} that 
contain $45$ parameters to be optimized (9 parameters for each Planar layer of a $4$-dimensional Planar flow) are used unless otherwise stated. We train \texttt{NF} using 
ADAM until convergence ($100,000$ iterations except where otherwise noted) with the initial step size 
set to $0.001$. And at each step, $10$ samples are used for gradient estimation. The initial distribution 
for \texttt{MixFlow}, \texttt{NF} and \texttt{NUTS} is set to be the mean-field Gaussian approximation,
and \texttt{HMC} and \texttt{NUTS} are initialized using the learned mean of the mean-field Gaussian approximation.
All parameters in \texttt{NF} are initialized using random samples from the standard Gaussian
distribution.

\cref{fig:neo_density} compares the sliced log density of tuned \texttt{MixFlow} and the importance sampling proposal of tuned \texttt{NEO},  
visualized in a similar fashion as \cref{fig:vis_plots} in \cref{sec:visualization}.
It is clear that \texttt{NEO} does not provide a high-quality density approximation, 
while the log density of \texttt{MixFlow} visually matches the target.
Indeed, unlike our method, due to the use of a nonequilibrium dynamic, 
\texttt{NEO} does not provide a similar convergence guarantee of 
its proposal distribution as simulation length increases. 

\cref{fig:2d_ksd} presents comparisons of the marginal sample qualities of \texttt{MixFlow} to 
\texttt{NUTS} and \texttt{NEO}---both are exact MCMC methods.
Results show that \texttt{MixFlow} produces comparable marginal sample quality, if not slightly better, 
than both MCMC methods. 
Note that it involves a nontrivial tuning process to
achieve the displayed performance of \texttt{NEO}. Concrete tuning strategies
are explained in the beginning of \cref{apdx:expt_detail}.  In fact, finding a
good combination of discretization step, friction parameter, 
simulation length, and mass matrix of momentum distribution is necessary
for \texttt{NEO} to behave reasonably; \cref{tab:aggregated_neo} summarizes 
the performance of \texttt{NEO} under various settings for both synthetic and real data experiments. 
The standard deviation of marginal sample KSD  can change drastically across different settings, 
meaning that the marginal sample quality can be sensitive to the choice of hyperparameters.  
Moreover, due to the usage of an unstable Hamiltonian dynamics (with friction),
one must choose hyperparameters carefully to avoid NaN values.  
The last column of \cref{tab:aggregated_neo} shows the number of hyperparameter combinations that lead to 
NaN values during sampling. 

\cref{fig:2d_elbo} compares the joint ELBO values across various leapfrog step
sizes for \texttt{MixFlow} against that of \texttt{NF}. We see that in all four examples, 
\texttt{NF} produces a smaller ELBO than \texttt{MixFlow} with a reasonable step size,
which implies a lower quality target approximation. Indeed,
\cref{fig:nf_scatter} shows that the trained \texttt{NF}s fail to capture the
shape of the target distributions. 
Although one may expect the performance of \texttt{NF} 
to improve if it were given more layers, we will show that this is not the case in a later paragraph.

\cref{fig:time_per_sample} compares the sampling efficiency of
\texttt{MixFlow} against \texttt{NUTS}, \texttt{HMC}, and \texttt{NF}.
We see that \texttt{MixFlow iid} is the slowest, because each
sample is generated by passing through the entire flow. 
However, we see that by taking all intermediate samples as in \texttt{MixFlow
single}, we can generate samples just as fast as \texttt{NUTS} and \texttt{HMC}.
On the other hand, while \texttt{NF} is fastest for sampling, 
it requires roughly $2$ minutes for training, which alone allows 
\texttt{MixFlow single}, \texttt{NUTS}, and \texttt{HMC} to 
generate over $1$ million samples in these examples.
A more detailed discussion about this trade-off for \texttt{NF} is presented later. 

\cref{fig:2d_ess} further shows the computational efficiency in terms of
effective samples size (ESS) per second. 
The smaller per second ESS of \texttt{MixFlow iid} is due to its slower sampling time. However, we emphasize that these samples are \iid.
\texttt{NUTS} overall achieves a higher per second ESS. 
\texttt{NUTS} is performant because of the much longer
trajectories it produces (it only terminates once it detects a ``U-turn''). 
This
is actually an illustration of a limitation of the ESS per unit time as a
measurement of performance. Because NUTS generates longer trajectories, it has a
lower sample autocorrelation and a higher ESS; but \cref{fig:running_estimation} shows that the actual
estimation performance of \texttt{NUTS} is comparable to the other methods. 
Note that it is also possible to
incorporate the techniques used in \texttt{NUTS} to our method, which we leave
for future work.

As mentioned above, ESS mainly serves as a practical index for the relative efficiency of 
using a sequence of dependent samples, as opposed to independent samples, to estimate
certain summary statistics of the target distribution. In this case, \texttt{MixFlow single}
can be very useful. 
\cref{fig:running_estimation} demonstrate the performance of \texttt{MixFlow single}, 
\texttt{NUTS}, \texttt{HMC}, and \texttt{NF} when estimating the coordinate-wise
means and standard deviations (SD) of target distributions.
We see that \texttt{MixFlow single}, \texttt{NUTS}, and \texttt{HMC} generally show
similar performance in terms of convergence speed and estimation precision. 
While \texttt{NF} converges very quickly due to \iid sampling, it does seem to struggle more
at identifying the correct statistics, particularly the standard deviation, given its limited approximation 
quality. It is worth noting that, unlike \texttt{MixFlow} and general MCMC methods, sample
estimates of target summaries obtained from \texttt{NF} are typically not
asymptotically exact, as the sample quality is fundamentally limited by the choice of variational 
family and how well the flow is trained. 

\begin{figure*}[t!]
    \centering 
\begin{subfigure}[b]{0.75\textwidth} 
    \scalebox{1}{\includegraphics[width=\textwidth]{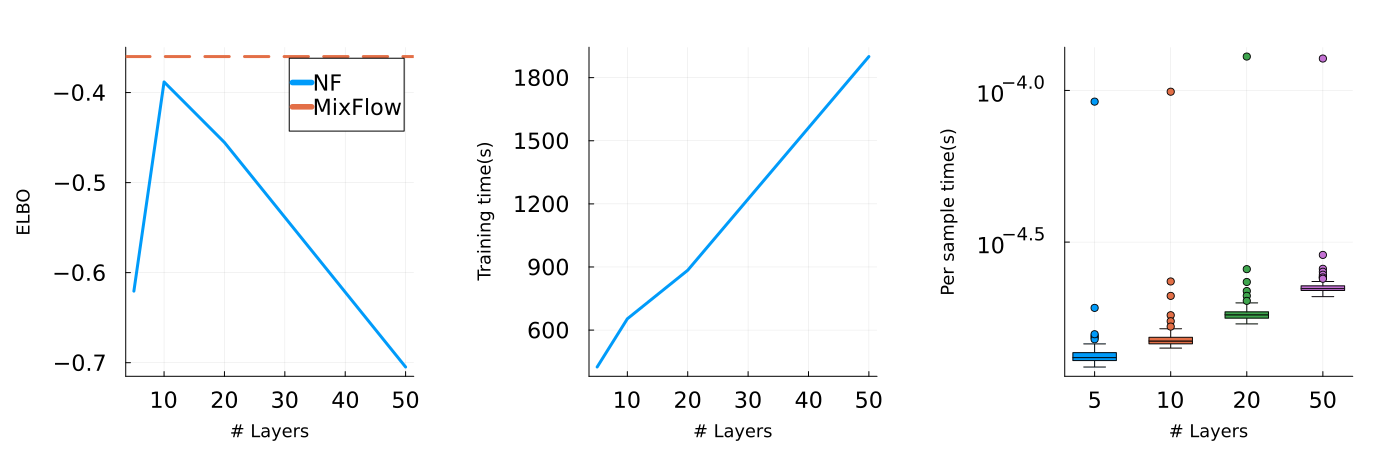}}
    \caption{(a) Banana distribution} \label{fig:banana_nf_layer}
\end{subfigure}
\hfill
\centering 
\begin{subfigure}[b]{0.75\textwidth} 
    \scalebox{1}{\includegraphics[width=\textwidth]{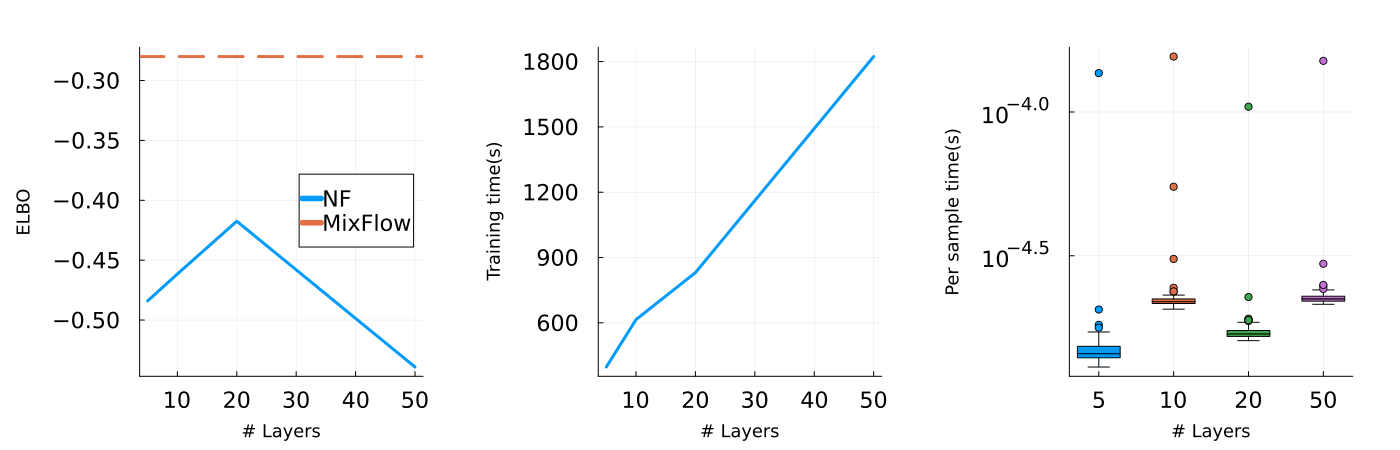}}
    \caption{(b) Neal's funnel} \label{fig:funnel_nf_layer}
\end{subfigure}
\hfill
\centering
\begin{subfigure}[b]{0.75\textwidth}
    \scalebox{1}{\includegraphics[width=\textwidth]{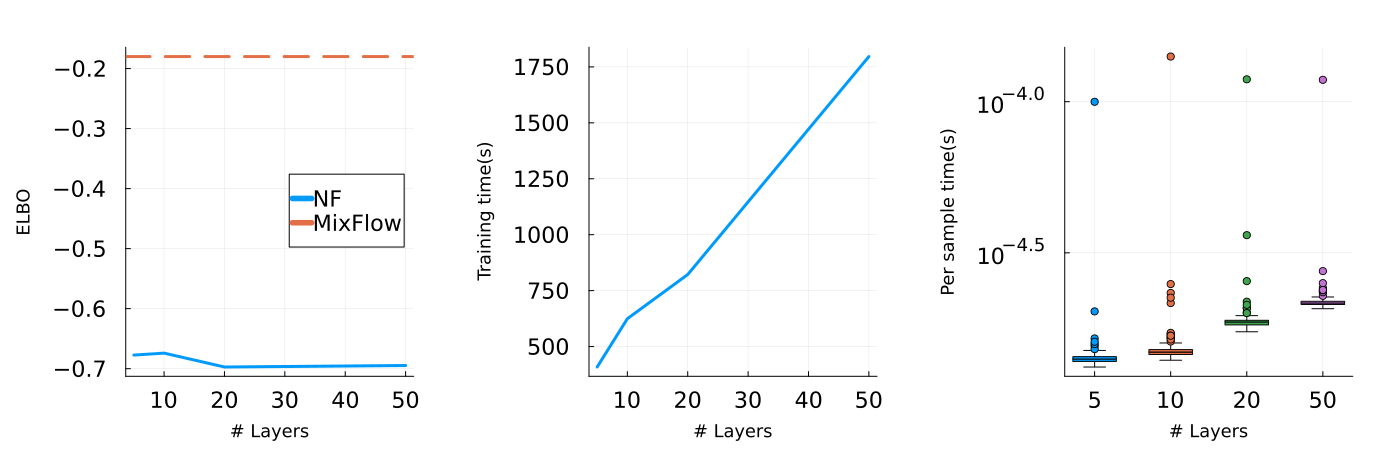}}
    \caption{(c) Cross distribution}\label{fig:cross_nf_layer}
\end{subfigure}
\hfill
    \centering 
\begin{subfigure}[b]{0.75\textwidth} 
    \scalebox{1}{\includegraphics[width=\textwidth]{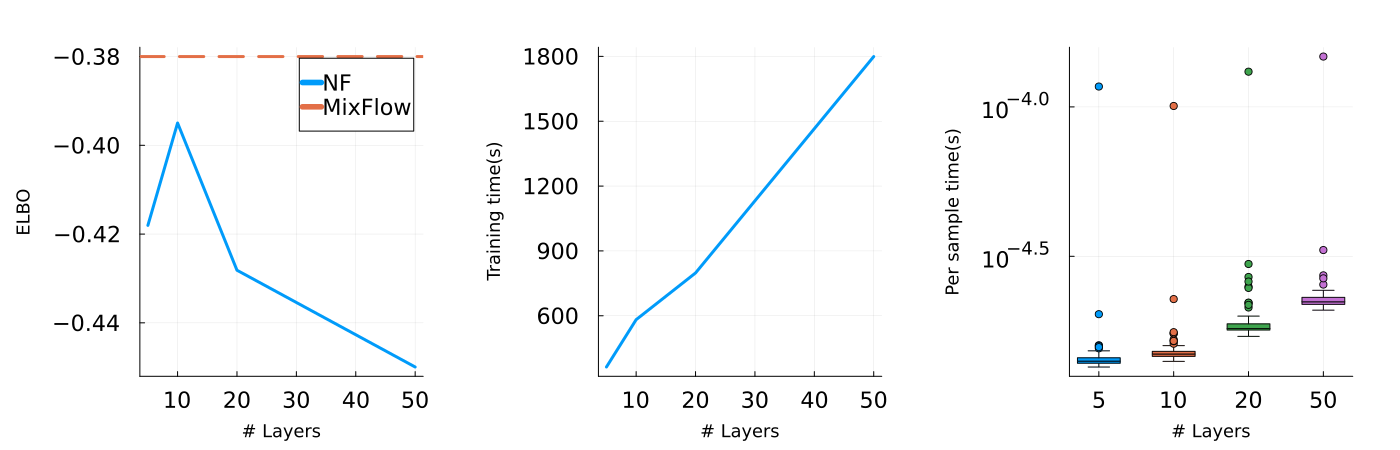}}
    \caption{(d) Warped Gaussian}\label{fig:warp_nf_layer}
\end{subfigure}

\caption{Additional \texttt{NF} results with number of planar layers ($5, 10, 20, 50$). 
First column shows the ELBO value. 
Each ELBO is estimated using $2000$ \iid samples from trained flow. The second and third columns 
correspond to the training time and per-sample time ($100$ trials) of \texttt{NF} given increasing planar layers. }
\label{fig:2d_more_nf}
\end{figure*}

Finally, we provide an additional set of results for \texttt{NF}
(\cref{fig:2d_more_nf}), examining its performance as we increase the number of
planar layers ($5, 10, 20, 50$). All settings for \texttt{NF} are identical to the above, except that we increase the optimization iteration to $500,000$ to
ensure the convergence of flows with increased numbers of layers. As
demonstrated in the second and third column of \cref{fig:2d_more_nf}, both
training time and sampling time scale with the number of layers roughly
linearly. Although a trained \texttt{NF} is still generally faster in sample
generation (see also \cref{fig:time_per_sample}), for these synthetic examples
with $4$-dimensional joint target distributions, training time can take up to
$30$ minutes.  More importantly, the corresponding target approximations of
\texttt{NF} are still not as good as those of \texttt{MixFlow} in all four
examples, even when we increase the number of layers to $50$, which corresponds
to optimizing $450$ parameters from the flow.  One may also notice from
\cref{fig:2d_more_nf} that a more complex normalizing flow does not necessarily
lead to better performance. This is essentially because the (usually
non-convex) KL optimization problem of standard \texttt{NF} becomes more
complex to solve as the flow becomes more flexible. As a result, even though
theoretically, \texttt{NF} becomes more expressive with more layers, there is
no guarantee on how well it approximates the target distribution.  In contrast,
\texttt{MixFlow} is optimization-free and is asymptotically exact---with a
proper choice of hyper-parameter, more computation typically leads to better
performance (\cref{fig:2d_elbo}).  With the $30$ minutes training time of
\texttt{NF}, \texttt{MixFlow iid} can generate $18,000$ \iid samples and
\texttt{MixFlow single} can generate over $10$ million samples, both of which are
more than sufficient for most estimations under these target distributions.

\captionsetup[subfigure]{labelformat=empty}

\subsection{Real data examples} \label{apdx:realexpt}
All settings for \texttt{MixFlow}, \texttt{NUTS}, \texttt{NEO} and \texttt{HMC} are identical to those in synthetic examples. All \texttt{NF}s are trained using ADAM for $200{,}000$ iterations with initial step size set to be $0.001$. 
We remark that comparison \texttt{NEO} is not included for high dimensional sparse
regression example---the most challenging example---due to its high consumption of RAM, which hits the ceiling of our computation resources.
Overall we observe similar phenomenons as for synthetic examples \cref{apdx:2d}.
Additionally, we include comparison to \texttt{UHA} on real data examples.
The tuning procedure of \texttt{UHA} involves architectural search (i.e.,
leapfrog steps between
refreshments, and number of refreshments), and hyperparameter optimization (i.e.,
leapfrog step size, and tempering path).
We choose among the combination of several fixed architectural settings of
leapfrog steps between refreshments ($L = 10, 20, 50$), and number of
refreshment ($N = 5, 10$), and optimize hyperparameters using ADAM 
for $20{,}000$ iterations (each gradient estimate is based on 5 independent
samples); each setting is repeated for $5$ times. \cref{tab:uha} presents
median, IQR of the resulting ELBOs of \texttt{UHA} under different architectural
settings. We pick the best settings for \texttt{UHA} based on the ELBO
performance and compare \texttt{UHA} under the selected settings with our
method in terms of target approximation quality and marginal sample quality.
Both ELBO and KSD of \texttt{UHA} are estimated using $5{,}000$ samples. 

\begin{figure}[t!]
\centering 
\begin{subfigure}[b]{.3\columnwidth} 
    \scalebox{1}{\includegraphics[width=\columnwidth]{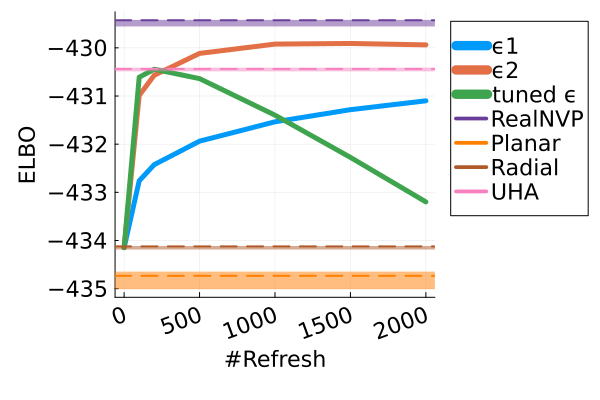}}
    \caption{(a) Linear regression} \label{fig:linreg_elbo_full}
\end{subfigure}
\hfill
\centering 
\begin{subfigure}[b]{.3\columnwidth} 
    \scalebox{1}{\includegraphics[width=\columnwidth]{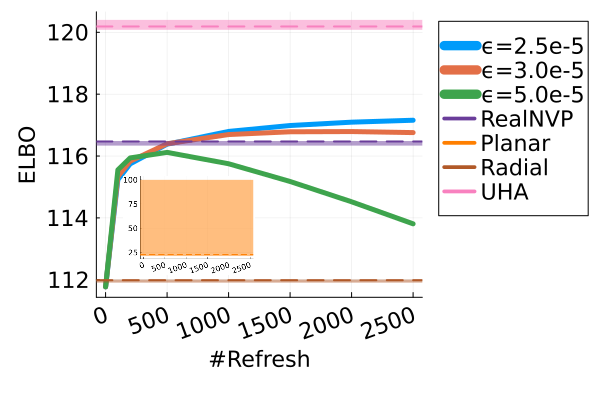}}
    \caption{(b) Linear regression (heavy tail)} \label{fig:lrh_elbo_full}
\end{subfigure}
\hfill
\centering 
\begin{subfigure}[b]{.3\columnwidth} 
    \scalebox{1}{\includegraphics[width=\columnwidth]{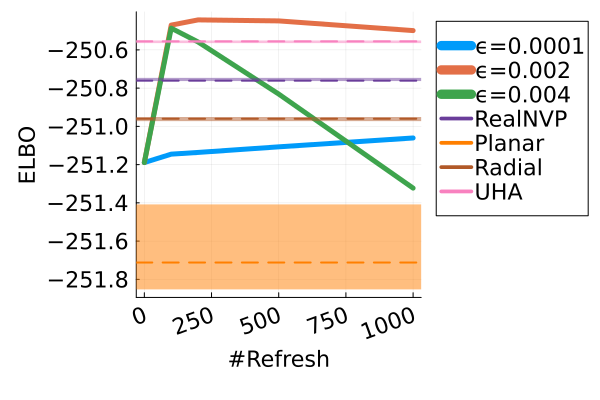}}
    \caption{(c) Logistic regression}
\end{subfigure}
\hfill
    \centering 
\begin{subfigure}[b]{.3\columnwidth} 
    \scalebox{1}{\includegraphics[width=\columnwidth]{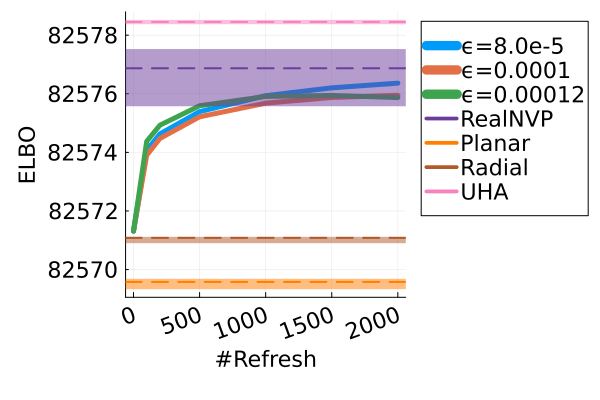}}
    \caption{(d) Poisson regression}
\end{subfigure}
\hfill
\centering 
\begin{subfigure}[b]{.3\columnwidth} 
    \scalebox{1}{\includegraphics[width=\columnwidth]{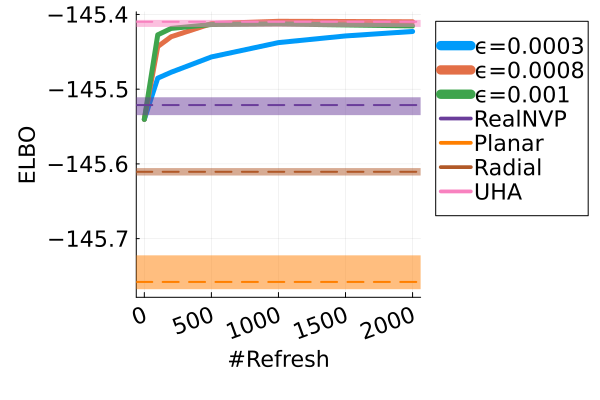}}
    \caption{(e) Student-t regression}
\end{subfigure}
\hfill
\begin{subfigure}[b]{.3\columnwidth} 
    \scalebox{1}{\includegraphics[width=\columnwidth]{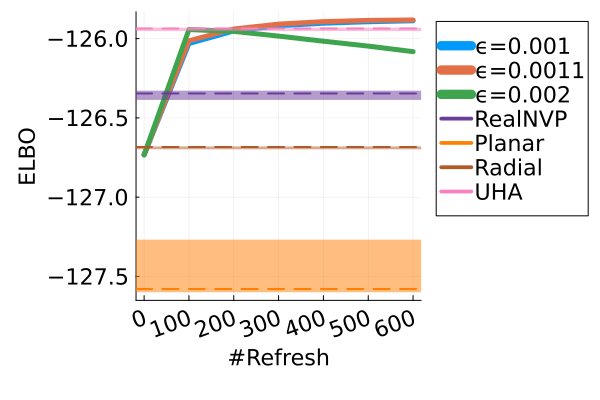}}
    \caption{(f) Sparse regression}
\end{subfigure}
\hfill
\centering 
\begin{subfigure}[b]{.3\columnwidth} 
    \scalebox{1}{\includegraphics[width=\columnwidth]{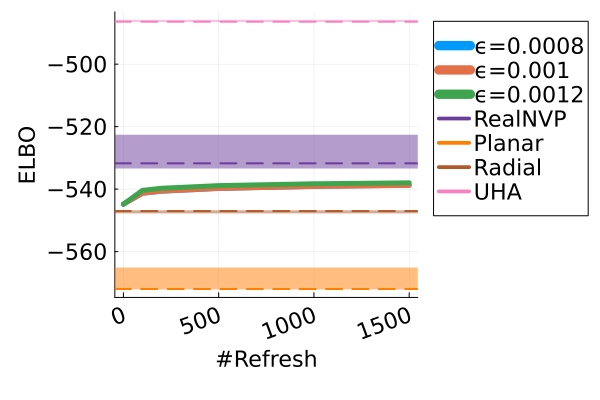}}
    \caption{(g) Sparse regression (high dim)}
\end{subfigure}
\hfill
\caption{ELBO comparison against \texttt{NF}s for real examples (except for linear regression, which is displayed in \cref{fig:linreg_elbo}. Each \texttt{NF} method is tuned under various settings, and only the best one is present for each example. 
Each figure also shows the effect of step sizes on \texttt{MixFlow}; overly large or small step sizes influence the 
performance of \texttt{MixFlow} negatively.  
Lines indicate the median, and error regions indicate $25^{th}$ to $75^{th}$ percentile from $5$ runs.}
\label{fig:real_elbo_tune}
\end{figure}

\cref{fig:real_elbo_tune} offers a comparison of the ELBO performance between
\texttt{UHA} and \texttt{MixFlow}. While \texttt{MixFlow} demonstrates similar
or superior ELBO performance in the linear regression, logistic regression,
student-t regression, and sparse regression problems, it is outperformed by
\texttt{UHA} in the remaining three real data examples. The higher ELBO
of \texttt{UHA} in these examples
might be due to its incorporation of annealing, which facilitates  
exploration of complex target distributions. We leave studying the use of annealing
in \texttt{MixFlows} to future work.

\begin{figure*}[t!]
\centering 
\begin{subfigure}[b]{.24\columnwidth} 
    \scalebox{1}{\includegraphics[width=\columnwidth]{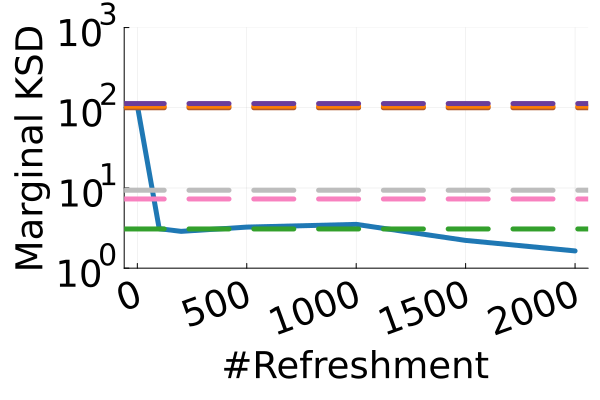}}
    \caption{(a) linear regression \label{fig:linreg_elbo_neo}}
\end{subfigure}
\hfill
\centering 
\begin{subfigure}[b]{.24\columnwidth} 
    \scalebox{1}{\includegraphics[width=\columnwidth]{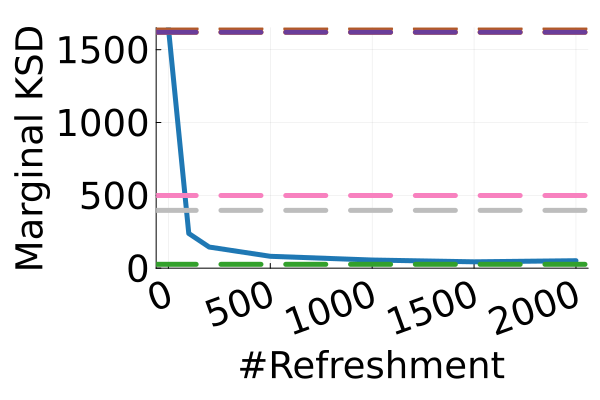}}
    \caption{(b) linear regression (heavy) \label{fig:lrh_elbo_neo}}
\end{subfigure}
\hfill
\centering 
\begin{subfigure}[b]{.24\columnwidth} 
    \scalebox{1}{\includegraphics[width=\columnwidth]{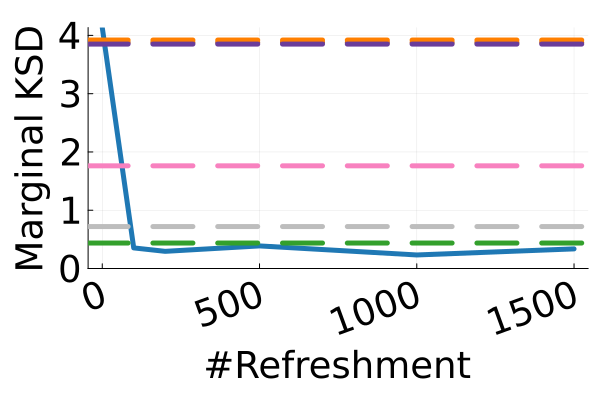}}
    \caption{(c) logistic regression \label{fig:logreg_elbo_neo}}
\end{subfigure}
\hfill
\centering
\begin{subfigure}[b]{.24\columnwidth} 
    \scalebox{1}{\includegraphics[width=\columnwidth]{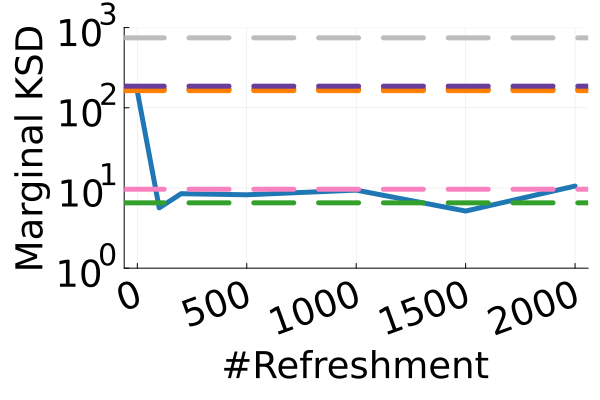}}
    \caption{(d) poisson regression \label{fig:poiss_elbo_neo}}
\end{subfigure}
\hfill
\centering
\begin{subfigure}[b]{.24\columnwidth} 
    \scalebox{1}{\includegraphics[width=\columnwidth]{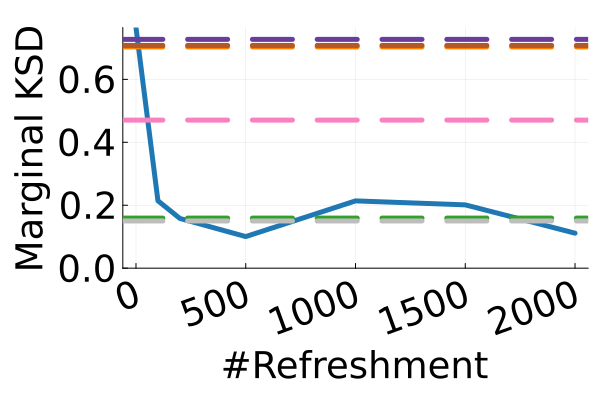}}
    \caption{(e) student t regression \label{fig:tr_elbo_neo}}
\end{subfigure}
\hfill
\centering
\begin{subfigure}[b]{.24\columnwidth} 
    \scalebox{1}{\includegraphics[width=\columnwidth]{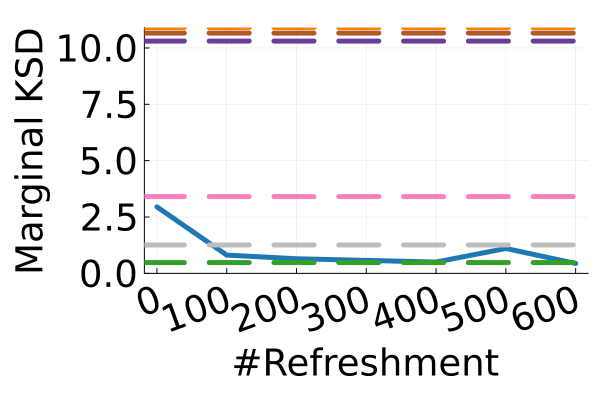}}
    \caption{(f) sparse regression \label{fig:sp_elbo_neo}}
\end{subfigure}
\hfill
\centering
\begin{subfigure}[b]{.24\columnwidth} 
    \scalebox{1}{\includegraphics[width=\columnwidth]{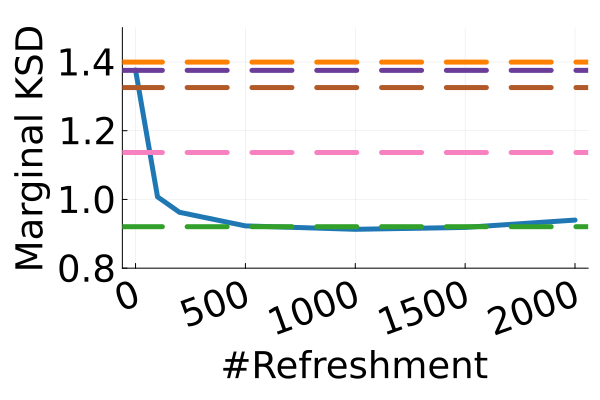}}
    \caption{(g) sparse regression (high dim) \label{fig:sp_big_elbo_neo}}
\end{subfigure}
\hfill
\centering 
\begin{subfigure}[b]{.24\columnwidth} 
	\centering
    \scalebox{1}{\includegraphics[width=0.5\columnwidth]{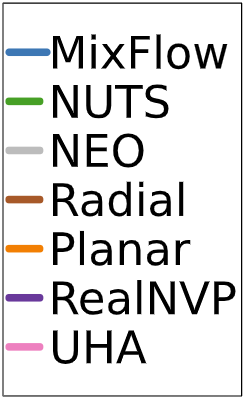}}
    \caption{}
\end{subfigure}
\hfill
\caption{KSD comparison for real data examples. }
\label{fig:real_metric_neo}
\end{figure*}

However, it is essential to note that for variational methods that augment the
original target space, a higher ELBO does not necessarily equate to
improved marginal sample quality. As demonstrated in \cref{fig:real_metric_neo},
the marginal sample quality of \texttt{UHA}, as evaluated by marginal KSD, is
worse than that of \texttt{MixFlow} and the two Monte Carlo methods (\texttt{NUTS}
and \texttt{NEO}), although it surpasses that of parametric flows (e.g.
RealNVP).

It is also worth noting that tuning augmentation methods like \texttt{UHA} can be
very expensive. For instance, in the Student-t regression problem,
\texttt{UHA} required 20 minutes ( $1223.8$ seconds) to run $20{,}000$ optimization
steps, with each gradient estimated using 5 Monte Carlo samples. This is
approximately 40 times slower than RealNVP, which required around 10 minutes for
$200{,}000$ optimization steps with stochastic gradients based on 10 Monte Carlo
samples. Conversely, evaluating one ELBO curve (averaged over $1{,}000$ independent
trajectories) for \texttt{MixFlow} across six
different flow lengths (i.e., flow length = 100, 200, 500, 1000, 1500, 2000)
only took 8 minutes, while adapting \texttt{NUTS} with $20{,}000$ iterations was
completed within a few seconds.

\subsubsection{Bayesian linear regression}\label{apdx:linear}
We consider two Bayesian linear regression problems, with both a
standard normal prior and a heavy tail prior using two sets of real data. 
The two statistical models take the following form:
\[
	\label{eq:linreg_model}
& \log\sigma^2 \distiid \distNorm(0,1),\quad
 y_j \given \beta, \sigma^2 \distind \distNorm\left(x_j^T\beta, \sigma^2\right)\\
& \textbf{Normal:} \beta \distiid \distNorm(0, 1) \quad  \textbf{Cauchy:} \beta \distiid \distCauchy(0,1) 
\]
where $y_j$ is the response and $x_j\in\reals^p$ is the feature vector for data point $j$.
For linear regression problem with a normal prior, we use the Boston housing prices
dataset \citep{Harrison78}. Dataset available in the \texttt{MLDatasets} Julia package at \url{https://github.com/JuliaML/MLDatasets.jl}.
containing $J=506$ suburbs/towns in the Boston area; the goal is to use suburb information to predict the median house price. We standardize all features and responses. 
For linear regression with a heavy-tail
prior, we use the communities and crime dataset \citep{community95}, 
available at \url{http://archive.ics.uci.edu/ml/datasets/Communities+and+Crime}.
The original dataset contains $122$ attributes that potentially connect to crime; the goal is to
predict per-capita violent crimes using the information of the community, such as
the median family income, per capita number of poluce officers, and etc.
For the data preprocessing, we drop observations with missing values, and using Principle component analysis for feature dimension reduction;
we selected $50$ principal components with leading eigenvalues.
The posterior dimensions of the two linear regression inference problems are
$15$ and $52$, repsectively.

\subsubsection{Bayesian generalized linear regression}  \label{apdx:genlinear}

We then consider two Bayesian generalized
linear regression problems---a hierachical logistic regression and a poisson
regression:
\[
\textbf{Logis. Reg.: }&
\alpha \sim \distGam(1, 0.01), \beta \mid \alpha  \sim \distNorm(0, \alpha^{-1} I), \\
& y_{j} \given \beta \distind \distBern\left(\frac{1}{1+e^{-x_{j}^{T} \beta}}\right),\\ 
\label{eq:logreg_model}
\textbf{Poiss. Reg.: } 
& \beta \sim \distNorm(0, I) , \\
& y_{j} \given \beta \distind \distPoiss\left(\log\left( 1+e^{-x_{j}^{T} \beta}\right)\right),
\label{eq:poireg_model}
\]
For logistic regression,
we use a bank marketing dataset \citep{Moro14} downsampled to $J=400$ data
points. Original dataset is  
available at \url{https://archive.ics.uci.edu/ml/datasets/bank+marketing}. 
the goal is to use client information to predict whether they subscribe to a term deposit. 
We include 8 features from the bank marketing
dataset \citep{Moro14}: client age, marital status, balance, housing loan status,
duration of last contact, number of contacts during campaign, number of days
since last contact, and number of contacts before the current campaign. For each
of the binary variables (marital status and housing loan status), all unknown
entries are removed. All features of the dataset are also standardized.
Hence the posterior dimension of the logistic regression problem is $9$ and the overall $(x,\rho,u)$ state
dimension of the logistic regression inference problems are 19.

For Poisson regression problem, we use an airport delays dataset with $15$
features and $J = 500$ data points (subsampled), resulting a $16$-dimensional
posterior distribution.
The airport delays dataset was constructed using flight delay data
  from \url{http://stat-computing.org/dataexpo/2009/the-data.html} and historical weather information
from \url{https://www.wunderground.com/history/}. 
relating daily weather information to the number of flights leaving an airport with a delay of more
than 15 minutes. 
All features are standardized as well.

\subsubsection{Bayesian Student-t regression}
We also consider a Bayesian Student-t regression problem, of which the
posterior distribution is heavy-tail. The Student-t regression model is
as follows:
\[
y_i \mid X_i, \beta \sim \distT_5(X_i^T \beta, 1), \quad \beta \distiid \distCauchy(0, 1).   
\]
In this example, we use the creatinine dataset \citep{liu1995ml}, containing a clinical trial on $34$ male patients with $3$ covariates. 
Original dataset is available in
\url{https://github.com/faosorios/heavy/blob/master/data/creatinine.rda}.
The 3 covariates consist of body weight in kg(WT), serum creatininte
concentration (SC), and age in years. The goal is to predict the endogenous
cretinine clearance (CR) using these covariates.
We apply the data transformation recommended by \citet{liu1995ml} by transferring response 
into $\log(\text{CR})$, and transferring covariats into $\log(\text{WT})$,
$\log(\text{SC})$, $\log(140- \text{age})$.

\subsubsection{Bayesian sparse regression}
Finally, we compare the methods on the Bayesian sparse regression problem applied
to two datasets: a prostate cancer dataset containing $9$ covariates and $97$
observations, and a superconductivity dataset \citep{hamidieh2018data}, containing $83$ features and
$100$ observations (subsampled). 
The prostate cancer dataset is available at
\url{https://hastie.su.domains/ElemStatLearn/datasets/prostate.data}.
The superconductivity dataset is available at
\url{https://archive.ics.uci.edu/ml/datasets/superconductivty+data}.
The model is as follows:
\[
&\log\sigma^2 \distiid \distNorm(0,1), \beta_i \distiid \frac{1}{2} \distNorm(0, \tau_1^2) + \frac{1}{2} \distNorm(0, \tau_2^2),\\
& y_j \given \beta, \sigma^2 \distind \distNorm\left(x_j^T\beta, \sigma^2\right)
	\label{eq:spreg_model}
\]
For both two datasets, we set $\tau_1 = 0.1, \tau_2 = 10$. 
The resulting posterior dimension for both datasets are $10$ and $84$ respectively.
When data information is weak, the posterior distribution in this model typically contains multiple modes \citep{Xu22}.
We standardize the covariates during the preprocessing procedure for both datasets.

\subsubsection{Additional experiment for real data examples} \label{apdx:addrealexpts}

\captionsetup[subfigure]{labelformat=empty}

\captionsetup[subfigure]{labelformat=empty}

\captionsetup[subfigure]{labelformat=empty}
\begin{figure}[t!]
\centering 
\begin{subfigure}[b]{.3\columnwidth} 
    \scalebox{1}{\includegraphics[width=\columnwidth]{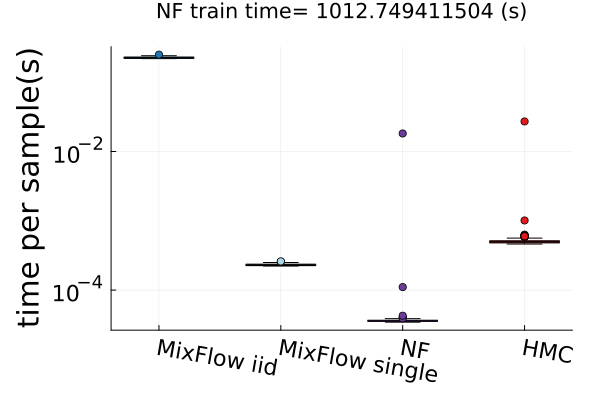}}
    \scalebox{1}{\includegraphics[width=\columnwidth]{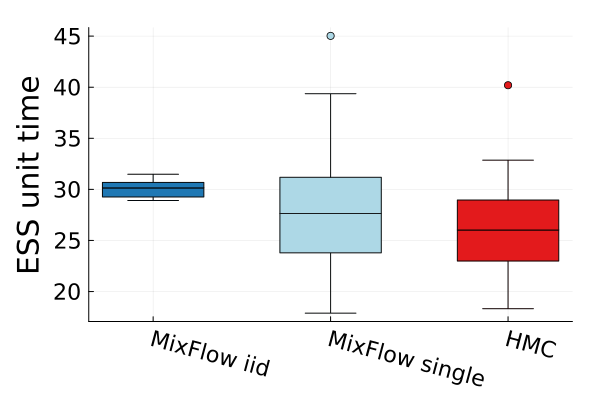}}
    \caption{(a) Linear regression} \label{fig:linreg_time}
\end{subfigure}
\hfill
\begin{subfigure}[b]{.3\columnwidth} 
    \scalebox{1}{\includegraphics[width=\columnwidth]{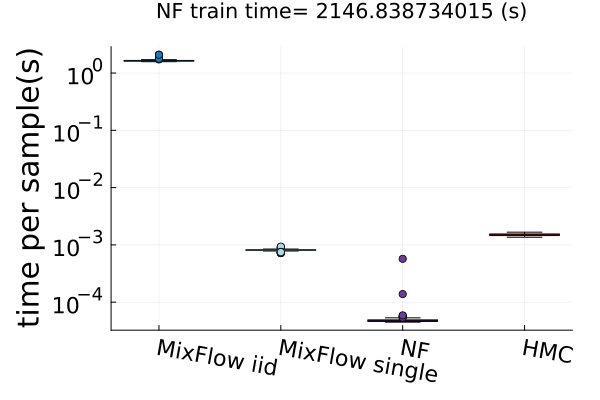}}
    \scalebox{1}{\includegraphics[width=\columnwidth]{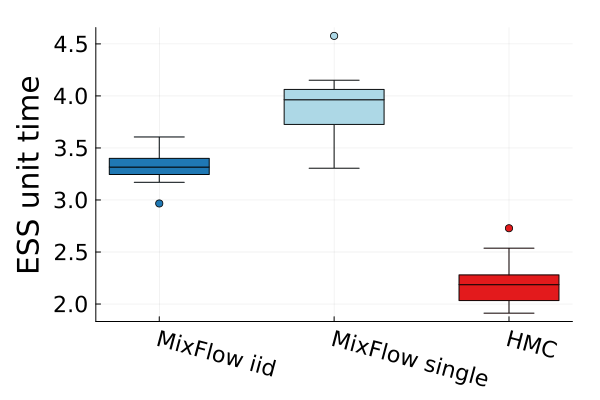}}
    \caption{(b) Linear regression (heavy tail)}\label{fig:lrh_time}
\end{subfigure}
\hfill
\centering 
\begin{subfigure}[b]{.3\columnwidth} 
    \scalebox{1}{\includegraphics[width=\columnwidth]{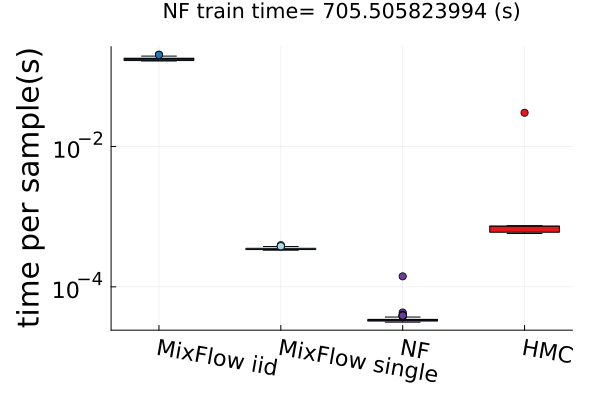}}
    \scalebox{1}{\includegraphics[width=\columnwidth]{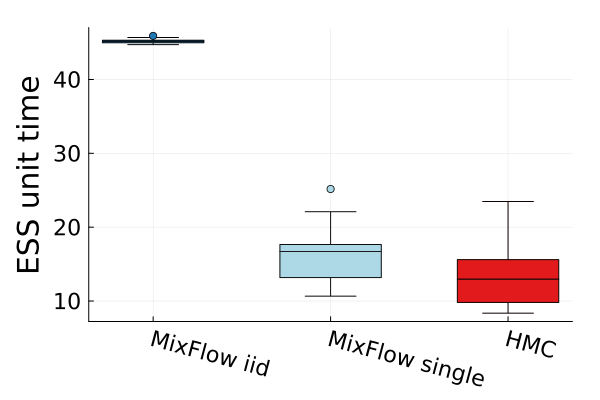}}
    \caption{(c) Logistic regression}\label{fig:logreg_time}
\end{subfigure}
\hfill
\begin{subfigure}[b]{.3\columnwidth} 
    \scalebox{1}{\includegraphics[width=\columnwidth]{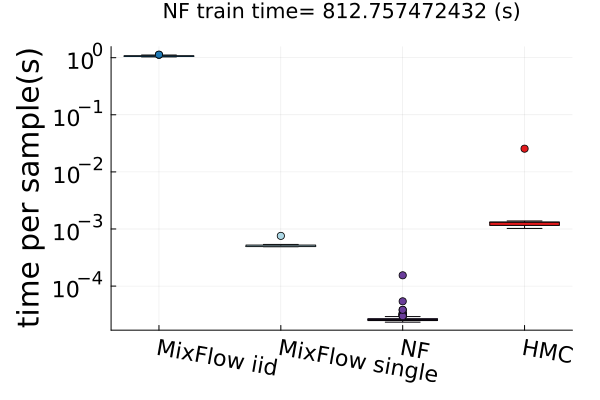}}
    \scalebox{1}{\includegraphics[width=\columnwidth]{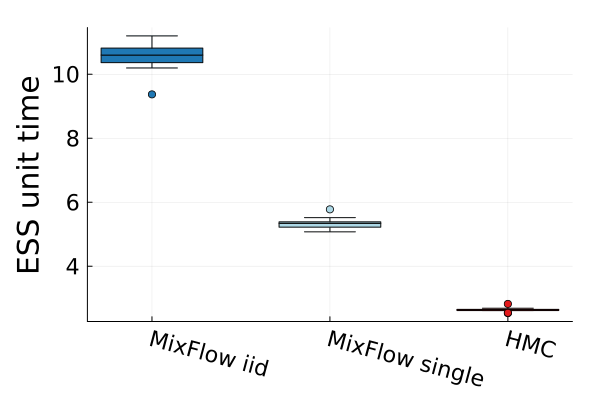}}
    \caption{(d) Poisson regression}\label{fig:poiss_time}
\end{subfigure}
\begin{subfigure}[b]{.3\columnwidth} 
    \scalebox{1}{\includegraphics[width=\columnwidth]{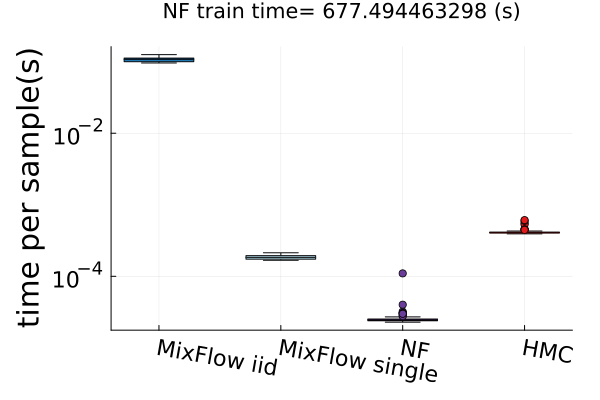}}
    \scalebox{1}{\includegraphics[width=\columnwidth]{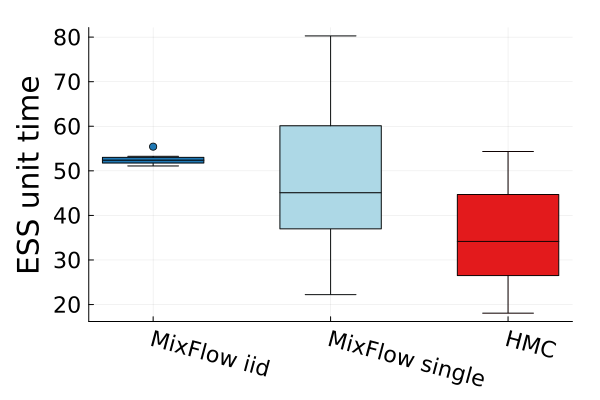}}
    \caption{(e) Sparse regression}\label{fig:sp_time}
\end{subfigure}
\hfill
\caption{Time results (sampling time comparison to \texttt{NF, NUTS, NEO}, and \texttt{HMC}, and per second ESS comparison to \texttt{HMC}; each repeated 100 trials) for two linear regression problems (\cref{fig:linreg_time,fig:lrh_time}), two generalized linear regression problems (\cref{fig:logreg_time,fig:poiss_time}), and one sparse regression problem (\cref{fig:sp_time}).  }
\label{fig:real_more_timing}
\end{figure}

\captionsetup[subfigure]{labelformat=empty}
\begin{figure}[t!]
    \centering 
\begin{subfigure}[b]{.33\columnwidth} 
    \scalebox{1}{\includegraphics[width=\columnwidth]{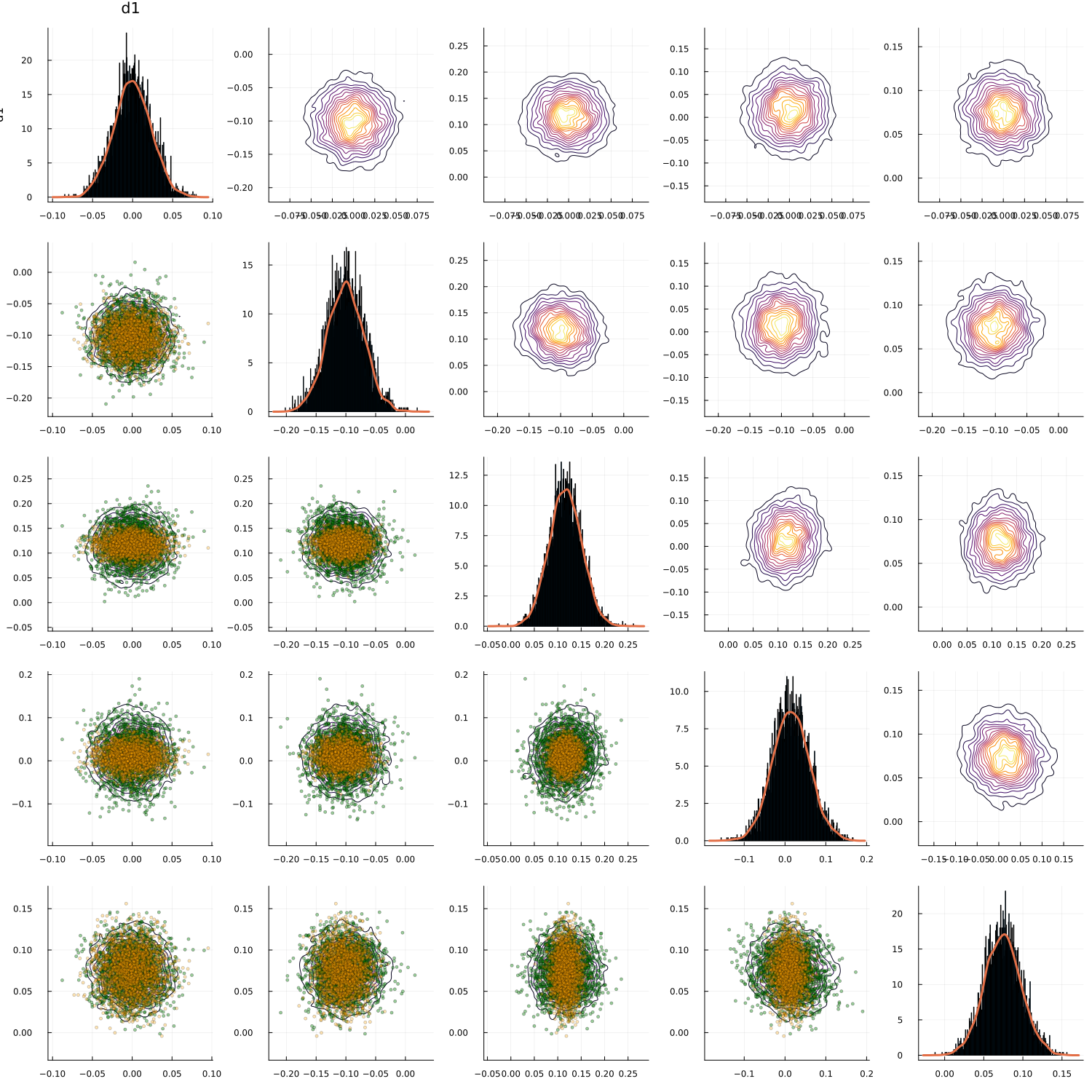}}
    \caption{(a) Linear reg.: first 5 dimensions}
\end{subfigure}
\hfill
\centering 
\begin{subfigure}[b]{.33\columnwidth} 
    \scalebox{1}{\includegraphics[width=\columnwidth]{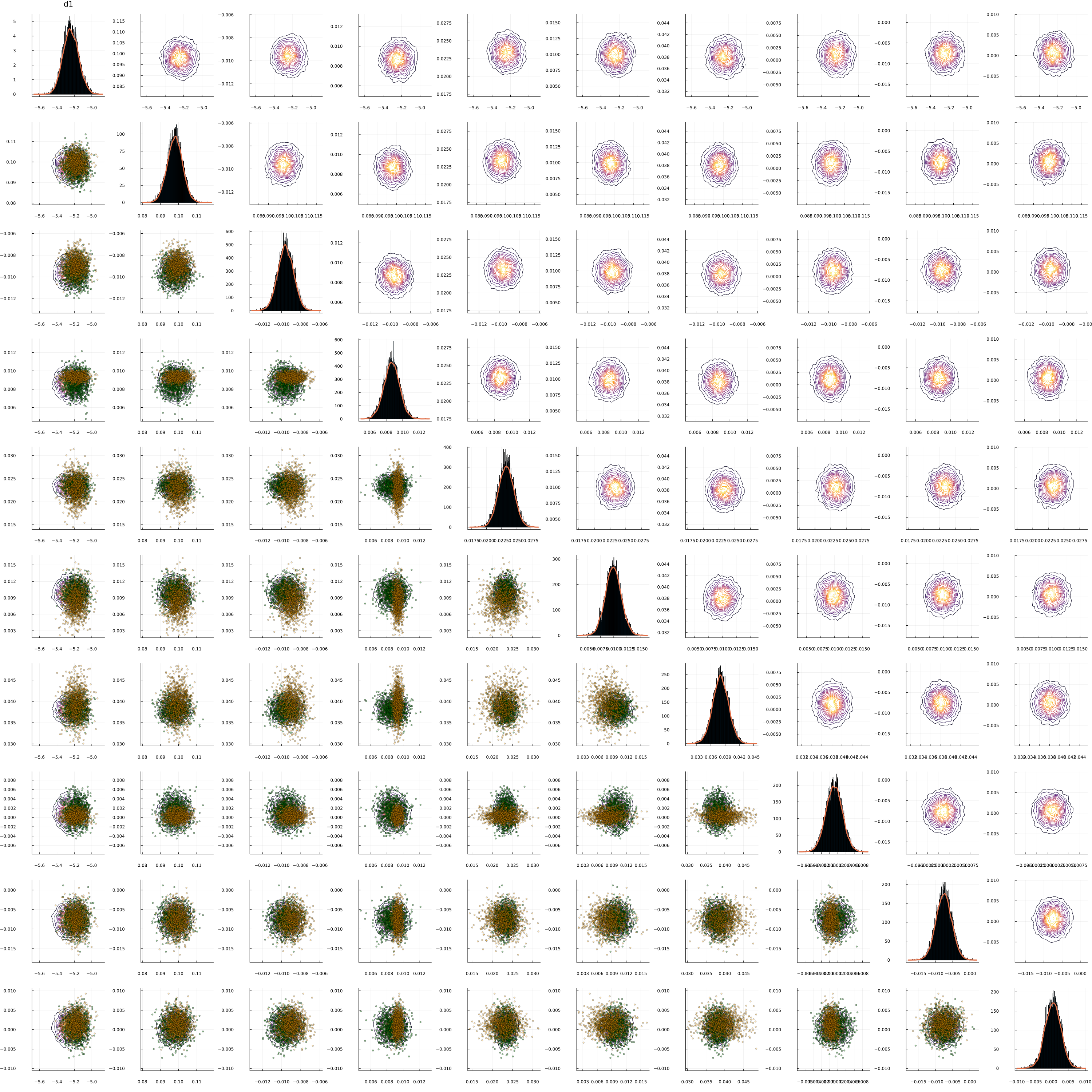}}
    \caption{(b) Linear reg. (Cauchy): first 10 dimensions}
\end{subfigure}
\hfill
\begin{subfigure}[b]{.33\columnwidth} 
    \scalebox{1}{\includegraphics[width=\columnwidth]{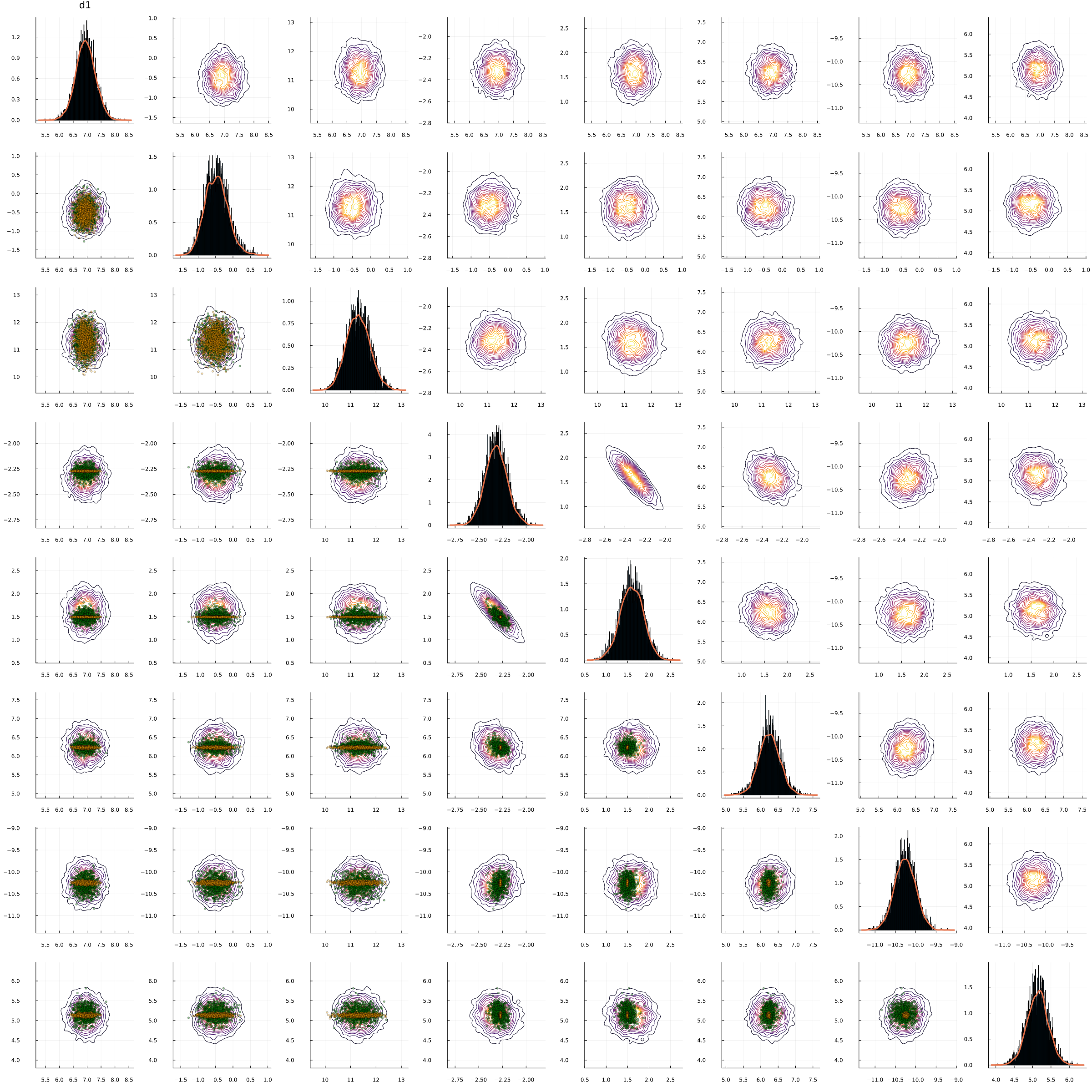}}
    \caption{(c) Poisson reg.: all 8 dimensions}
\end{subfigure}
\hfill
\centering 
\begin{subfigure}[b]{.8\columnwidth} 
    \scalebox{1}{\includegraphics[width=\columnwidth]{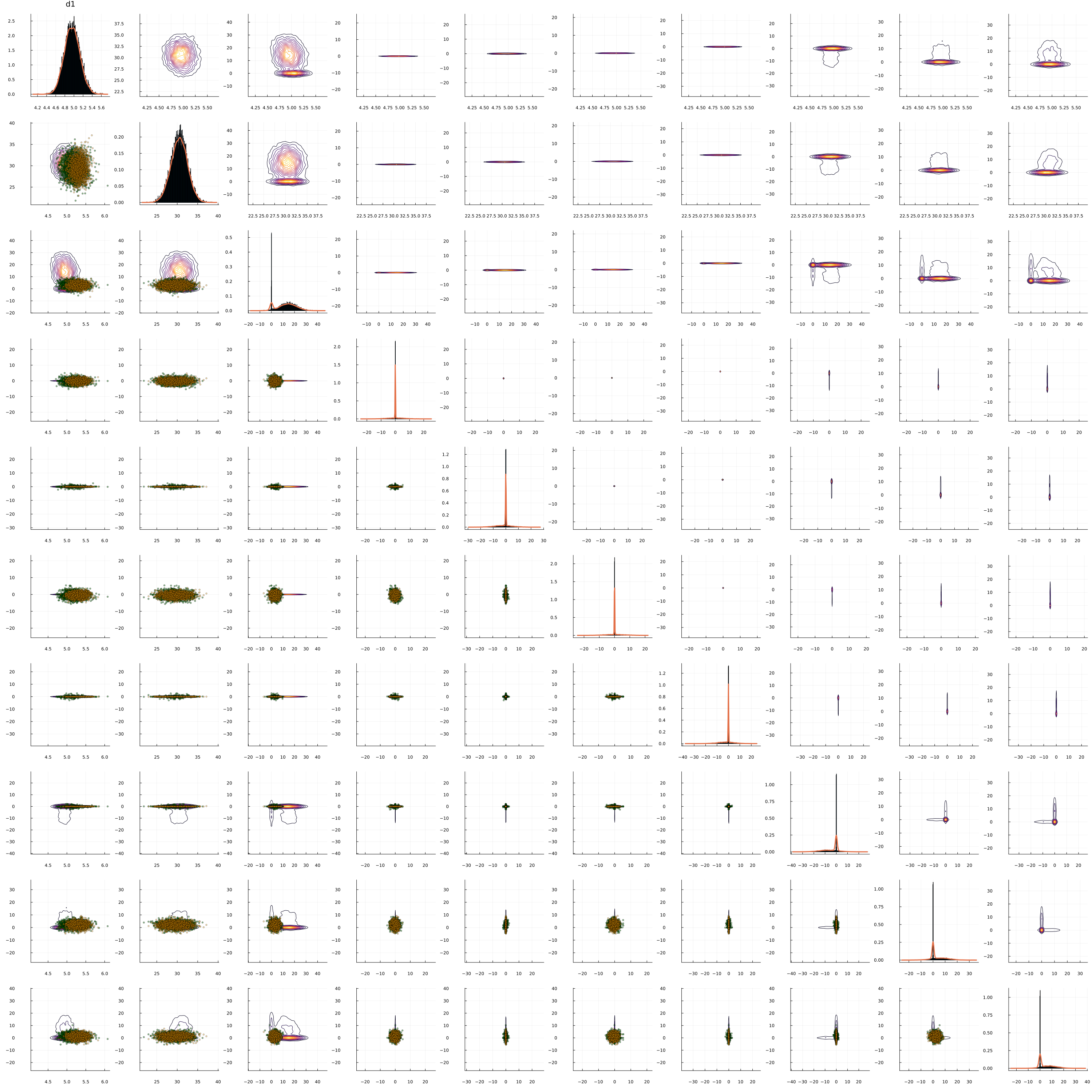}}
    \caption{(d) Sparse reg. (high dim): first 10 dimensions}
\end{subfigure}
\caption{Sample quality visualization of $2{,}000$ \iid samples draw from each of \texttt{MixFlow} (green scatters) and \texttt{NF} (orange scatters) on 4 real data examples: linear regression, linear regression with heavy tail prior, poisson regression, and high-dimensional sparse regression. Pairwise kernel density estimation (KDE) is based on $20{,}000$ \texttt{NUTS} samples, which is initialized with the Gaussian mean of the mean-field Gaussian approximation to posterior distribution, uses $20{,}000$ steps for adaptation (targeting at an average acceptance rate $0.7$). \texttt{NF} is chosen to be the same one as compared in \cref{fig:real_timing,fig:real_more_timing}.}
\label{fig:post_vis}
\end{figure}

\begin{table}[]
\caption{Comparison of ELBO between \texttt{MixFlow} and \texttt{RealNVP}
with different
number of layers on real data example. Each setting of \texttt{RealNVP} is
run in $5$ trials. ELBOs of \texttt{MixFlow} are estimated using $1000$
independent trajectories and ELBOs of \texttt{RealNVP} are estimated using
$2000$ independent samples.} 
\label{tab:realnvp}
\centering
\begin{tabular}{ccccccccccccccccccccc}
                    & \multicolumn{5}{c}{\texttt{MixFlow}}                                                   \\
                    \hline\\
                    & \textbf{STEPSIZE}      & \textbf{\#LEAPFROG} & \textbf{\#REFRESHMENT} & \textbf{ELBO}                 & \textbf{\#SAMPLE}  \\
                    \hline\\
Lin Reg             & 0.0005   & 30         & 2000          & -429.98              & 1000       \\
Lin Reg Heavy       & 0.000025 & 50         & 2500          & 117.1                & 1000     \\
Log Reg             & 0.002    & 50         & 1500          & -250.5               & 1000      \\
Poiss Reg           & 0.0001   & 50         & 2000          & 82576.9              & 1000     \\
Student t Reg       & 0.0008   & 80         & 2000          & -145.41              & 1000     \\
Sparse Reg          & 0.001    & 30         & 600           & -125.9               & 1000      \\
Sparse Reg High Dim & 0.0012   & 50         & 2000          & -538.36              & 2000     \\
                    & \multicolumn{5}{c}{\texttt{RealNVP}}                                              \\
                    \hline\\
                    & \textbf{\#LAYER}  & \textbf{\#HIDDEN}   & \textbf{MEDIAN}        & \textbf{IQR}                  & \textbf{\#NAN}     \\
                    \hline\\
Lin Reg             & 5        & 15         & -429.43       & (-429.56, -429.42)   & 0         \\
Lin Reg             & 10       & 15         & -429.41       & (-429.43, -429.36)   & 1         \\
Lin Reg Heavy       & 5        & 52         & 116.47        & (116.34, 116.49)     & 0        \\
Lin Reg Heavy       & 8        & 52         & 116.65        & (116.56, 116.73)     & 3        \\
Lin Reg Heavy       & 10       & 52         & 116.26        & (116.11,116.41)      & 3        \\
Log Reg             & 5        & 9          & -250.76       & (-250.76, -250.75)   & 3         \\
Log Reg             & 8        & 9          & -250.65       & (-250.75, -250.64)   & 2         \\
                    & \multicolumn{5}{c}{\texttt{RealNVP}}                                              \\
                    \hline\\
                    & \textbf{\#LAYER}  & \textbf{\#HIDDEN}   & \textbf{MEDIAN}        & \textbf{IQR}                  & \textbf{\#NAN}     \\
                    \hline\\
Poiss Reg           & 3        & 16         & 82576.87      & (82575.57, 82577.53) & 1        \\
Poiss Reg           & 5        & 16         & 82580.77      & (82580.13, 82583.72) & 2      \\
Poiss Reg           & 8        & 16         & N/A           & N/A                  & 5        \\
Student t Reg       & 5        & 4          & -145.52       & (-145.53, -145.51)   & 0         \\
Student t Reg       & 10       & 4          & -145.52       & (-145.52, -145.49)   & 0        \\
                    & \multicolumn{5}{c}{\texttt{RealNVP}}                                              \\
                    \hline\\
                    & \textbf{\#LAYER}  & \textbf{\#HIDDEN}   & \textbf{MEDIAN}        & \textbf{IQR}                  & \textbf{\#NAN}     \\
                    \hline\\
Sparse Reg          & 5        & 10         & -126.33       & (-126.36, -126.32)   & 0         \\
Sparse Reg          & 8        & 10         & -126.35       & (-126.36, -126.33)   & 1        \\
Sparse Reg High Dim & 5        & 20         & -531.75       & (-533.41, -522.60)   & 0        \\
Sparse Reg High Dim & 8        & 20         & N/A           & N/A                  & 5  
\end{tabular} 
\end{table}

\begin{table}[]
\caption{Comparison of ELBO between \texttt{MixFlow} and \texttt{PlanarFlow}
with different
number of layers on real data example. Each setting of \texttt{PlanarFlow} is
run in $5$ trials. ELBOs of \texttt{MixFlow} are estimated using $1000$
independent trajectories and ELBOs of \texttt{PlanarFlow} are estimated using
$2000$ independent samples.} 
\label{tab:planar}
\centering
\begin{tabular}{ccccccccccccccccccccc}
                    & \multicolumn{5}{c}{\texttt{MixFlow}}                                                   \\
                    \hline\\
                    & \textbf{STEPSIZE}      & \textbf{\#LEAPFROG} & \textbf{\#REFRESHMENT} & \textbf{ELBO}                 & \textbf{\#SAMPLE}  \\
                    \hline\\
Lin Reg             & 0.0005   & 30           & 2000                 & -429.98 & 1000      \\
Lin Reg Heavy       & 2.50E-05 & 50           & 2500                 & 117.1   & 1000   \\
Log Reg             & 0.002    & 50           & 1500                 & -250.5  & 1000      \\
Poiss Reg           & 0.0001   & 50           & 2000                 & 82576.9 & 1000    \\
Student t Reg       & 0.0008   & 80           & 2000                 & -145.41 & 1000    \\
Sparse Reg          & 0.001    & 30           & 600                  & -125.9  & 1000     \\
Sparse Reg High Dim & 0.0012   & 50           & 2000                 & -538.36 & 2000    \\
                    & \multicolumn{4}{c}{\texttt{PlanarFlow}}                                              \\
                    \hline\\
                    & \textbf{\#LAYER}  & \textbf{MEDIAN}        & \textbf{IQR}                  & \textbf{\#NAN}     \\
                    \hline\\
Lin Reg             & 5        & -434.73      & (-435.01, -434.65)   & 0       &      \\
Lin Reg             & 10       & -438.79      & (-439.25, -436.77)   & 0       &  \\
Lin Reg             & 20       & -442.4384675 & (-444.64, -440.93)   & 0       &  \\
Lin Reg Heavy       & 5        & 79.65        & (-75.92, 99.10)      & 0       &     \\
Lin Reg Heavy       & 10       & 23.13        & ( 21.29, 100.10)     & 0       &   \\
Lin Reg Heavy       & 20       & -879.1909019 & (-1641.19, -392.80)  & 0       &  \\
Log Reg             & 5        & -251.71      & (-251.85, -251.41)   & 0       &     \\
Log Reg             & 10       & -251.8       & (-252.03, -251.80)   & 0       &    \\
Log Reg             & 20       & -252.0937812 & (-252.20, -251.93)   & 0       &     \\
                    & \multicolumn{4}{c}{\texttt{PlanarFlow}}                                              \\
                    \hline\\
                    & \textbf{\#LAYER}  & \textbf{MEDIAN}        & \textbf{IQR}                  & \textbf{\#NAN}     \\
                    \hline\\
Poiss Reg           & 5        & 82569.58     & (82569.33, 82569.68) & 0       &   \\
Poiss Reg           & 10       & 82568.49     & (82568.12, 82569.57) & 0       &  \\
Poiss Reg           & 20       & 82566.40441  & (82552.57, 82568.05) & 2       &  \\
Student t Reg       & 5        & -145.69      & (-145.71, -145.64)   & 0       &     \\
Student t Reg       & 10       & -145.72      & (-145.73, -145.69)   & 0       &    \\
Student t Reg       & 20       & -145.7579931 & (-145.77, -145.72)   & 0       &  \\
                    & \multicolumn{4}{c}{\texttt{PlanarFlow}}                                              \\
                    \hline\\
                    & \textbf{\#LAYER}  & \textbf{MEDIAN}        & \textbf{IQR}                  & \textbf{\#NAN}     \\
                    \hline\\
Sparse Reg          & 5        & -127.58      & (-127.60, -127.42)   & 0       &     \\
Sparse Reg          & 10       & -127.75      & (-127.87, -127.68)   & 0       &    \\
Sparse Reg          & 20       & -127.6677415 & (-128.48, -127.65)   & 0       &  \\
Sparse Reg High Dim & 5        & -571.97      & (-572.09, -565.10)   & 0       &   \\
Sparse Reg High Dim & 10       & -571.7       & (-573.22, -565.55)   & 0       &   \\
Sparse Reg High Dim & 20       & -577.1338885 & (-580.30, -569.39)   & 0       & 
\end{tabular}
\end{table}

\begin{table}[]
\label{tab:radial}
\caption{Comparison of ELBO between \texttt{MixFlow} and \texttt{RadialFlow}
with different
number of layers on real data example. Each setting of \texttt{RadialFlow} is
run in $5$ trials. ELBOs of \texttt{MixFlow} are estimated using $1000$
independent trajectories and ELBOs of \texttt{RadialFlow} are estimated using
$2000$ independent samples.} 
\centering
\begin{tabular}{ccccccccccccccccccccc}
                    & \multicolumn{5}{c}{\texttt{MixFlow}}                                                   \\
                    \hline\\
                    & \textbf{STEPSIZE}      & \textbf{\#LEAPFROG} & \textbf{\#REFRESHMENT} & \textbf{ELBO}                 & \textbf{\#SAMPLE}  \\
                    \hline\\
Lin Reg             & 0.0005   & 30         & 2000                  & -429.98 & 1000     \\
Lin Reg Heavy       & 2.50E-05 & 50         & 2500                  & 117.1   & 1000  \\
Log Reg             & 0.002    & 50         & 1500                  & -250.5  & 1000      \\
Poiss Reg           & 0.0001   & 50         & 2000                  & 82576.9 & 1000    \\
Student t Reg       & 0.0008   & 80         & 2000                  & -145.41 & 1000   \\
Sparse Reg          & 0.001    & 30         & 600                   & -125.9  & 1000     \\
Sparse Reg High Dim & 0.0012   & 50         & 2000                  & -538.36 & 2000 \\
                    & \multicolumn{4}{c}{\texttt{RadialFlow}}                                              \\
                    \hline\\
                    & \textbf{\#LAYER}  & \textbf{MEDIAN}        & \textbf{IQR}                  & \textbf{\#NAN}     \\
                    \hline\\
Lin Reg             & 5        & -434.14    & (-434.36, -434.12)    & 0       &     \\
Lin Reg             & 10       & -434.12    & (-434.19, -434.12)    & 0       &   \\
Lin Reg             & 20       & -433.82    & (-434.21, -433.73)    & 0       &  \\
Lin Reg Heavy       & 5        & 111.99     & (111.91, 112.01)      & 0       &   \\
Lin Reg Heavy       & 10       & 111.99     & (111.89 , 112.00)     & 0       &   \\
Lin Reg Heavy       & 20       & 111.7      & (111.61, 111.71)      & 0       &  \\
Log Reg             & 5        & -251.33    & (-251.49, -251.31)    & 0       &     \\
Log Reg             & 10       & -251.17    & (-251.21 , -251.06)   & 0       &     \\
Log Reg             & 20       & -250.96    & (-250.97, -250.95)    & 0       &   \\
                    & \multicolumn{4}{c}{\texttt{RadialFlow}}                                              \\
                    \hline\\
                    & \textbf{\#LAYER}  & \textbf{MEDIAN}        & \textbf{IQR}                  & \textbf{\#NAN}     \\
                    \hline\\
Poiss Reg           & 5        & 82570.5    & (82570.19, 82570.57)  & 0       &  \\
Poiss Reg           & 10       & 82571.02   & (82570.76 , 82571.05) & 0       &  \\
Poiss Reg           & 20       & 82571.08   & (82570.90, 82571.12)  & 0       &   \\
Student t Reg       & 5        & -145.61    & (-145.62, -145.61)    & 0       &   \\
Student t Reg       & 10       & -145.59    & (-145.60, -145.59)    & 0       &    \\
Student t Reg       & 20       & -145.58    & (-145.60, -145.57)    & 0       &    \\
                    & \multicolumn{4}{c}{\texttt{RadialFlow}}                                              \\
                    \hline\\
                    & \textbf{\#LAYER}  & \textbf{MEDIAN}        & \textbf{IQR}                  & \textbf{\#NAN}     \\
                    \hline\\
Sparse Reg          & 5        & -127.28    & (-127.37, -127.27)    & 0       &   \\
Sparse Reg          & 10       & -127       & ( -127.03, -126.88)   & 0       &     \\
Sparse Reg          & 20       & -126.68    & ( -126.69, -126.61)   & 0       &    \\
Sparse Reg High Dim & 5        & -548.05    & (-548.54, -547.98)    & 0       &  \\
Sparse Reg High Dim & 10       & -547.2     & (-547.74, -547.16)    & 0       &   \\
Sparse Reg High Dim & 20       & -547.03    & (-547.78, -546.76)    & 0       &  
\end{tabular}
\end{table}

\begin{table}[]
\centering
\caption{Minimum, average, and standard deviation of KSD obtained from NEO for each of the examples across all tuning settings. \#Fail indicates the number of tuning settings that resulted in NaN outputs. The final KSDs obtained from MixFlow are also included for comparion.}
\label{tab:aggregated_neo}
\begin{tabular}{cccccc}
                              & \textbf{MIXFLOW KSD} & \textbf{MIN. KSD} & \textbf{AVG. KSD} & \textbf{SD. KSD} & \textbf{\#FAIL} \\
                               \hline\\
\texttt{banana}  & 0.06                       & 0.06              & 1.3               & 1.69             & 7/24            \\
\texttt{cross}     & 0.13                     & 0.11              & 2.78              & 3.09             & 12/24           \\
\texttt{Neal's funnel} & 0.04                 & 0.04              & 0.76              & 1.84             & 6/24            \\
\texttt{warped Gaussian} & 0.15               & 0.23              & 0.55              & 0.25             & 0/14            \\
\texttt{linear regression} & 1.64             & 9.37              & 143.56            & 84.16            & 18/24           \\
\texttt{linear regression (heavy tail)} & 51.43 & 396.97            & 1589.71           & 599.47           & 18/24           \\
\texttt{logistic regression} & 0.34           & 0.72              & 19.28             & 11.7             & 18/24           \\
\texttt{Poisson regression} & 10.57            & 639.08            & 1642.81           & 670.8            & 18/24           \\
\texttt{t regression} & 0.11                  & 0.15              & 4.79              & 5.03             & 12/24           \\
\texttt{sparse regression}  & 0.44            & 1.26              & 43.83             & 25.58            & 18/24          
\end{tabular}
\end{table}

\begin{table}[]
    \caption{ELBO results of \texttt{UHA} with different
number of leapfrogs and number of refreshments on real data examples. 
Each setting of \texttt{UHA} is run in $5$ trials. 
ELBOs of \texttt{UHA} are estimated using $5000$ independent samples.} 
\label{tab:uha}
\centering
\begin{tabular}{ccccccccccccccccccccc}
\hline\\
& \textbf{\#LFRG} & \textbf{\#REF} & \textbf{MEDIAN} & \textbf{IQR} & \textbf{\#SAMPLE} \\
\hline\\
Lin Reg                	& 10		& 5				& -430.438		& (-430.488, -430.411)		& 5000      \\
Lin Reg                	& 20		& 5				& -430.589		& (-430.596, -430.541)		& 5000      \\
Lin Reg                	& 50		& 5				& -432.215		& (-432.268, -432.146)		& 5000      \\
Lin Reg                	& 10		& 10			& -430.468		& (-430.511, -430.457)		& 5000      \\
Lin Reg                	& 20		& 10			& -431.404		& (-431.441, -431.202)		& 5000      \\
Lin Reg                	& 50		& 10			& -432.462		& (-432.56, -432.327)		& 5000      \\
Lin Reg Heavy           & 10		& 5				& 120.188		& (120.08, 120.401)		& 5000      \\
Lin Reg Heavy           & 20		& 5				& 119.126		& (118.806, 119.796)		& 5000      \\
Lin Reg Heavy           & 50		& 5				& 119.134		& (118.994, 119.247)		& 5000      \\
Lin Reg Heavy           & 10		& 10			& 118.222		& (118.125, 118.245)		& 5000      \\
Lin Reg Heavy           & 20		& 10			& 118.191		& (118.151, 118.202)		& 5000      \\
Lin Reg Heavy           & 50		& 10			& 118.059		& (117.718, 118.175)		& 5000      \\
Log Reg                	& 10		& 5				& -250.588		& (-250.607, -250.553)		& 5000      \\
Log Reg                	& 20		& 5				& -250.555		& (-250.565, -250.551)		& 5000      \\
Log Reg                	& 50		& 5				& -250.601		& (-250.64, -250.567)		& 5000      \\
Log Reg                	& 10		& 10			& -250.643		& (-250.662, -250.584)		& 5000      \\
Log Reg                	& 20		& 10			& -250.578		& (-250.615, -250.563)		& 5000      \\
Log Reg                	& 50		& 10			& -250.913		& (-251.006, -250.839)		& 5000      \\
Poiss Reg                	& 10		& 5				& 82578.439		& (82578.43, 82578.459)		& 5000      \\
Poiss Reg                	& 20		& 5				& 82578.453		& (82578.372, 82578.522)		& 5000      \\
Poiss Reg                	& 50		& 5				& 82578.421		& (82578.342, 82578.465)		& 5000      \\
Poiss Reg                	& 10		& 10			& 82578.45		& (82578.435, 82578.488)		& 5000      \\
Poiss Reg                	& 20		& 10			& 82578.449		& (82578.368, 82578.458)		& 5000      \\
Poiss Reg                	& 50		& 10			& 82578.369		& (82578.322, 82578.414)		& 5000      \\
Student t Reg		& 10		& 5				& -145.46		& (-145.461, -145.438)		& 5000      \\
Student t Reg		& 20		& 5				& -145.424		& (-145.451, -145.423)		& 5000      \\
Student t Reg		& 50		& 5				& -145.427		& (-145.43, -145.421)		& 5000      \\
Student t Reg		& 10		& 10			& -145.441		& (-145.454, -145.404)		& 5000      \\
Student t Reg		& 20		& 10			& -145.439		& (-145.466, -145.411)		& 5000      \\
Student t Reg		& 50		& 10			& -145.41		& (-145.417, -145.407)		& 5000      \\
Sparse Reg			& 10		& 5				& -125.936		& (-125.953, -125.931)		& 5000      \\
Sparse Reg			& 20		& 5				& -125.954		& (-125.981, -125.944)		& 5000      \\
Sparse Reg			& 50		& 5				& -126.279		& (-126.28, -126.226)		& 5000      \\
Sparse Reg			& 10		& 10			& -125.978		& (-125.988, -125.931)		& 5000      \\
Sparse Reg			& 20		& 10			& -126.032		& (-126.057, -125.961)		& 5000      \\
Sparse Reg			& 50		& 10			& -126.557		& (-126.559, -126.525)		& 5000      \\
Sparse Reg High Dim	& 10		& 5				& -501.201		& (-501.201, -500.778)		& 5000      \\
Sparse Reg High Dim	& 20		& 5				& -499.467		& (-499.471, -499.387)		& 5000      \\
Sparse Reg High Dim	& 50		& 5				& -498.968		& (-499.001, -498.656)		& 5000      \\
Sparse Reg High Dim	& 10		& 10			& -488.7			& (-488.799, -488.561)		& 5000      \\
Sparse Reg High Dim	& 20		& 10			& -487.216		& (-487.662, -487.042)		& 5000      \\
Sparse Reg High Dim	& 50		& 10			& -486.423		& (-486.477, -485.83)		& 5000      \\
\end{tabular}
\end{table}